\documentclass{article} % For LaTeX2e
\usepackage{iclr2027_conference,times}

\usepackage{graphicx}
\usepackage{booktabs}
\usepackage{multirow}
\usepackage[most]{tcolorbox}
\newcommand{\titletext}{GameBoyWorlds: A Testbed for \\Self-Improvement in Embodied Video Games}
\usepackage{amsmath,amsfonts,bm}

\def\eqref#1{equation~\ref{#1}}
\def\1{\bm{1}}

\DeclareMathAlphabet{\mathsfit}{\encodingdefault}{\sfdefault}{m}{sl}
\SetMathAlphabet{\mathsfit}{bold}{\encodingdefault}{\sfdefault}{bx}{n}

\usepackage{hyperref}
\usepackage{url}

\usepackage{enumitem}

\title{\titletext}

\author{
Dhananjay Ashok\textsuperscript{1}, Adam Shen\textsuperscript{2, \thanks{Equal contribution, order alphabetical.} }, Aslan Huo Feng\textsuperscript{3, *}, Chinmay Khanna\textsuperscript{1, *},\\
\bfseries Jun Rui Huang\textsuperscript{1, *}, Raghav Sarmukaddam\textsuperscript{1, *}, Surendira Balaji Natarajan\textsuperscript{1, *}, \\
\bfseries Xiaotong Cui\textsuperscript{1, *},
Xincan Zhang\textsuperscript{4, *}, Thomson Yen\textsuperscript{2}, Hongseok Namkoong\textsuperscript{2},\\
\bfseries Jonathan May\textsuperscript{1} and Jesse Thomason\textsuperscript{5}\\[1.5ex]
\normalfont\small\textsuperscript{1}Information Sciences Institute, University of Southern California \quad \textsuperscript{2}Columbia University\\
\normalfont\small\textsuperscript{3}University of Chicago \quad \textsuperscript{4}Purdue University \quad \textsuperscript{5}Georgia Institute of Technology\\
}

\iclrfinalcopy % Uncomment for camera-ready version, but NOT for submission.
\begin{document}

\maketitle

\begin{abstract}
Powered by expert guidance, embodied agents can operate in interactive environments; however, it is unclear whether they can learn autonomously and continually from their own experience. 
To evaluate such self-improvement methods, we introduce a two-legged benchmark: \textbf{GameBoyWorlds}, the first testbed for agentic self-improvement in complex video games~\footnote{https://dhananjayashok.github.io/GameBoyWorlds-Benchmark/}. \textbf{GameBoyWorlds-Execution} evaluates task execution on a collection of 5 distinct game series. Agents are allowed access to dedicated training games but are provided no demonstrations, documentation, or rewards. Agents must ground themselves in the environment through self-directed exploration and by inferring actionable knowledge from their own experience. 
At test time, agents must complete short-horizon execution tasks that evaluate their ability to navigate, interact, and engage with game-specific mechanics (e.g., combat) in potentially unseen games. Out-of-the-box frontier models complete fewer than 50\% of the 500 tasks due to failures in multimodal grounding, establishing that self-improvement methods have considerable room to push performance. We demonstrate that contemporary approaches to self-improvement are lacking, with standard implementations of world modelling and autonomous skill discovery failing, and a novel strategy that uses curiosity-based exploration to write guides achieving only partial success. \textbf{GameBoyWorlds-Playthrough} tests end-to-end game completion in two fan-made Pokémon games. We show that while frontier models have been pre-exposed to official releases such as Pokémon Red, they lack essential information on the games in our testbed. Instead of relying on their parametric knowledge to succeed, agents must learn from their own experience and autonomously improve over the course of the playthrough. We show that a sophisticated agentic pipeline with multimodal memory and hierarchical subgoals fails to reach even the first major milestone in both games, establishing GameBoyWorlds as an ambitious target for self-improving agents.
\end{abstract}

\section{Introduction}
\label{sec:introduction}
Driven by advances in Deep Learning~\citep{lecun2015deep}, Natural Language Processing~\citep{acl-2026-long} and Computer Vision~\citep{iccv2025}, embodied agents perform complex tasks in interactive environments~\citep{duan2022survey}. Reinforcement learning (RL) creates policies that navigate homes~\citep{savva2019habitat}, Minecraft agents cross several game milestones~\citep{fan2022minedojo}, and Vision-Language Action (VLA) agents perform dexterous manipulation~\citep{intelligence2025pi_}. However, these successes rest on human artifacts. RL requires hand-crafted reward signals, systems such as MineDojo rely on human gameplay, and VLAs like $\pi_0$ are pretrained on large corpora of human teleoperation. This reliance on human expertise and labour is a critical dependency, making progress contingent on data that is challenging to scale~\citep{doshi2024scaling}. 

% JON: could cite some major investment in human demonstration

An alternative data source is an agent's experience, i.e., data generated via interaction with an environment~\citep{SilverWelcomeTT}. The paradigm of self-improvement leverages this data, allowing embodied agents to continually improve without expert guidance~\citep{ghasemipour2025selfimproving}. Self-improving systems could autonomously adapt to unfamiliar environments and may develop superhuman capabilities over the course of their deployment~\citep{silver2017mastering, jumper2021highly}. 

% JON: could cite major investment in recursive self improvement

Humans are experts at continual self-improvement, as demonstrated when they start a new video game~\citep{pmlr-v80-dubey18a}. A child playing Pokémon for the first time can learn through experience alone. After understanding character movement and interaction, they must use these execution capabilities to play through the game. From memorising maps to understanding the use-cases of the game's many items, the child continues to learn as they play. Eventually, they acquire the capacity to achieve sophisticated, long-term goals in what was once an unfamiliar environment. 

\begin{figure}
    \centering
    \includegraphics[width=\linewidth]{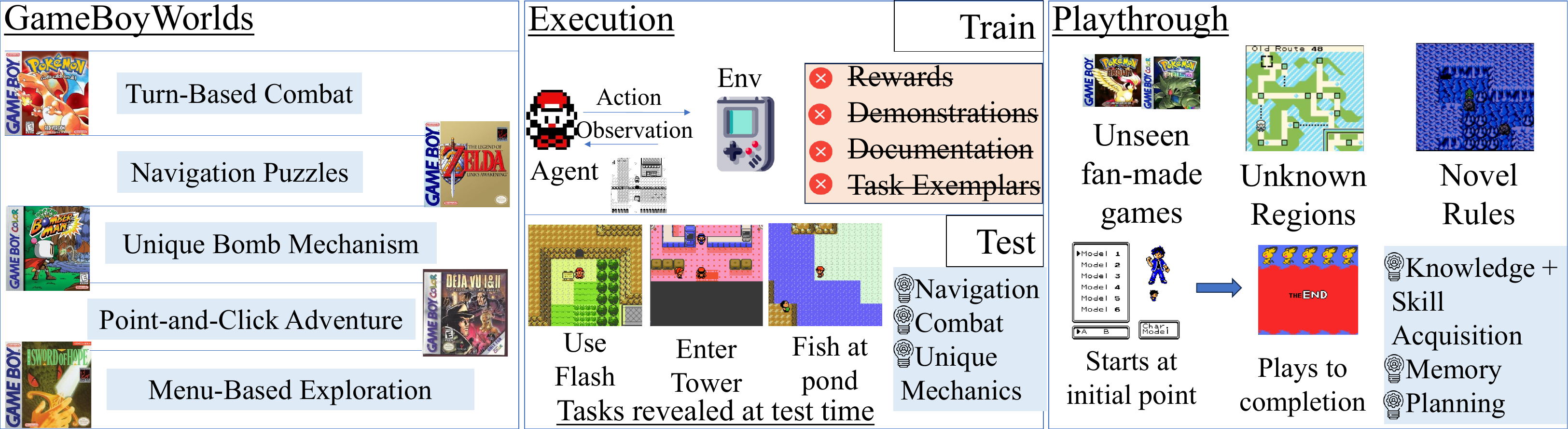}
    \caption{GameBoyWorlds, a testbed for embodied self-improvement based on a collection of diverse games from the GameBoy universe (left). The execution leg (middle) allows agents to conduct self-driven exploration in game environments and tests their short-horizon task execution capabilities. The playthrough leg (right) tests deployment-time self-improvement in unseen games.}
    \label{fig:intro}
\end{figure}

Existing game benchmarks are evaluation-only~\citep{zhang2025videogamebench, paglieri2025balrog, ouyang2026gameworld, hu2026lmgamebench} or train with rewards from test environments~\citep{mnih2013playing, kempka2016vizdoom, vinyals2017starcraft, hambro2022insights, hafner2022benchmarking, pleines2025pokemon, park2026orak}. While frontier model releases demonstrate that agents can play through video games~\citep{comanici2025gemini, bousquette2026pokemon}, they do so on popular titles like Pokémon Red, which are highly contaminated in the pretraining corpora of the models~\citep{karten2025pokeagent}. It remains unclear whether agents can acquire capabilities and information through self-improvement alone, as opposed to relying on their pretrained knowledge of the domain. To evaluate models for these capabilities, we introduce a two-legged testbed defined on games from the GameBoy universe: \textbf{GameBoyWorlds} (Figure~\ref{fig:intro}), the first benchmark for self-improvement in complex video-game environments. 

The first leg is \textbf{GameBoyWorlds-Execution}, a testbed for short-horizon task execution across five different game series, with each marking a dedicated training game and a distinct, test-only game. The games include a diverse range of navigation dynamics and specialized mechanics. For example, games from the Pokémon series feature top-down, egocentric navigation and turn-based combat, while games from the Deja Vu series use menu-based navigation and tool-based interaction. 
During training, agents are given free access to the training game, but are provided with no task exemplars, demonstrations, documentation, or game-specific reward signals. Agents must conduct self-driven exploration to form an understanding of each game's unique mechanics and acquire skills relevant to the game series. At test time, agents are placed in a moment of gameplay and must execute a task specified in natural language. Success requires agents to be adept at multimodal perception, localized planning and execution of game-specific mechanics. For example, given the task `Activate the switch using a bomb' in the game `Bomberman Quest', agents must locate the switch on the screen, plan a path through obstacles to reach it, trigger the game-specific mechanic of planting a bomb on the correct tile and navigate away from the bomb's blast radius to complete the task. With 50 hand-crafted tasks per game, we test agents on a wide range of 500 tasks, including navigation, interaction, combat and game-specific control across five unique game titles and 10 distinct games.

Frontier models struggle without game-specific adaptation, with task completion below 50\% for Gemini-3.6-Flash, GPT-5-Mini and Claude-4.5-Haiku due to consistent failures in perception and poor game-mechanic understanding. While self-improvement procedures have substantial room to push performance, standard implementations of world modelling~\citep{Hafner2020Dream} and autonomous skill discovery~\citep{zhou2025proposeragentevaluator} fail catastrophically, degrading agent performance. We introduce a novel pipeline that uses curiosity-driven exploration~\citep{burda2018largescale} as a search procedure to surface `surprising' subtrajectories of gameplay. We sweep these subtrajectories with the VLM and distill insights from them into game-specific guidance documents to aid task execution. Access to these documents at test time can improve task completion by 22\% on select games, establishing our approach as a viable means of self-improvement. However, performance drops by 30\% in the worst case, suggesting that more powerful approaches are needed to fully address the testbed. 

The second leg, \textbf{GameBoyWorlds-Playthrough}, mirrors the experience of a human playing an unfamiliar game and evaluates an agent's ability to play two fan-made Pokémon games from start to finish. In contrast to frontier model releases, our testbed uses scarcely documented fan-made games with novel regions and game mechanics. While frontier models score near-perfectly in knowledge checks on games like Pokémon Red, they achieve less than 30\% accuracy on questions regarding how to progress through the games in our testbed. Instead of relying on parametric knowledge, agents must acquire knowledge as well as the capacity to conduct long-term planning directly from their experience in the games. We craft an initial agent architecture, complete with three-tiered hierarchical controllers, dedicated memory systems, and self-reflective plan creation. This system fails to clear the first major milestone (obtain your first Pokémon) in both games, establishing the testbed as an ambitious target for agents that continually self-improve during deployment. 

We release our testbeds with an emulation system, providing unified interfaces for any GameBoy game. We hope GameBoyWorlds promotes further research in self-improving embodied agents. 

\section{GameBoyWorlds-Execution: Embodied Learning in Games}
\label{sec:execution}
We model games as Partially Observable Markov Decision Processes $(\mathcal{S}, \mathcal{A}, \mathcal{O}, \mathcal{T}, \mathcal{E})$. $\mathcal{S}$ denotes emulator states; $\mathcal{A} = \{\text{$\uparrow$, $\downarrow$, $\leftarrow$, $\rightarrow$, A, B, STRT}\}$ is an action space of buttons.
The observation space $\mathcal{O}$ consists of raw screen frames. 
$\mathcal{T}: \mathcal{S} \times \mathcal{A} \times \mathcal{S} \to [0, 1]$ defines the probability of transitioning to a subsequent state given a state and action, and the observation emission function $\mathcal{E}: \mathcal{S} \to \mathcal{O}$ is deterministic. No extrinsic, game-specific rewards are provided. At test time, agents are given start state $s_0\in\mathcal{S}$, task $t$ specified in natural language and a completion function $\mathcal{C}_t: \mathcal{S} \to \{0, 1\}$. 
Task $t$ is complete if the agent generates $(a_1, \ldots a_c), a_i\in\mathcal{A}$ such that $\mathcal{C}_t(s_c) = 1$ where $s_i \sim \mathcal{T}(s_{i-1}, a_i)$. Our primary metric of interest is \textbf{task completion gain}, i.e., the increase in task completion after self-improvement in the training environments, as opposed to the initial capabilities of the agent. We define self-improvement as procedures that \textbf{exclusively use data generated from the agent's interactions with the training environment}; i.e., not distillation from human trajectories, a stronger teacher or external knowledge bases. While reward shaping is permitted, we require all reward designs to be intrinsic, i.e., task- and game-agnostic. We only allow training in the \textbf{first installment} of each series, forcing approaches to extract insights that apply across each game series. 

We port 10 games from five series, complete with 50 tasks and completion functions each, for a total of 500 test tasks across our benchmark. The games are: 
\begin{itemize}[nosep,noitemsep]
\item \textbf{Pokémon} (Red and Crystal), a role-playing game set in a world with creatures known as Pokémon, with navigation, combat, and interaction tasks like `Fish for a Pokémon'. \item \textbf{Legend of Zelda} (Link's Awakening and Oracle of the Seasons), top-down exploration, real-time combat, and dungeons with navigation tasks like `Go to the lava floor'. \item \textbf{Deja Vu} (1 and 2), A point-and-click adventure in a 1940s noir setting. Tasks involve tool-based interactions like `Unlock the door'. \item \textbf{Sword of Hope} (1 and 2), an RPG where the player uses a command menu to interact with their surroundings. Tasks include combat tests like `Win a battle encounter'. \item \textbf{Bomberman} (Quest and Pocket), a strategic, maze-based role-playing game requiring navigating grid-like arenas and mechanics like `Defeat the Skull Enemy using a bomb'. 
\end{itemize}
Completion functions are implemented with frame comparisons, providing robust success verification, and all tasks are executable by humans. For implementation details, see Appendix~\ref{app:gameboyworlds}.

%\noindent\textbf{Harry Potter} (1 and 2): An RPG based on the fantasy series where the player controls Harry and explores the Hogwarts castle. Tasks include interaction tests like `Buy books at Flourish and Blotts'. 

%\noindent\textbf{Runes of Virtue} (1 and 2): An action RPG of the \textit{Ultima} franchise. The player navigates underground labyrinths in real-time. Tasks involve puzzle tests like `Unlock the basement ladder'. 

%\noindent\textbf{Survival Kids} (1 and 2): A survival-focused RPG where the player is stranded on a deserted island. Tasks involve resource management tests like `Cook the meat over the fire'. 

%\noindent\textbf{Harvest Moon} (1, 2 and 3): A farming RPG where the player is tasked with restoring a neglected farm. Tasks include livestock management tests like `Buy a cow and name it LUNA'. 

\section{Evaluating Baseline Task Execution Capabilities}
\label{sec:evaluation}

\begin{figure}[th]
    \centering
    \includegraphics[width=0.484\linewidth]{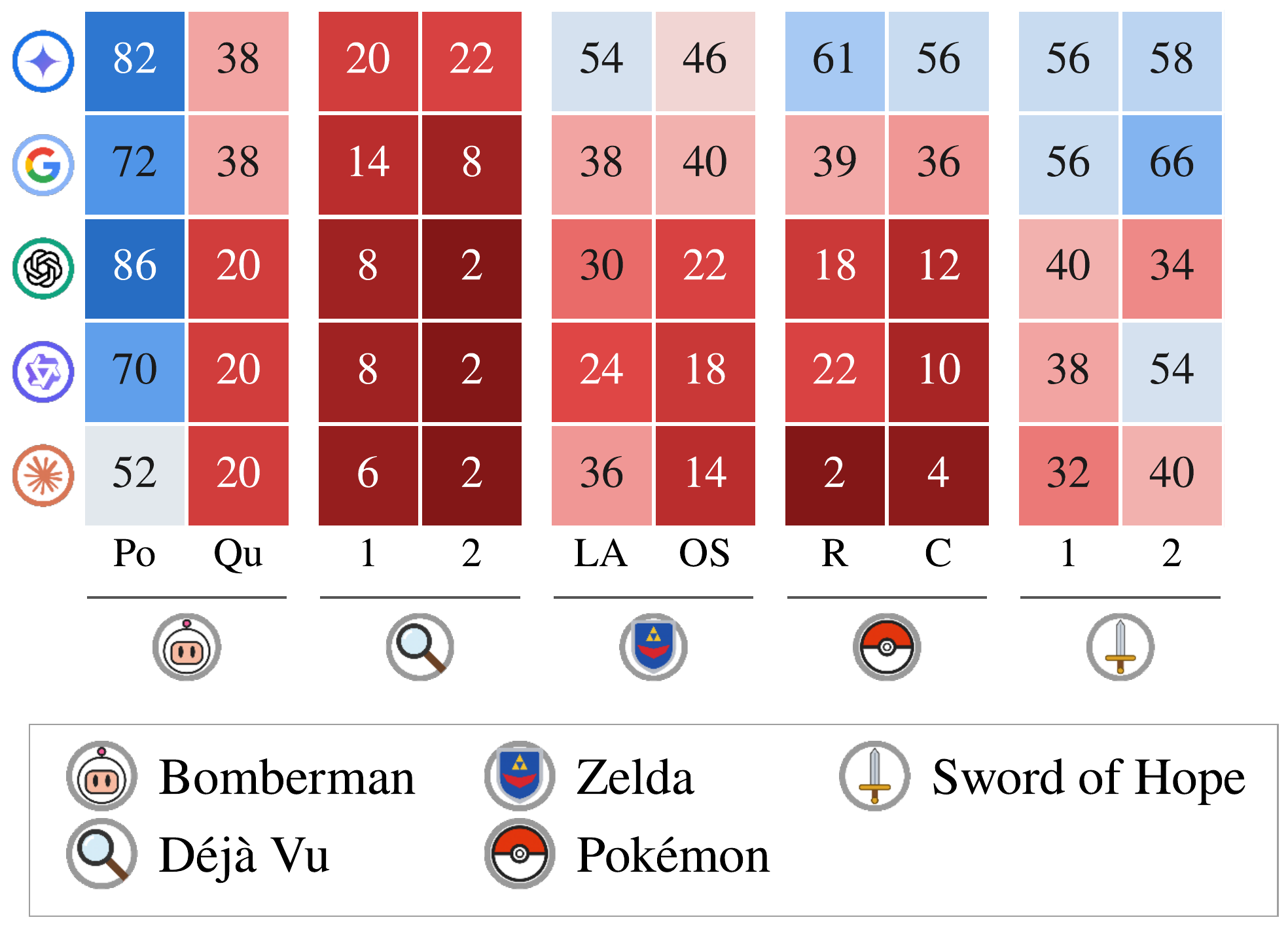}\hfill
    \includegraphics[width=0.496\linewidth]{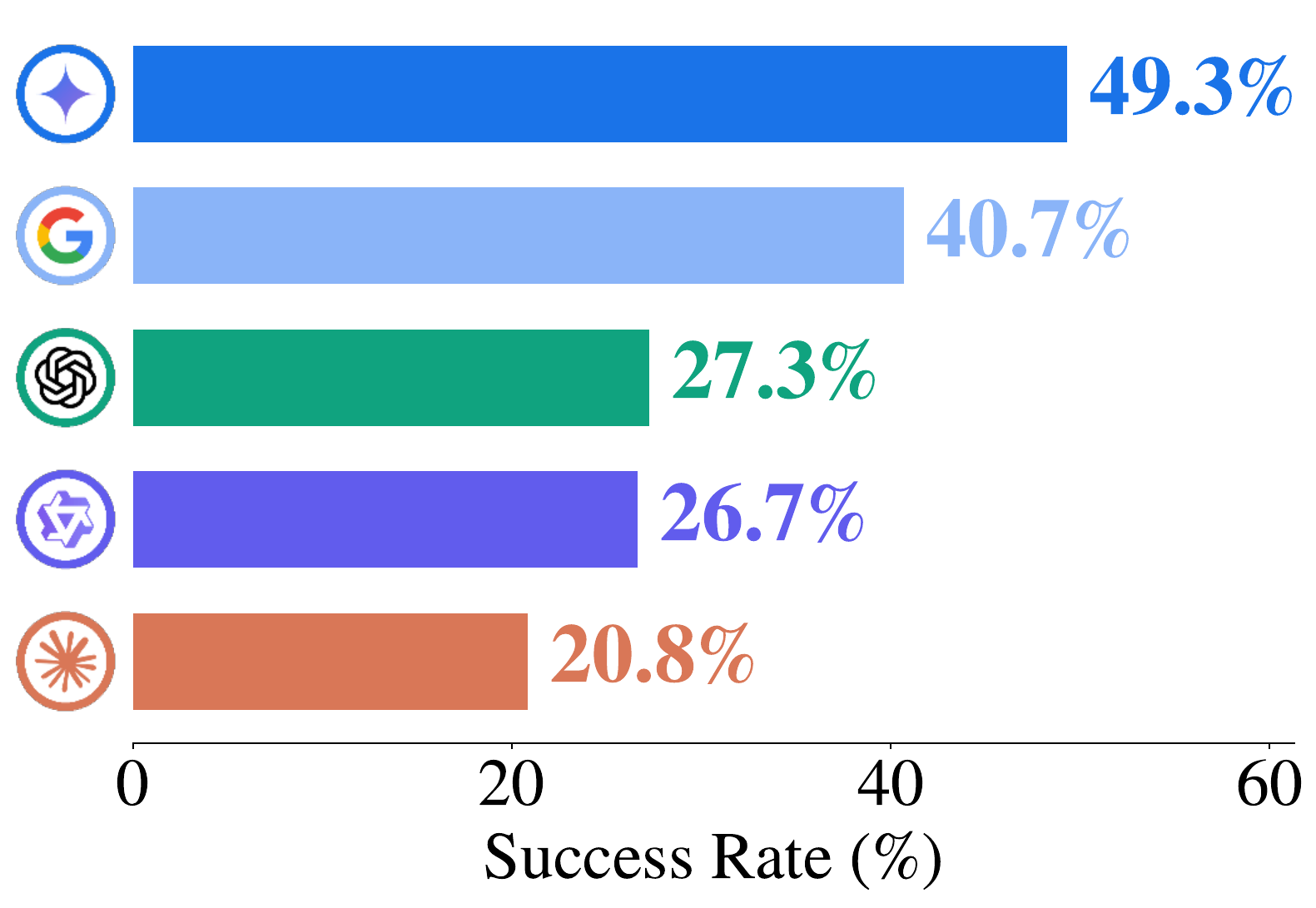}
    \par\vspace{4pt}
    \includegraphics[width=0.9\linewidth]{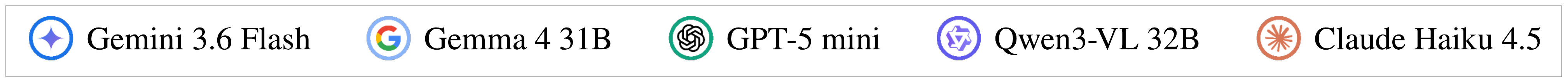}
    \caption{Model performance when deployed on GameBoyWorlds-Execution. Left: success rate (\%) per game. Right: overall success rate. Models fail to exceed 50\% average task completion.}
    \label{fig:benchres}
\end{figure}
We first benchmark frontier models when deployed out-of-the-box. After several initial designs, we settle on a hierarchical framework~\citep{belkhale2024rt} consisting of a supervisor and executor. The supervisor handles \textbf{subgoal planning and error correction:} Given a task and observation, a supervisor decomposes the task into subgoals, which are passed to the executor. After execution, the supervisor judges if the subgoal is cleared, and if so, issues the next subgoal. Otherwise, the supervisor critiques the steps, diagnoses the failure, and redeploys the executor with a hint. \textbf{Self-Revision:} Should the executor fail despite multiple rounds of hints, the supervisor considers whether the subgoals are flawed. If so, it creates a new set of subgoals and retries. 

At every step, the executor reasons over the supervisor's instructions and its history to decide on a single button press. Alternatives such as outputting sequences do not improve performance. After every step, the executor describes the change in the screen. This description is tracked along with the history and used in all functions that involve understanding an executor's steps (e.g., supervisor error correction). Visual history consistently improved performance relative to alternative schemes. 

The supervisor and executor share the same underlying VLM, and each part of the agent framework is implemented with game-agnostic prompts. For implementation details, complete with ablations over alternative frameworks that were considered, refer to Appendix~\ref{app:zeroshot}. We evaluate Gemini-3.6-Flash, Claude-4.5-Haiku~\citep{anthropic2024claude35sonnet} and GPT-5-Mini~\citep{singh2025openai}. We also evaluate open-source models: Gemma-4-31B and Qwen3-VL-30B-Instruct~\citep{bai2025qwen3}. 

\subsection{Results}

Frontier models often fail (Figure~\ref{fig:benchres}), with the best average performance obtained by Gemini-3.6-Flash at 49.3\%. Deja Vu, with its unique point-and-click mechanics, proves the hardest game with a maximum task success rate of 22\%. Bomberman Pocket is the easiest to solve, with GPT-5-mini achieving the highest success rate (84\% ). Open-source models compare favourably against GPT-5-mini and Claude-Haiku, and Gemini-4-31B significantly ($p<0.05$, Wilson Interval~\citep{wilson1927probable}) outperforms them. We hypothesize that since the training paradigm of the Gemma models involves distillation from Gemini, which shows superior benchmark performance among the frontier models, it may have obtained substantial multimodal grounding and perception. 

\begin{figure}[h]
    \centering
    \includegraphics[width=0.45\linewidth]{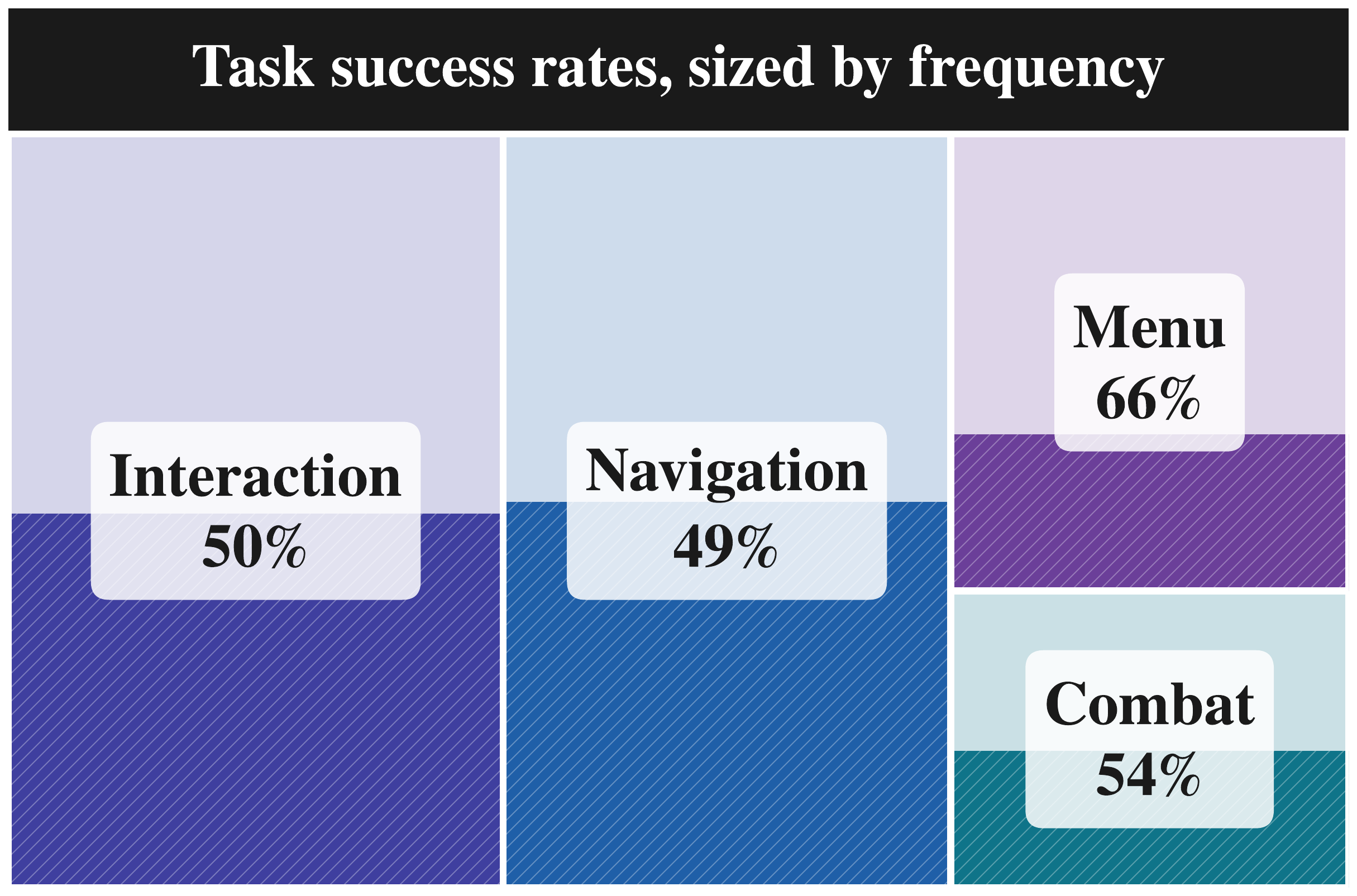}
    \includegraphics[width=0.45\linewidth]{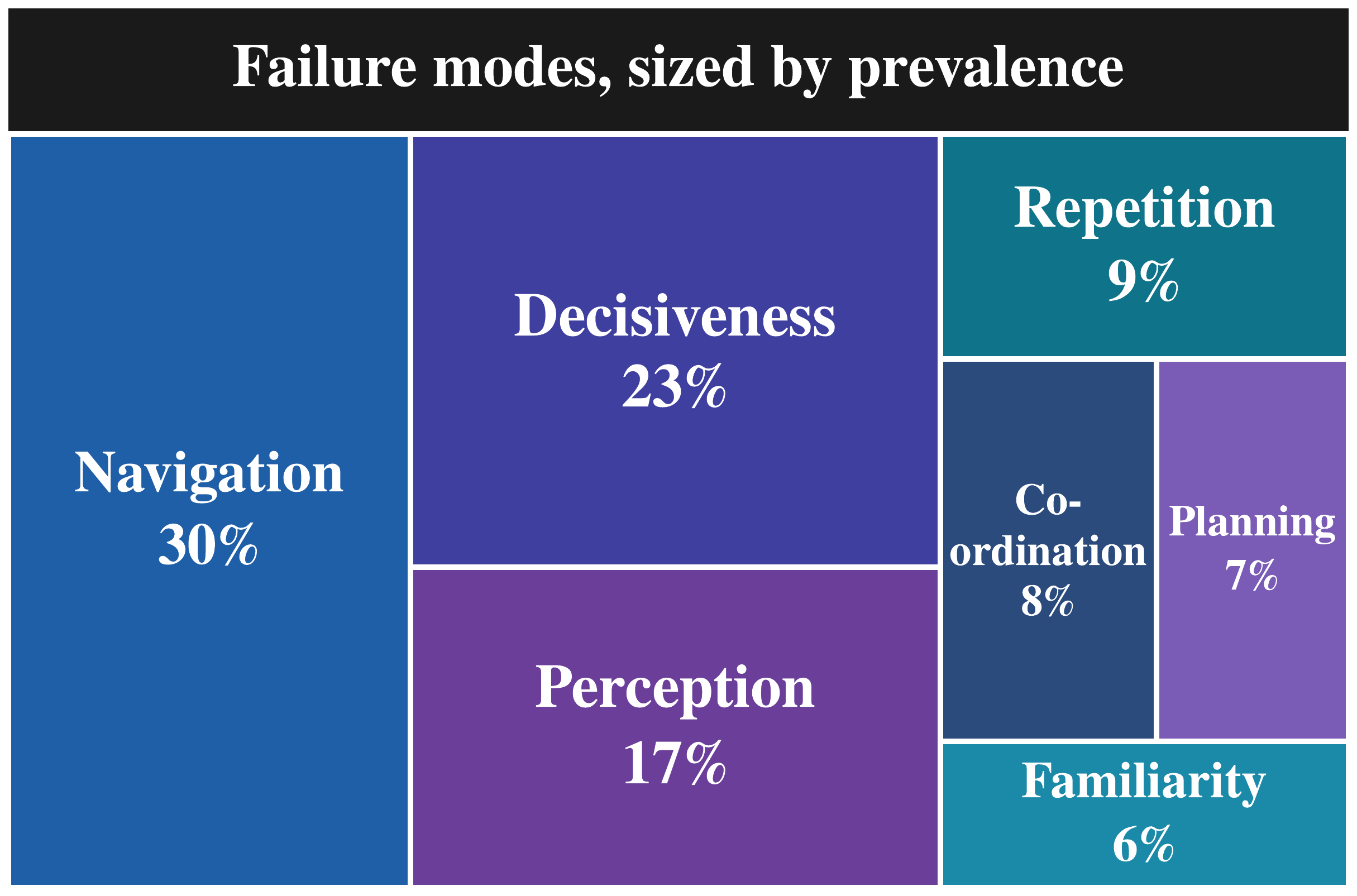}
    \caption{Success rate by common task categories (left) and failure modes (right). While models are often more proficient in menu usage, they are often poor at basic navigation and interaction.}
    \label{fig:pie}
\end{figure}

We annotate task categories and find (Figure~\ref{fig:pie}) that while models are relatively competent in static, menu-based tasks, they consistently struggle with the most basic interaction tasks, which often require the agent to pathfind to an object or NPC and then initiate an interaction. We use Claude-Code to pseudo-annotate failure modes for the Gemini-3.6-Flash model and verify a subset of the classifications by hand. We see that models consistently struggle with precise perception and alignment on screen, as well as timely interaction. Other failure modes include constant repetition of the same action sequence, poor coordination between agent hierarchies, and insufficient knowledge of game-specific rules. For examples and further discussion, see Appendix~\ref{app:zeroshot_examples}. Overall, agent performance is far from optimal, implying that models are well-situated for iterative self-improvement loops. 

\section{Evaluating Existing Self-Improvement Methods}
\label{sec:selfimprove}

Having established GameBoyWorlds-Execution's difficulty, we next evaluate approaches to the self-improvement paradigm in GameBoyWorlds. We consider standard implementations of dominant methods from self-supervised learning~\citep{doersch2015unsupervised}, unsupervised reinforcement learning~\citep{laskin2021urlb} and open-endedness~\citep{zhang2024omni}, when applied to Gemma-4-31B.  

\noindent\textbf{World Modeling:} World models are internal simulators that learn the dynamics of an environment. They train on autonomously collected experience data to estimate an observation-space equivalent of the transition function $\mathcal{T}$ with $\mathcal{W}: \mathcal{O}\times\mathcal{A}\to\mathcal{O}$. Given a strong world model, an agent can plan over the likely consequences of their actions~\citep{sutton1981adaptive} or cheaply train in a hallucinated environment~\citep{Hafner2020Dream, hafner2020mastering}. Having enabled strong performance in robotic manipulation tasks and game environments, world modelling is a powerful, general-purpose means of self-improvement. We train game-specific world models and deploy them under a planning framework. Specifically, we adapt the executor from above (Section~\ref{sec:evaluation}). Instead of predicting a single action, it queries the world model using the current observation and all possible actions. The executor then compares all predicted next observations simultaneously, identifies the next observation which seems to most advance towards its goal and selects the associated action (implementation details in Appendix~\ref{app:wm}).

\noindent\textbf{Autonomous Skill Discovery:} Recent work~\citep{zhang2023bootstrap, zheng2025skillweaver} leans on VLMs to bootstrap skill acquisition. By autonomously conducting rounds of self-evaluated practice, these methods accumulate skills without any human guidance. We adopt a variation of the Proposer-Agent-Evaluator framework~\citep{zhou2025proposeragentevaluator}, a pipeline with three distinct phases: zero-shot task proposal, task practice and behaviour cloning. Given a state in the training environment, the agent first inspects the screen and draws from its parametric knowledge of the domain to suggest tasks that could be accomplished in the given scenario. The agent attempts to complete the task and evaluates its own progress. If a task is judged to be successfully completed, it is carried over to the practice phase; otherwise, it is discarded. In the practice phase, the agent is placed in the starting state from which the task was completed, but before it is given control in the environment, a small number of random buttons are pressed. The agent then tries to complete the task from the perturbed starting state. If successful (as judged by self-reflection), the trajectory is saved to a separate fine-tuning dataset. We train a VLA on the dataset using behavioral cloning, and deploy it as the executor in the framework above (Section~\ref{sec:evaluation}). For specific implementation details, see Appendix~\ref{app:skill}.  

\subsection{Curiosity-based exploration} 
Curiosity-based RL uses intrinsic rewards that promote the visitation of novel states~\citep{burda2018exploration}. In simpler environments like Atari, curiosity can entirely obviate the need for extrinsic rewards ~\citep{burda2018largescale}. In comparison, GameBoy games are substantially more complex and cannot be cleared with curiosity alone. In this work, we adapt curiosity-based RL to serve as a search procedure to identify `interesting' snippets of gameplay. Our core intuition is that subtrajectories that end in a highly novel observation likely implicitly contain information regarding key, game-specific mechanics. We hope to capture a wide range of such `interesting' subtrajectories, and use them to infer game-specific insights that will be useful for downstream task execution. 

\subsubsection{Iterative Curiosity-Based Exploration as Search}
We reward the agent for observing frames that are different from the other frames in its experience buffer. Given frame observations of the GameBoy screen $o_1, o_2\in\mathbb{R}^{H\times W}$, we flatten and apply a random linear projection to each of them to obtain their latent representations $z_i\in\mathbb{R}^d$ and consider the cosine similarity between latents as an inverse measure of difference. For every step, given the buffer of previously seen screens $O\in\mathbb{R}^{N\times H \times W}$ and the latest observation $o$, we derive the latents and compute the intrinsic reward with $1-\max_{i\in[1, N]}(\text{sim}(z, z_i))$. Our design also includes: 

\noindent\textbf{Buffer Reset:} The buffer is reset at the start of every episode to ensure that the environment and reward schedule remain \textit{stationary}, a requirement for asymptotic convergence of RL methods~\citep{sutton1981adaptive}. In practice, we observe that agents trained under this schedule often learn a single `interesting' trajectory that seeks out a single, high-reward transition (e.g., entering new maps).

\noindent\textbf{Iterative exploration:} To discover a wider range of behaviours from the same initial state, we run multiple generations of curiosity-based exploration sequentially. Generations differ by the initial frame buffer used; the buffers of later generations are pre-populated with the frames seen by all previous generations and are always reset to this initial setting. In effect, agents trained in the second generation are only rewarded for discovering states that were not surfaced in the first generation.

\noindent\textbf{Hyperparameter Sweep:} RL is notoriously sensitive to hyperparameters~\citep{eimer2023hyperparameters}. However, variance in final policies leads to a diversity of high-reward trajectories. Hence, each generation of curiosity-based exploration is a sweep over hyperparameters. We retain the top $k$ runs as ordered by total reward obtained by the final policy as use them to populate the buffers of the next generation. 

\begin{figure}
    \centering
    \includegraphics[width=\linewidth]{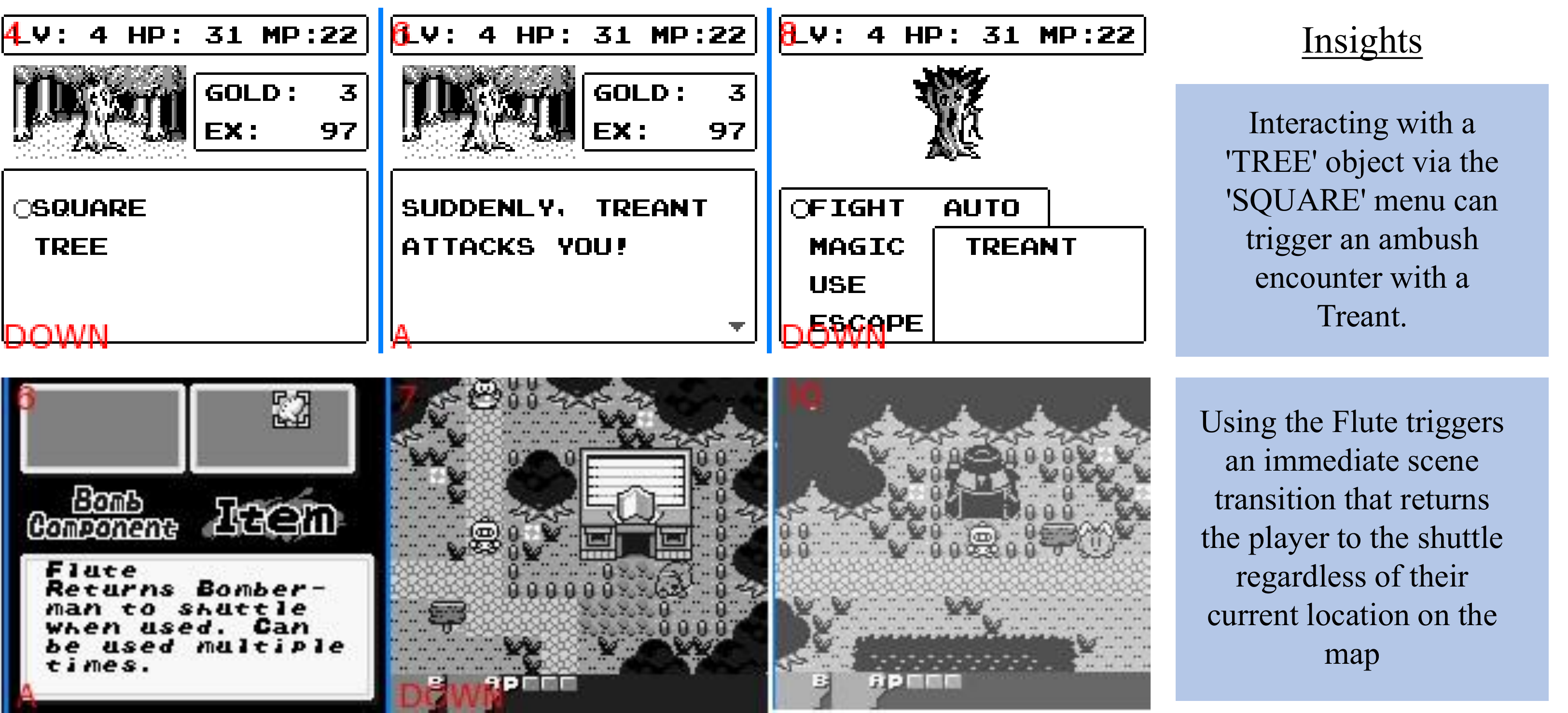}
    \caption{Insights drawn from `interesting' trajectories surfaced during curiosity based exploration.}
    \label{fig:curiosity}
\end{figure}

In practice, the exploration regularly surfaces behaviors that harness a game-specific mechanic (Figure~\ref{fig:curiosity}), even those that the VLA agent is unaware of or incapable of performing consistently. For complete implementation details and examples of discovered mechanics, see Appendix~\ref{app:curiosity}.

\subsubsection{Insight Distillation}
The previous step gives us an undifferentiated mass of interesting trajectories mixed with noisy trajectories that do not capture any unique game mechanic or task. We next use VLMs to autonomously extract actionable insights from these snippets of gameplay. We put the highest reward trajectories from all generations of curiosity exploration through the following pipeline:

\noindent\textbf{Terminal State Grouping:} We group trajectories based on the similarity of their final frames, with a simplifying assumption that sequences which end in similar frames expose similar game mechanics. 

\noindent\textbf{Trajectory Description}: We sample trajectories from each group and verbalize what occurs between each frame, with a reference on how the actions lead to a change in the screen. In effect, this process identifies the `task' being completed by the agent in the trajectory, e.g. `Read the sign'. If no task can be identified, the group is ignored, giving the VLM a path to filter noisy trajectories. The VLM is then prompted to generate non-obvious, game-specific insights that can be inferred from the trajectory. For example, in Deja Vu, the VLM observes a trajectory where the player inspects an item via the `eye' tool in the bottom bar. As a result, it extracts the insight: `to inspect an item, do not simply click on it, but first select the eye icon in the bottom right and then click on it'. 

\noindent\textbf{Insight Documentation}: Finally, the VLM is prompted to iteratively merge all of the uncovered insights and organize them into a list of entries in an insight document. Each entry in the document consists of a task description (e.g., `use a warp item'), a relevant scenario (e.g., `top-down overworld') and an actionable insight or guidance message relevant to that specific task. Given insight documents, we incorporate them into the supervisor during task execution. When the agent is provided with a new initial state and task, it first reviews every entry in its insight document and determines whether each entry is relevant to the task at hand. All relevant insights are combined into a single prompt, and this prompt is provided as additional knowledge context for every step that the supervisor engages in (i.e., subgoal creation, error correction and self-reflection). For prompts, intermediate artifacts and sample entries from insight documents, see Appendix~\ref{app:synthetic}. 

\subsection{Results}
\begin{figure}[h]
    \centering
    \includegraphics[width=\linewidth]{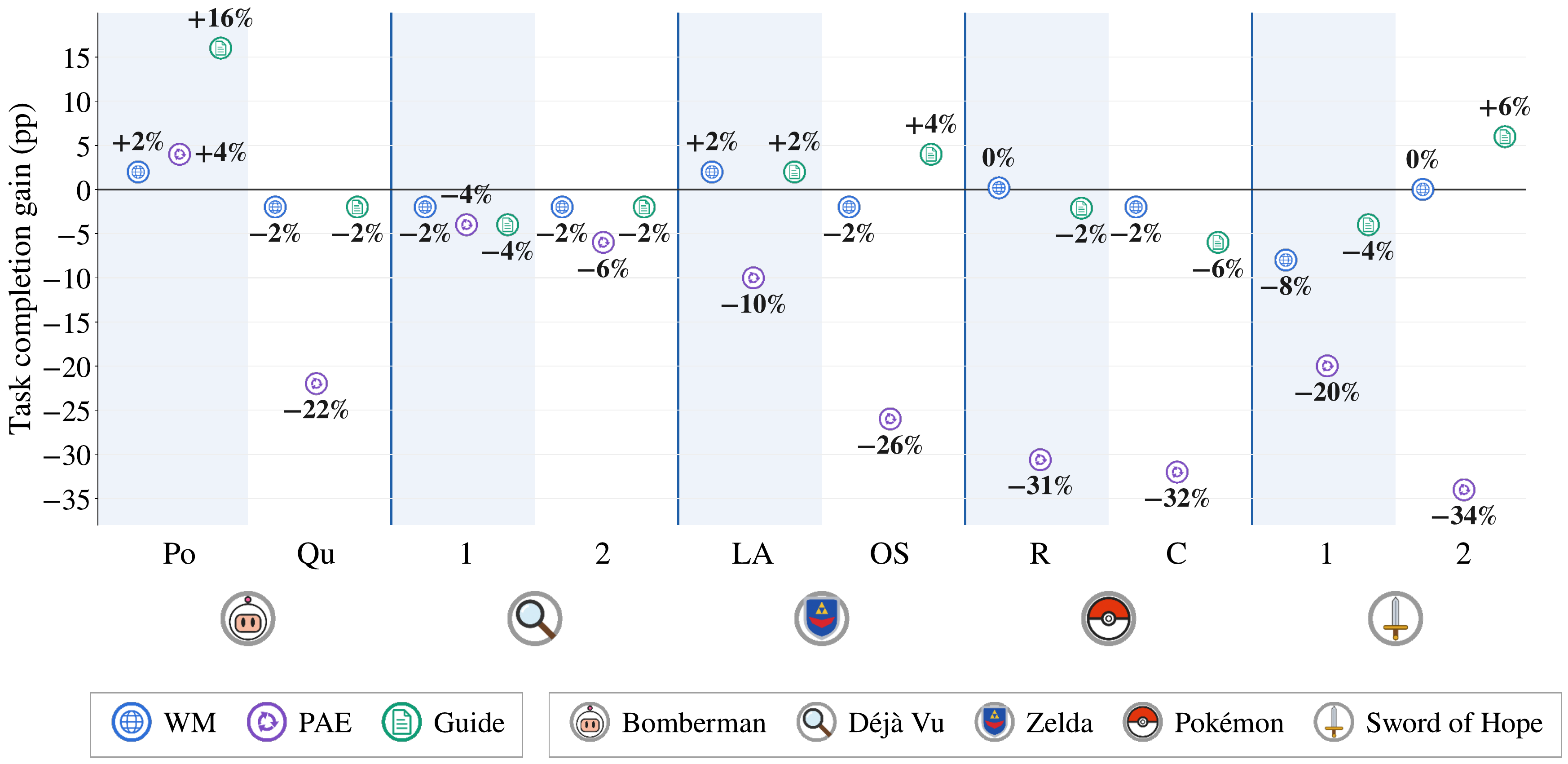}
    \caption{World modeling (WM) and autonomous skill acquisition (PAE) fail on GameboyWorlds-Execution. Our curiosity pipeline improves performance on average (+2\%), but lacks robustness.}
    \label{fig:selfimprove}
\end{figure}

No approach consistently improves performance across all game environments
(Figure~\ref{fig:selfimprove}). 

World modeling reduces
average task completion by $3.4\%$. Qualitatively, we observe that the world model is unable
to predict the next observation with sufficient accuracy to serve as a useful
tool. The skill discovery approach performs even worse, degrading task
completion by $44.3\%$ on average ($40.7\% \rightarrow 22.6\%$), with
catastrophic collapses on PokémonCrystal ($36.0\% \rightarrow 4.0\%$),
PokémonRed ($38.8\% \rightarrow 8.2\%$) and
ZeldaOracleOfSeasons ($40.0\% \rightarrow 14.0\%$). The performance degradation is always greater for test-only games, suggesting an overfitting to the training distribution. We hypothesize the extent of degradation is caused by an inability to accurately judge task completion during practice, a known failure scenario for skill acquisition methods~\citep{song2024mind}. The only method that achieves partial success is our novel utilization of
curiosity-based exploration to write game-specific documentation. Here, task
completion increases by as much as $22\%$ or $10\%$ on select games like
BombermanPocket ($72.0\% \rightarrow 88.0\%$) and
ZeldaOracleOfSeasons ($40.0\% \rightarrow 44.0\%$). The boost applies
to both games seen during training time and unseen test
games like SwordOfHope2 ($66.0\% \rightarrow 72.0\%$). However, even
this approach leads to considerable performance degradation on some games
(DejaVu1, $-29\%$; DejaVu2, $-25\%$; PokémonCrystal, $-17\%$), and the average task completion rate across improves by only $2.0\%$, ($40.7\% \to 41.5\%$). The
failure of these standard implementations establishes
GameBoyWorlds-Execution as a challenging benchmark and calls for further
research on acquiring short-horizon task execution capabilities via self-improvement.

\begin{figure}[th]
    \centering
    \includegraphics[width=\linewidth]{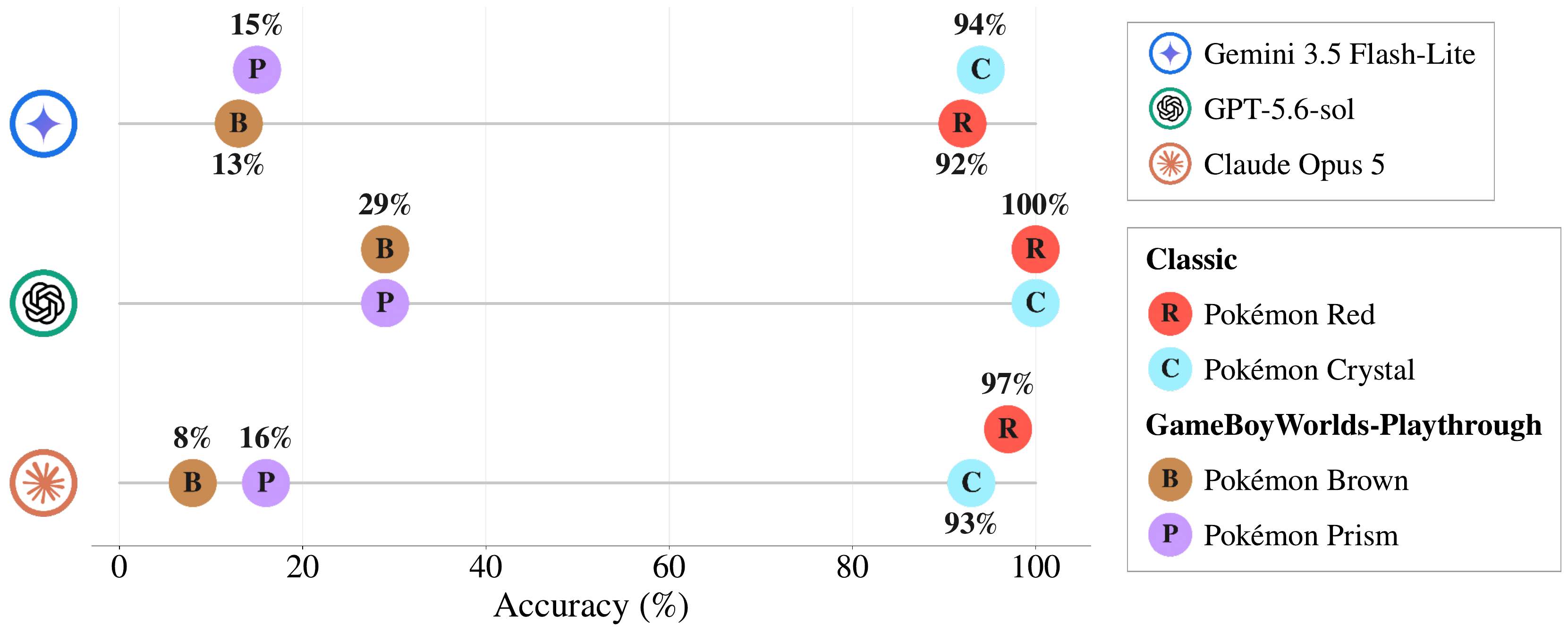}
    \caption{Contamination checks show that frontier models nearly perfectly recover the progression steps in the classic Pokémon games but recover less than 30\% of the steps in the fan-made versions.}
\label{fig:contamination-gap}
\end{figure}

\section{GameBoyWorlds-Playthrough: End-to-End Gameplay}
\label{sec:playthrough}

Playing a game through to completion requires dynamic multimodal memory, active knowledge acquisition, reasoning under uncertainty and long-term planning. We ask whether models are capable of self-improving during deployment to develop these capabilities in-game. 

\noindent\textbf{Existing benchmarks are heavily contaminated:} Frontier models can be benchmarked on Pokémon games as a means of establishing their planning and visual grounding abilities~\citep{karten2026continual}. However, key information on how to clear these games is already contained in the parametric knowledge of frontier models, before they even start the game. For example, when asked about how to clear Pokémon Red, GPT-5 already knows that to enter Saffron City, it can first navigate to the roof of the Celadon City Department Store, purchase a Fresh Water from a vending machine, and give it to the thirsty guards blocking the gates. Human players must discover this relatively arbitrary dependency through active exploration, but GPT-5 bypasses information-gathering entirely. The consequence of this contamination is that existing benchmarks are unable to evaluate whether agents can truly acquire information and improve during deployment. We address this gap with \textbf{GameBoyWorlds-Playthrough}, a testbed for start-to-finish completion on Pokémon Brown and Pokémon Prism, fan-made games that introduce novel regions and mechanics. In both games, agents start at the spawn and must clear 10 distinct milestones: obtaining your first Pokémon, obtaining each of the eight gym badges, and finally, defeating the Elite Four. 

We write questions that poll the model for contamination of progression steps and confirm (Figure~\ref{fig:contamination-gap}) that while frontier models nearly perfectly recover the progression steps in the classic Pokémon games (Red and Crystal), they recover less than 30\% of the steps in the fan-made versions. Success in GameBoyWorlds-Playthrough requires not only consistent short-horizon execution, but also long-term planning, active knowledge seeking, dynamic multimodal memory, and continual improvement during deployment. For details on the games and contamination checks, see Appendix~\ref{app:contamination}.

Poor model performance on GameBoy-Execution (Figure~\ref{fig:benchres}) suggests that additional scaffolds are required to enable long-horizon gameplay. We implement a workable framework that provides action interfaces and perception modules that enable navigation, mapping, interaction, hierarchical subgoal formation, and knowledge accumulation. These subsystems do not provide the model with game-specific knowledge or use privileged information (e.g., ROM memory) but rather grow as it accumulates experience. As implementation is Pokémon-specific, we defer details to Appendix~\ref{app:playthroughagent}.

\begin{figure}
    \centering
    \includegraphics[width=\linewidth]{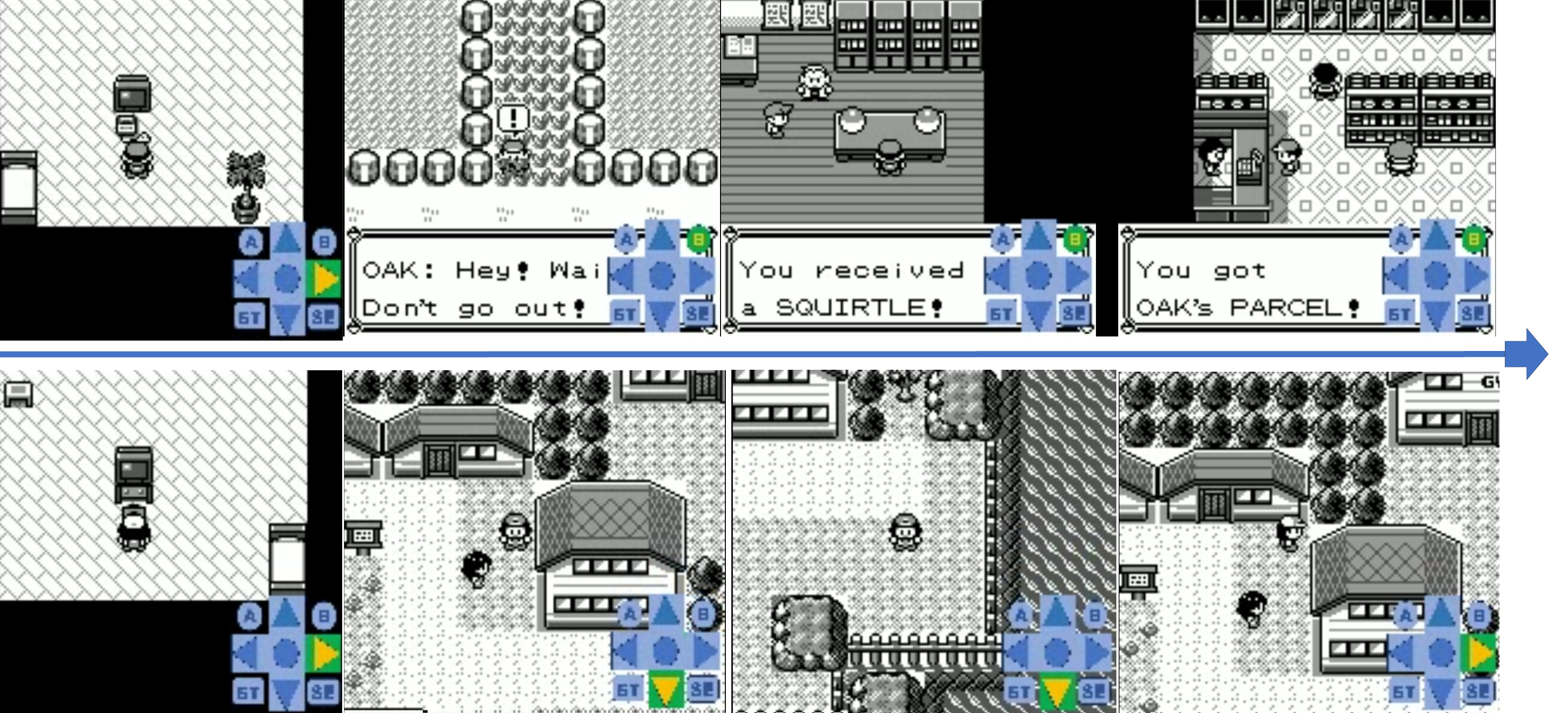}
    \caption{Substantial contamination in pretraining enables the agent to clear several milestones in Pokémon Red, while the same system remains stuck in the starter town in Pokémon Brown.}
    \label{fig:redvbrown}
\end{figure}

\noindent\textbf{Results:} We deploy this agent with Gemini-3.6-Flash and observe (Figure~\ref{fig:redvbrown}) an immediate divergence in performance on classic games v.s. our test set. In Pokémon Red, on the very first frame, the agent creates a subgoal to `Trigger the Prof. Oak cutscene by attempting to walk out of Pallet Town', evidence that pretraining contamination enables misleadingly high-quality planning. Within 100 agent loops, it navigates out of the starter house, triggers the cutscene, goes into the Pokémon Lab, selects a starter Pokémon, defeats its rival in a battle, traverses through the first route to reach the next city (Viridian City), walks into the PokéMart to trigger the cutscene where the vendor gives the player a parcel that must be delivered to Prof. Oak, and navigates back down to the Pokémon Lab in an attempt to deliver the parcel. The agent becomes stuck when its long term memory fails, and it keeps attempting to select a new starter Pokémon despite already having one. When playing Pokémon Brown, in comparison, the agent does not make any meaningful progress. It exits the starter house and walks into various other houses in the first city, speaking to the NPCs within. However, in Pokémon Brown, the way to obtain your first Pokémon is to head south to the neighbouring town, where the player meets the Pokémon professor. In the first 100 steps, the agent never attempts to explore beyond the starting town, after which it gets stuck in a loop and hence never obtains its first Pokémon. We observe similar trends on Pokémon Crystal (agent obtains the first Pokémon quickly) v.s. Pokémon Prism (agent never obtains the starter Pokémon), establishing that GameBoyWorlds-Playthrough is an ambitious target that requires a deeper level of active exploration, knowledge acquisition and deployment-time self-improvement. 

\section{Related Work}
\label{sec:related-work}
Frontier model releases show they can complete Pokémon games from start to finish, seemingly contradicting our results, which show models failing to execute simple tasks. We attribute this difference to the harness used in these works, which reads directly from game memory states and simplifies pathfinding and perception through game-specific state-space engineering. For example, the harness used in Gemini-Plays-Pokemon reads tile information like navigability directly from memory states~\citep{Breunig2025}. Our playthrough agent (Section~\ref{sec:playthrough}) supports this hypothesis, with handcrafted perception helpers endowing it with superior execution capabilities. While such decisions enable agents to progress through the game, they remove core challenges regarding perception and interaction. Our results demonstrate that without these human-crafted scaffolds, even frontier models are incapable of playing through complex, embodied video games like Pokémon. 

The most comparable benchmark to GameBoyWorlds is ARC-AGI-3~\citep{foundation2026arc}, an interactive reasoning benchmark which challenges AI agents to explore novel environments, acquire goals on the fly, and learn continuously. However, this challenge uses relatively simplistic, fully observable environments that were handcrafted for the sake of the task. As a consequence, models have already achieved a score of 99.9\% on ARC-AGI-3. The games in GameBoyWorlds introduce substantially more sophisticated and complex mechanics. Dropping a bomb in Bomberman instantly alters environmental topology by creating solid obstacles, requiring agents to compute and execute escape routes under strict real-time fuse deadlines and dynamic blast radii. Titles like Pokémon Brown require agents to manage shifting inventories, conduct non-linear exploration, and execute overarching goals spanning thousands of discrete steps. Hence, GameBoyWorlds introduce challenges that simple, puzzle-based environments like ARC-AGI-3 fundamentally cannot capture.

%Unsupervised RL + Curiosity based exploration

%https://odysseus-project.github.io/?family=land&scene=world1-level1&mode=compare

%https://cdn.patronus.ai/SpeedrunBench.pdf

%Continual Harness: Online Adaptation for Self-Improving Foundation Agents

%InSight: Self-Guided Skill Acquisition
%via Steerable VLAs

\section{Conclusion}
\label{sec:conclusion}
We introduce GameBoyWorlds, a testbed for self-improvement in embodied video games. Agents are expected to train in games and acquire capabilities without any explicit task, human demonstrations, game-specific external knowledge or task exemplars. With an execution testbed of 500 tasks, we demonstrate that frontier models lack sufficient multimodal agentic capabilities to achieve good progress. We adapt modern self-improvement approaches to the GameBoyWorlds setting and find that no method consistently raises performance across all games, showing that the benchmark remains challenging and largely unsolved. Finally, we create a playthrough challenge that tests agents in unfamiliar games, forcing them to actively acquire knowledge and skills during deployment. We hope GameBoyWorlds helps accelerate progress on self-improvement in embodied environments.

%\newpage

\bibliography{iclr2027_conference}
\bibliographystyle{iclr2027_conference}

\appendix

\newpage

\section{GameBoyWorlds Implementation}
\label{app:gameboyworlds}

\subsection{Emulation and Environment Interface}
GameBoyWorlds is implemented as a Python package built on the PyBoy emulator~\citep{pyboy}, and exposes every game through the standard Gymnasium API~\citep{towers2026gymnasium}. The package is organised into two modules. The \textit{emulation} module handles the loading of ROMs and save states, button inputs, screen rendering and the tracking of game events. The \textit{interface} module wraps the emulator into Gym-compliant environments, complete with action spaces, observation spaces and episode termination.

\noindent\textbf{Save States:} Every episode begins by loading an emulator save state, i.e., a complete snapshot of the GameBoy's memory at a particular moment of gameplay. Resetting the environment restores this snapshot exactly. Since the emulator is deterministic, the same save state and action sequence always produce the same trajectory, which makes every test task exactly reproducible. All states are created with a consistent in-game configuration (e.g. text speed and text box frame style), since several state checks rely on the appearance of on-screen text boxes.

\noindent\textbf{Actions and Observations:} Each environment step presses a single button for $5$ emulator ticks, releases it, and then waits a further $14$ ticks for the game to respond, for a total of $19$ ticks (just under a third of a second of gameplay at the GameBoy's $60$ frames per second). The observation returned to the agent is the final rendered frame, a $144\times160$ single-channel greyscale image at the native GameBoy resolution. The intermediate frames rendered during the step are retained internally and passed to the event trackers, so that brief on-screen events are not missed between observations.

\noindent\textbf{Extensibility:} The framework is not specific to the 10 games in our benchmark. Any GameBoy or GameBoy Color ROM can be added by registering the ROM, creating a default save state, and implementing a game-specific state parser. At the time of writing, the package supports several titles beyond those used in GameBoyWorlds-Execution (e.g. Harry Potter, Harvest Moon, Survival Kids and Runes of Virtue), as well as four fan-made Pokémon ROM hacks. We distribute the save states and screen captures (but not the ROMs themselves) through Hugging Face, and users must legally obtain the ROMs for each game.

\subsection{Screen Capture System}
\label{app:screencapture}
Many GameBoy titles, and ROM hacks in particular, have no public documentation of their memory layout, so reading game state directly from memory is not feasible in general. Instead, we identify game events through the screen itself. Each game's state parser defines a set of \textbf{named screen regions}, i.e., fixed rectangles on the screen at which a particular event produces a characteristic image (e.g. the edge of the start menu in the top right corner, or the dialogue box in the middle of the screen). Each region is linked to one or more \textbf{reference captures}: the pixel contents of that region at the moment an event occurs, recorded once by a human and saved to disk.

\begin{figure}[ht]
    \centering
    \includegraphics[width=\linewidth]{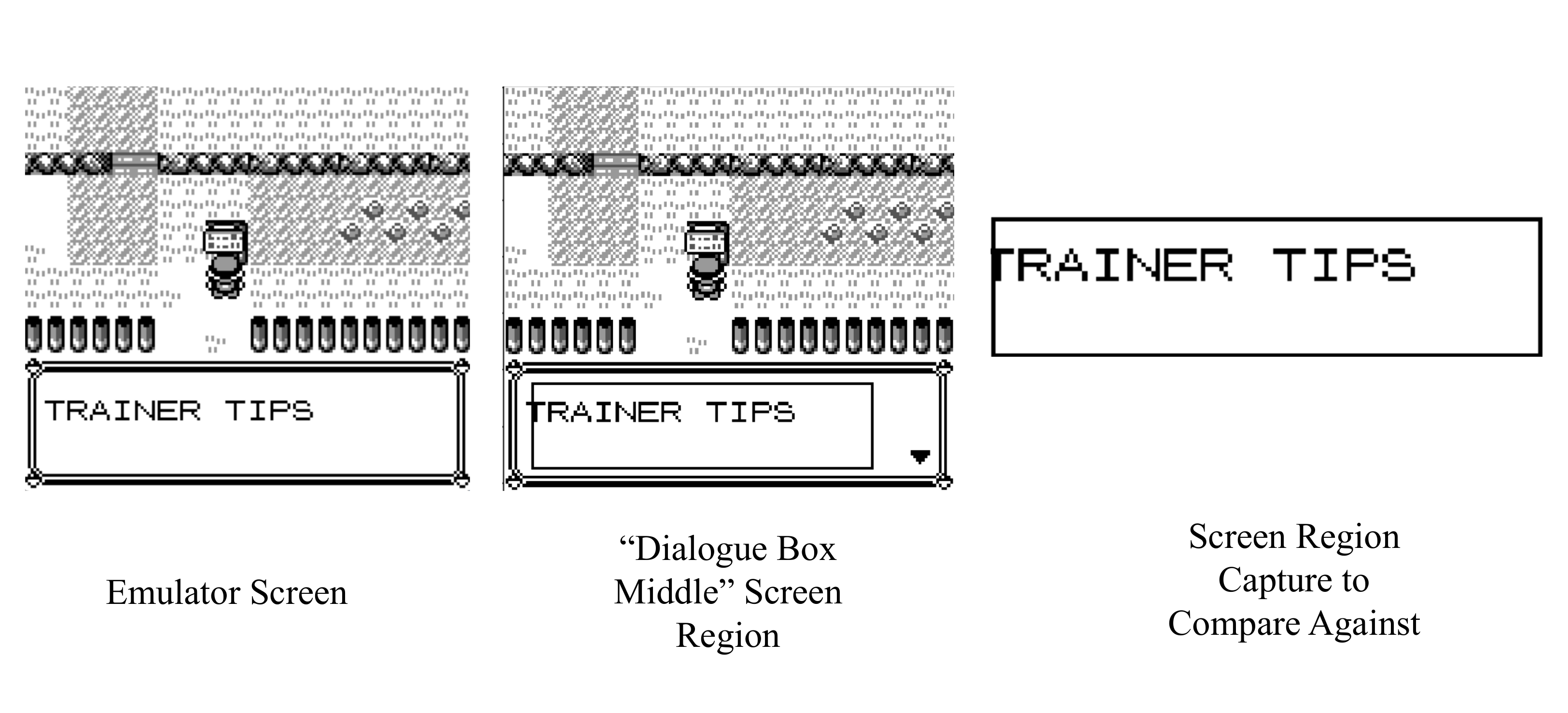}
    \caption{Capture of a screen region that can be used to compare against for the completion of the task `Read the trainers tips sign'.}
    \label{fig:capture}
\end{figure}

Given a frame, the parser crops the named region and computes the mean absolute pixel difference against the reference capture (Figure~\ref{fig:capture}). The region is considered a match if this difference is at most $0.01$. Since GameBoy frames are rendered deterministically on a small, discrete palette, the same event produces the same pixels every time it occurs, and this threshold amounts to requiring a near-exact match. In turn, spurious matches are rare, and the check is cheap enough to run on every rendered frame.

A single region can hold several reference captures, allowing one location on the screen to identify different events (e.g. the dialogue box in Pokémon Prism displays a distinct message on receiving each gym badge). Where a game's memory layout is well documented (Pokémon Red and Crystal), parsers may also read values from memory, although all tasks in GameBoyWorlds-Execution can be verified from the screen alone.

\subsection{Task Implementation}
\label{app:tasks}
Every test task is implemented by hand. For each task, one of the authors plays the game up to the moment at which the task should begin, and saves this moment as the task's initial state $s_0$. The author then completes the task from this state and, at the moment of completion, records a reference capture of a screen region that reliably and uniquely signals success (e.g. the dialogue box announcing that an item has been obtained). The completion function $\mathcal{C}_t$ is then implemented as a check against this capture: the task is considered complete as soon as any frame rendered during an episode matches the reference capture. Finally, each task is validated by loading its initial state and replaying it to confirm that the completion function fires at the moment of success, and not before.

An important consequence of this procedure is that \textbf{every task in the benchmark is achievable}. Each initial state was created from real gameplay, and each completion function was built from a screen that an author reached by playing from that very state. No task is unsolvable due to a missing item, an unreachable location or a broken completion check.

Each task is additionally annotated with a natural language description, a task category (e.g. navigation, interaction or combat) and the training game of its series (Section~\ref{sec:execution}). Termination in our environments always corresponds to success: we deliberately provide no failure signals, so that agents cannot use early termination as a source of feedback during evaluation.

\subsection{GameBoyWorlds-Playthrough Progress Markers}
\label{app:markers}
The two games in GameBoyWorlds-Playthrough follow the classic structure of the Pokémon series: the player must defeat the leaders of eight gyms, each of whom awards a badge, before challenging the Elite Four. We therefore track progress with one subgoal per gym badge, and define the final task completion as defeating the Elite Four and becoming the league champion. This gives a total of eight subgoals followed by a single completion condition for each game. For Pokémon Prism, the badges are the Pyre, Nature, Charm, Midnight, Muscle, Haze, Raucous and Naljo badges. For Pokémon Brown, they are the Marine, Hail, Sprout, Sparky, Fist, Equity, Star and Psi badges.

As with the execution tasks, the markers are implemented with the screen capture system: each badge is detected by matching the dialogue box message shown when the badge is received, and completion is detected by matching the message shown upon becoming champion. Both the playthroughs for Pokémon Prism and Brown begin directly after the game's introduction.

\subsection{Limitations}
Checking completion from the screen imposes a structural limitation: a task can only be verified if its completion produces a distinct screen that reliably appears on every successful attempt. Tasks whose success is not accompanied by such a screen cannot be verified by our system, and are excluded from the benchmark. For example, a change to an internal game variable that is never displayed, or an outcome that is only shown if the player later opens a particular menu, cannot be checked. This constraint biases the task distribution towards events with immediate visual feedback (e.g. dialogue, menu screens and map transitions). 

\subsection{Game-Specific Tasks}
\label{app:gameboyworlds_specific}
Each game contains unique mechanics and features, making the tasks implemented highly game-dependent. We describe the core differences below:

\noindent\textbf{Pokémon} (Red and Crystal): Tasks span the full breadth of the RPG. \textit{Navigation} tasks require moving through towns, buildings and caves, often under constraints imposed by the map (e.g. `Exit the cave using the exit on the bottom right', `Reach the other side of the ice maze', `Enter the burned tower'). \textit{Interaction} tasks require engaging with characters and objects in the overworld, which in turn requires facing them from the correct tile (e.g. `Speak to Bill', `Look into the mirror', `Fish until you catch a Pokemon'). \textit{Menu} tasks require operating the game's nested menus for the party, the bag, the Pokédex and the PC (e.g. `Use a Potion on Charmander', `Teach Pidgeot Toxic', `Withdraw Staryu from the box'). \textit{Combat} tasks take place in the turn-based battle system, and range from winning battles to catching wild Pokémon and exploiting type matchups (e.g. `Defeat Gym Leader Brock', `Catch Pikachu', `Defeat Golbat with a super effective attack'). Several tasks combine these categories, such as purchasing items at a shop (e.g. `Buy one Antidote', `Sell exactly 1 Psychic TM').

\noindent\textbf{Legend of Zelda} (Link's Awakening and Oracle of the Seasons): The majority of tasks are \textit{navigation} tasks in a real-time, top-down world, in which the player must find paths around obstacles, through buildings and between screens (e.g. `Go inside the shop', `Fall from the bridge', `Go inside the skeleton house'). \textit{Interaction} tasks require approaching and engaging with characters and objects (e.g. `Talk to the farmer', `Pick up the sword from the shore', `Talk to the fairy'). \textit{Inventory} tasks require managing the items assigned to the two action buttons (e.g. `Equip the sword', `Swap the shroom with the shield', `Open the inventory'). A smaller set of tasks chain navigation with interaction, such as clearing obstacles with an item to reach a new area (e.g. `Go inside the dark forest by chopping bushes', `Go to the girl's house and move the pots', `Kill all the bats in the dark forest to acquire money from the chest').

\noindent\textbf{Deja Vu} (1 and 2): As point-and-click adventures, all interaction is mediated by a cursor and a bar of verb commands, which must be selected before clicking on an object. \textit{Interact} tasks, the largest category, require applying the correct verb to an object in the scene or inventory (e.g. `Open the coat pocket', `Unlock the car door', `Buy 2 chips'). \textit{Inspect} tasks require examining an object to reveal its description (e.g. `Check the gun', `Check the timetable', `Check the dead man on the floor'). \textit{Take} tasks require moving an object from the scene into the inventory (e.g. `Take the coat from the front door', `Take gum from the trench coat', `Take the ring'). \textit{Navigation} tasks require moving between rooms through exits or the map (e.g. `Get into the cellar', `Enter the taxi', `Exit the casino'). \textit{Combat} tasks require using a weapon or tool on a target (e.g. `Hit the mugger', `Shoot the lock', `Use the knife to open the door'), and \textit{Use} tasks require equipping items onto the player (e.g. `Put on the trench coat').

\noindent\textbf{Sword of Hope} (1 and 2): The player acts almost exclusively through a grid of commands, so the majority of tasks test precise \textit{menu} control. Menu navigation tasks require moving the cursor to a specific command or entry (e.g. `Navigate the cursor onto LOOK in the command grid', `In the weapons shop, move the cursor to the third weapon', `In the Magic submenu, move the cursor to the Firebal spell'). Menu reasoning tasks require entering a submenu and then backing out without committing to an action (e.g. `Open the Magic submenu and cancel back to exploration without casting', `Open the shop buy menu and exit without purchasing', `Open the Teleport destination list and cancel back to exploration'). \textit{Interaction} tasks chain commands to affect the world (e.g. `Buy Wheat from a shop', `Look at surroundings to reveal a previously hidden path', `Use a key or quest item to unlock passage to a new area'), and \textit{combat} tasks take place in the menu-driven battle system (e.g. `Finish a battle encounter without a game over', `Escape from a battle using the Escape command', `Defeat a boss using offensive magic spells').

\noindent\textbf{Bomberman} (Quest, Pocket and Max): Tasks centre on the series' core mechanic of placing bombs that detonate after a delay. \textit{Combat} tasks require defeating enemies with bombs while escaping the blast radius (e.g. `Defeat the Skull Enemy using a bomb', `Kill the Forest Stage Boss and take the exit', `Activate the switch using a bomb'). \textit{Navigation} tasks require finding a route through grid-like arenas, often by destroying blocks along the way (e.g. `Stage 3: Go to the stage exit', `Forest Area 2: Navigate to find and take the exit', `Navigate from the Forest Entrance to enter the Ancient Ruins'). \textit{Interaction} tasks require collecting power-ups and speaking to characters (e.g. `Pick up a Fire Up', `Talk to the Old Man', `Use the Power Glove to activate the switch'), and \textit{menu} tasks require operating the game's pause and equipment screens (e.g. `Open the bomb select screen', `Open the Charabom select screen', `Open the pause menu').

\noindent\textbf{Harry Potter} (Philosopher's Stone and Chamber of Secrets): An RPG in which exploration is interleaved with a turn-based, card-driven battle system. \textit{Navigation} tasks require moving between the rooms and shops of the wizarding world (e.g. `Enter Flourish and Blotts bookshop', `Enter Ron's room in the Burrow', `Enter the apothecary'). Most tasks combine navigation with \textit{interaction}, requiring the player to find and engage with a character or object (e.g. `Talk to Hagrid inside Gringotts', `Find Dobby in the bedroom', `Board the flying car'). \textit{Battle} tasks require selecting actions and card attacks against enemies in specific positions (e.g. `Use the Broom Doom card attack', `Fight a gnome in the top left position', `Win the duel'), and \textit{menu} tasks require operating shop and equipment menus (e.g. `Buy 5 chocolate frogs and confirm you have them in inventory', `Equip the pointed hat and then the plain work robe', `Select a card deck before leaving').

\noindent\textbf{Harvest Moon} (1, 2 and 3): A farming RPG in which the player manages a farm and its livestock. Almost every task combines \textit{navigation} with \textit{interaction}, as tools, animals and signs must be reached before they can be used (e.g. `Pick up the water can', `Place the egg in the incubator to hatch the chick', `Harvest only the center eggplant of the 3x3 crop field without disturbing the others'). Pure \textit{navigation} tasks require reaching buildings and areas of the world (e.g. `Enter the cow barn', `Enter the hospital', `Enter the forest'). \textit{Menu} tasks require completing transactions through shop menus, often under exact constraints on what is bought (e.g. `Buy a cow and name it ABBY', `Buy and only buy a fodder set at the farmer's union', `Get an estimate for building a bridge').

\noindent\textbf{Runes of Virtue} (1 and 2): An action RPG of the \textit{Ultima} franchise, set largely in multi-floor underground caverns. \textit{Navigation} tasks require descending through floors of a cavern and finding its entrances and exits (e.g. `Enter the Cavern of Cowardice', `Cavern of Dishonour: Enter floor 4', `Exit the Cavern of Dishonour'). \textit{Dialogue} tasks require locating and speaking to specific characters (e.g. `Talk to the king for the first time', `Talk to Nystul', `Talk to the blacksmith and fail to buy a shield'). \textit{Interaction} tasks require obtaining and using items (e.g. `Cavern of Cowardice: Open the chest on floor 3', `Obtain the Rune of Honour', `Give cheese to Sherry'). The remaining tasks test the inventory menu (`Open the inventory menu') and survival mechanics (`Die').

\noindent\textbf{Survival Kids} (1 and 2): A survival RPG in which the player is stranded on a deserted island and must manage hunger, thirst and time of day. \textit{Survival} tasks, often carried out through the inventory menu, require crafting and consuming resources (e.g. `Light a fire with kindling', `Cook the meat over the fire', `Recover hunger by eating fruit'). \textit{Interaction} tasks require gathering materials from the environment (e.g. `Pick up the tree bark', `Find the hidden bag', `Cut grass before the sharp stone'). \textit{Menu} tasks require selecting, equipping and discarding items (e.g. `Equip the knife', `Select Take from the clam menu', `Drop trash from the inventory'), and \textit{navigation} tasks require exploring the island as new paths are cleared (e.g. `Clear the first blocked path', `Find the river', `Find the second new path').

\subsection{Evaluated Subset}
In total, the games described above constitute $1000$ tasks across $20$ games ($50$ tasks per game). In this work, we compute full benchmark results only on the $10$ games from the five series described in Section~\ref{sec:execution} (Pokémon, Legend of Zelda, Deja Vu, Sword of Hope and Bomberman, excluding Bomberman Max), for a total of $500$ test tasks. The remaining $500$ tasks, spanning Bomberman Max, Harry Potter, Harvest Moon, Runes of Virtue and Survival Kids, are fully implemented in GameBoyWorlds, and we leave their evaluation to future work.

\section{Zeroshot Agent}
\label{app:zeroshot}

This section details the agent framework used to benchmark frontier models in Section~\ref{sec:evaluation}. The framework consists of two components: a \textbf{supervisor}, which plans, judges progress and writes guidance, and an \textbf{executor}, which interacts with the environment one button press at a time. Both components are driven by the same underlying VLM, and every prompt is game-agnostic: the only game-specific information provided is the name of the game.
\subsection{Episode Structure}
Each test episode is allotted a budget of $75$ steps. A step is consumed by every executor response, including responses that could not be parsed or that named an invalid action, so that the budget bounds the agent's interaction identically across all framework variants. Supervisor calls do not consume steps. An episode ends as soon as the environment signals task completion, or when the budget is exhausted. Success is always determined by the environment's completion function (Appendix~\ref{app:tasks}) and never by the agent's own judgement.

\subsection{Supervisor}
\label{app:zeroshot_supervisor}

\noindent\textbf{Subgoal Planning:} Given the task and the initial observation, the supervisor writes a plan, i.e., an ordered sequence of subgoals. Each subgoal is handed to the executor in isolation, without sight of the other subgoals. Hence, the planning prompt requires every subgoal to stand on its own, to refer only to objects that are visible on the screen, to avoid naming buttons, and to end in a visually checkable termination condition. If no plan can be parsed from the response, the episode proceeds with the task itself as the only subgoal. The prompt is given in Prompt~\ref{prompt:subgoal_planning}.

\noindent\textbf{Attempts:} The supervisor works through the plan in order. For each subgoal, it deploys the executor for an \textit{attempt} of at most $5$ steps, with the subgoal given as the executor's task. Intermediate subgoals have no environment signal, so the executor is additionally permitted to end an attempt early by declaring the subgoal complete (see \textit{Completion Check} below). The final subgoal is treated differently: the executor is given the original task, with the final subgoal as its hint, and may not declare completion, since only the environment can clear the task. Each intermediate subgoal is allowed $3$ failed attempts before the supervisor moves on to the next subgoal. The final subgoal is retried until the budget is exhausted.

\noindent\textbf{Subgoal Judgement:} After each attempt at an intermediate subgoal, the supervisor decides whether the subgoal was accomplished. The frames of the attempt are split into windows of at most $8$ frames, and each window is summarised separately. The supervisor then consolidates these summaries and inspects the final screen to produce a verdict. The executor's reason for stopping is shown as a claim to be verified, as an executor that declared itself finished may be mistaken. The prompts are given in Prompts~\ref{prompt:subgoal_judgement_window_summary} and~\ref{prompt:subgoal_judgement_verdict}.

\noindent\textbf{Regression Check:} A failed attempt may undo progress on a subgoal that was already cleared (e.g. by closing a menu that an earlier subgoal opened). The verdict above only considers the current subgoal, and cannot detect this. Hence, after every failed attempt, the supervisor compares the screen at the moment the most recent subgoal was judged complete against the current screen, and determines whether that progress has been lost. If so, the loss is reported to both the error correction and self-revision stages. The prompt is given in Prompt~\ref{prompt:regression_check}.

\noindent\textbf{Self-Revision:} After every failed attempt, the supervisor considers whether the failure is due to the executor fumbling a sound plan, or due to the plan itself being flawed. The prompt is deliberately strict: repeated failure alone is not accepted as evidence of a flawed plan, and the supervisor must name something visible on the screen that the plan is incompatible with. If the plan is judged flawed, the current subgoal and all subsequent subgoals are replaced, while subgoals that have already been cleared are left untouched. The replacement subgoals are subject to the same requirements as the original plan, and restart with a fresh attempt count. The plan may be revised at most $2$ times per episode. The prompt is given in Prompt~\ref{prompt:self_revision}.

\noindent\textbf{Error Correction:} If the plan is not revised and attempts remain, the supervisor writes a hint for the next attempt. The hint prompt is shown the window summaries of the failed attempt, the action taken at each step alongside the executor's stated reasoning for it, the reason the attempt was judged incomplete, any detected regression, and the previous hint. In contrast to the plan, the hint is permitted to name buttons, so that the supervisor can correct the executor's misunderstanding of the controls directly. However, it may not prescribe button counts or sequences, as a sequence written from one screen quickly becomes invalid once the executor begins to act. If no hint can be parsed from the response, the previous hint is retained. The prompt is given in Prompt~\ref{prompt:error_correction}.

The planning, self-revision and error correction prompts contain a slot for knowledge recorded from past playthroughs. In the zeroshot agent, this slot is always filled with \texttt{(nothing recorded)}. The same prompts are used, with the slot filled, by the insight-guided agent in Section~\ref{app:insight}.

\subsection{Executor}
\label{app:zeroshot_executor}

\noindent\textbf{Single Actions:} At every step, the executor is shown the current screen, its task, the hint from the supervisor (if any), the list of available actions and its history (see \textit{Visual History} below). It reasons about the best next action and outputs a single action. The hint is presented as privileged guidance: the executor is instructed to act on it without referring to it explicitly in its reasoning. The hint block also notes how many actions have already been taken in the current attempt, and that earlier attempts may have already changed the game state. If the executor produces $4$ consecutive responses that cannot be parsed or that name an invalid action, the attempt is terminated. Otherwise, the executor is shown an error message explaining the problem on its next step. The prompts are given in Prompts~\ref{prompt:executor_step} and~\ref{prompt:executor_step_hint_block}.

\noindent\textbf{Visual History:} After every action, the executor is shown the screens before and after the action, and describes in a single sentence what changed as a result. The descriptions of the $5$ most recent actions are inserted into the \texttt{[CONTEXT\_SECTION]} slot of the step prompt. If any of these descriptions report that nothing changed, the executor is additionally warned that it may be stuck in a loop. The prompts are given in Prompts~\ref{prompt:visual_history_change_description} and~\ref{prompt:visual_history_context_section}.

\noindent\textbf{Completion Check:} When the executor is attempting an intermediate subgoal, a completion check follows every action. The check is shown the screens before and after the action, the action and the reasoning behind it, and the $8$ most recent actions, and must decide whether the subgoal is now fully complete. A positive verdict ends the attempt and passes control back to the supervisor for judgement. The prompt is given in Prompt~\ref{prompt:completion_check}.

\subsection{Alternative Frameworks}
\label{app:zeroshot_alternatives}
The framework above was selected after comparing alternatives along three axes. All alternatives share the step budget, prompts and parameters of the framework above, and differ only in the component under study.

\noindent\textbf{Supervisor:} We consider three supervisors of increasing complexity. \textit{No Supervisor} deploys a single executor on the task for the entire budget, with no hints. \textit{Error Correction} runs the executor in attempts on the task itself and writes a hint after every failure, but does not plan, and hence has no subgoal judgement, regression check or self-revision. \textit{Subgoal Planning} is the full supervisor described in Section~\ref{app:zeroshot_supervisor}.

\noindent\textbf{Action Selection:} We consider three ways for the executor to decide on an action. \textit{Single} is the scheme described above. \textit{Scored} asks the executor to assign every available action a usefulness score from $1$ to $5$ with a one-sentence justification, and takes the highest scoring action (ties are broken by the order of the action list). It replaces the final lines of the step prompt with Prompt~\ref{prompt:executor_step_scored_variant}.

\textit{Sequence} asks the executor to commit to a sequence of $1$ to $5$ actions, which are executed in order. It replaces the final lines of the step prompt with Prompt~\ref{prompt:executor_step_sequence_variant}. With this scheme, the completion check runs only once the full sequence has been executed.

\noindent\textbf{History:} We consider three forms of history for the executor. \textit{None} leaves the \texttt{[CONTEXT\_SECTION]} slot empty. \textit{Actions} lists the $5$ most recent actions, each tagged with whether the screen changed as a result, with the same loop warning as above if any action produced no change. The action history is formatted as in Prompt~\ref{prompt:executor_step_action_history_variant}. \textit{Visual} is the scheme described above.

\noindent\textbf{Results:} We evaluate all $27$ combinations with Gemma-4-31B on three games: Deja Vu 2, Harvest Moon 3 and Legend of Zelda: Oracle of the Seasons. Since each combination is evaluated once, and a single task corresponds to only $2$ to $3$ percentage points, we summarise the results as paired comparisons: two components are compared under every setting of the remaining components, and we count how often each achieves the higher task completion rate.

Table~\ref{tab:zeroshot_ablation} compares visual history against no history, across all $9$ combinations of supervisor and action selection. Visual history achieves a higher completion rate in $19$ of $27$ comparisons and a lower one in $6$. Averaged over all combinations and games, task completion is 21.8\% with no history, 23.3\% with action history and 25.4\% with visual history. However, most individual differences amount to between $1$ and $4$ tasks.

\begin{table}[h]
\centering
\small
\caption{Paired comparison of visual history against no history with Gemma-4-31B. Each game contributes $9$ comparisons, one for every combination of supervisor and action selection. A win indicates that visual history achieves a strictly higher task completion rate.}
\label{tab:zeroshot_ablation}
\begin{tabular}{lccc}
\toprule
Game & Wins & Ties & Losses \\
\midrule
Deja Vu 2         & 6  & 2 & 1 \\
Harvest Moon 3    & 6  & 0 & 3 \\
Oracle of Seasons & 7  & 0 & 2 \\
\midrule
Total             & 19 & 2 & 6 \\
\bottomrule
\end{tabular}
\end{table}

Table~\ref{tab:zeroshot_ablation_subgoal} compares the Subgoal Planning supervisor against each alternative supervisor, across all $9$ combinations of action selection and history. Subgoal Planning outperforms the unsupervised executor in $19$ of $27$ comparisons, winning every comparison on Oracle of the Seasons and all but one on Harvest Moon 3. Against Error Correction, the comparison is evenly split, with $10$ wins, $8$ ties and $9$ losses. Deja Vu 2 remains challenging under every combination, with no framework exceeding 14.7\%.

\begin{table}[h]
\centering
\small
\caption{Paired comparison of the Subgoal Planning supervisor against each alternative supervisor with Gemma-4-31B. Each game contributes $9$ comparisons, one for every combination of action selection and history. A win indicates that Subgoal Planning achieves a strictly higher task completion rate.}
\label{tab:zeroshot_ablation_subgoal}
\begin{tabular}{lcccccc}
\toprule
 & \multicolumn{3}{c}{vs.\ No Supervisor} & \multicolumn{3}{c}{vs.\ Error Correction} \\
\cmidrule(lr){2-4} \cmidrule(lr){5-7}
Game & Wins & Ties & Losses & Wins & Ties & Losses \\
\midrule
Deja Vu 2         & 2  & 2 & 5 & 1  & 6 & 2 \\
Harvest Moon 3    & 8  & 1 & 0 & 4  & 2 & 3 \\
Oracle of Seasons & 9  & 0 & 0 & 5  & 0 & 4 \\
\midrule
Total             & 19 & 3 & 5 & 10 & 8 & 9 \\
\bottomrule
\end{tabular}
\end{table}

Our overall claim is not the the framework chosen in Section~\ref{sec:evaluation} is fundamentally superior to other choices, but rather that other choices produce comparable or worse results. 

\subsection{Example Trajectory}
\label{app:zeroshot_examples}
\begin{figure}
    \centering
    \includegraphics[width=\linewidth]{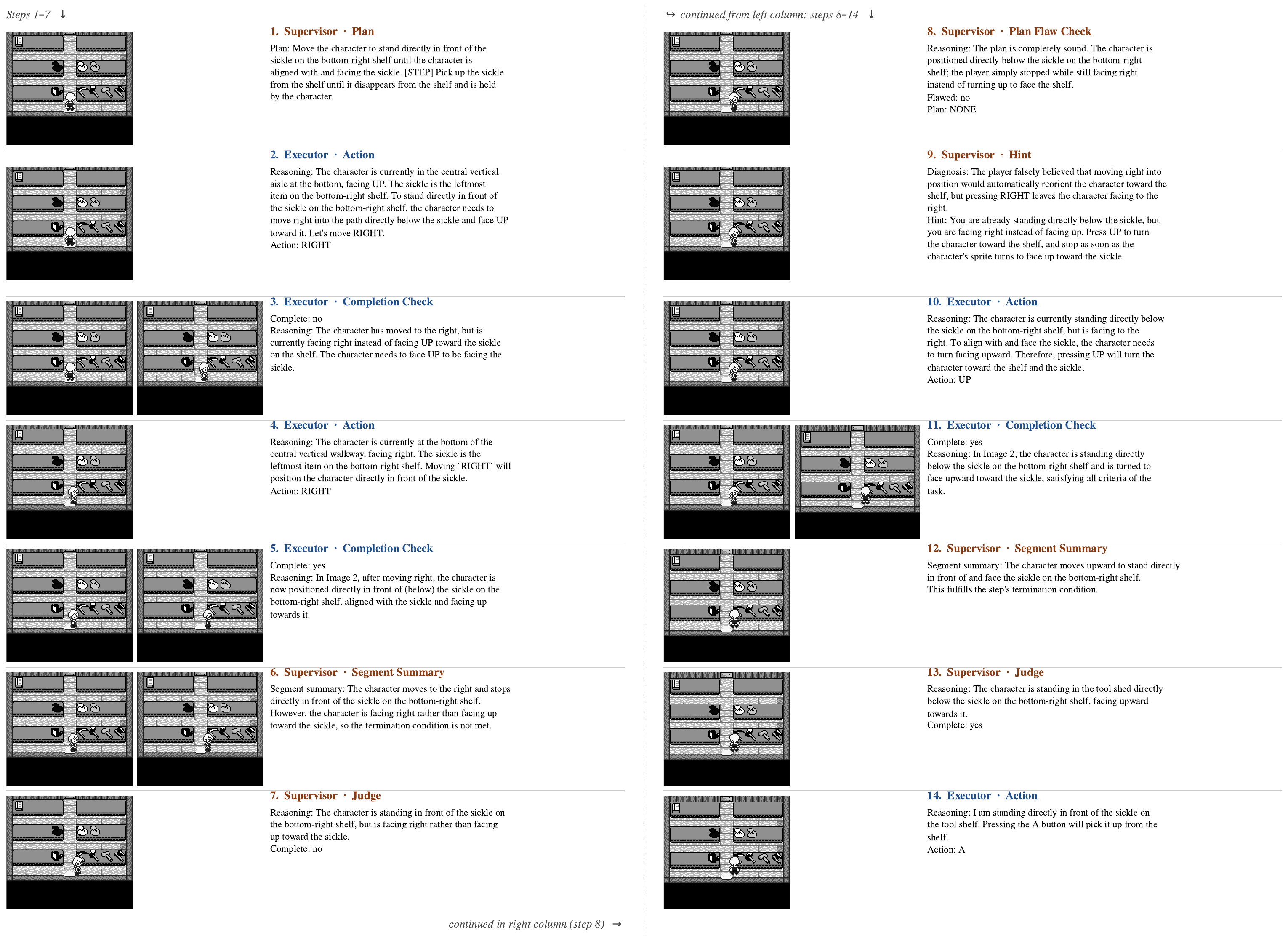}
    \caption{Example trajectory of agent successfully picking up a sickle in Harvest Moon 1. The episode terminates after the final frame. For brevity, the `Visual Difference' step, where the VLM summarizes what changed between frames, has been omitted.}
    \label{fig:trajectory}
\end{figure}

Refer to Figure~\ref{fig:trajectory}, which shows an agent completing the task `Pick up a sickle'. 

\section{Self Improvement Method Implementations}
\label{app:selfimprove}

\subsection{Curiosity-Based Exploration}
\label{app:curiosity}
Curiosity-based exploration is the source of experience for two of the methods we study: it collects the replay buffers on which world models are trained (Section~\ref{app:wm}), and it surfaces the trajectories from which insight documents are distilled (Section~\ref{app:insight}). We describe the full exploration setup here, and state the configuration used by each method in its own section. 

\noindent\textbf{Agent:} The curiosity agent is a PPO~\citep{schulman2017proximal} policy acting in the low-level action space of button presses. It observes a stack of the $2$ most recent frames, and plays episodes of at most $30$ steps from a training state. The environment provides no extrinsic reward, so the agent is trained on the intrinsic reward alone.

\noindent\textbf{Observation Novelty Reward:} Each frame is embedded with a fixed random patch projection: the frame is divided into non-overlapping $8\times8$ patches, each patch is mapped to $2$ dimensions by a random linear projection with weights drawn from a standard normal distribution, and the concatenated $720$-dimensional vector is normalised to unit length. The projection is generated from a fixed seed, so that the same projection is used by every run. The agent maintains a buffer of the embeddings of frames it has observed. Given the latest frame with embedding $z$ and a buffer $\{z_i\}_{i=1}^{N}$, the observation reward is
\begin{equation*}
r_{\text{obs}} = 1 - \max_{i\in[1, N]} \text{sim}(z, z_i),
\end{equation*}
where $\text{sim}$ is the cosine similarity. The frame is then added to the buffer, unless an embedding already in the buffer matches it to within $10^{-3}$ in every dimension. Should the buffer exceed $10{,}000$ embeddings, it is compressed to $5{,}000$ embeddings by $k$-means clustering, retaining only the cluster centres.

To clarify our motivation behind this choice of `difference', consider frame observations of the GameBoy screen $o_1, o_2, o_3\in\mathbb{R}^{H\times W}$ with latent representations $z_i\in\mathbb{R}^d$. We then refer to the Johnson-Lindenstrauss lemma~\citep{venkatasubramanian2011johnson}, which shows that if $\|o_1-o_2\| < \|o_2-o_3\|$, then with high probability, $\|z_1-z_2\| < \|z_2-z_3\|$. This result suggests we should use the distance between the latents as a proxy for difference. However, we observed that building a reward signal out of distance led to high variance and poor policy convergence. Instead, we consider the cosine similarity between latents.

\noindent\textbf{Text Novelty Reward:} The observation reward treats a new line of dialogue in an already familiar dialogue box as a near-duplicate frame, since the text occupies a small fraction of the screen. However, dialogue and menu text frequently carry the information that is most relevant to a game's mechanics. To reward the discovery of new text, each game's state parser in GameBoyWorlds defines text regions of the screen (e.g. the dialogue box and menu area for Deja Vu and Harvest Moon, the dialogue box and battle move list for Pokémon, and the full screen for Sword of Hope), and reports the contents of each region whenever text is displayed within it. Each captured region is split into $8$ vertical strips, and a separate buffer of raw strips is maintained for every region. The text reward for a region is the fraction of its strips that do not match any strip in that region's buffer to within $10^{-3}$ (or $1$ if the region has never been observed), and unseen strips are added to the buffer. The text reward $r_{\text{text}}$ is the maximum over all regions currently displaying text, and $0$ if no text is displayed.

\noindent\textbf{Combined Reward:} The two rewards are combined with a text weight $\alpha\in[0, 1]$:
\begin{equation*}
r = \alpha\, r_{\text{text}} + (1-\alpha)\, r_{\text{obs}}.
\end{equation*}
With $\alpha=0$, the agent is rewarded for visual novelty alone. With larger $\alpha$, the agent is additionally drawn towards triggering new dialogue and opening new menus.

\noindent\textbf{Buffer Reset:} Both buffers are reset at the start of every episode, so that the reward schedule remains stationary across episodes. In practice, this leads each agent to converge on a single high-reward trajectory.

\noindent\textbf{Iterative Rounds:} To discover a range of behaviours from the same training state, exploration proceeds in sequential rounds. At the end of each round, the buffers of the round are saved and merged with those of all previous rounds, and every episode of the next round begins with its buffers pre-populated with this merged set rather than empty. Hence, an agent in a later round is only rewarded for observations and text that no earlier round surfaced. When several values of $\alpha$ are used, a separate agent is trained for each value within every round, and all of them contribute to the round's buffers.

\noindent\textbf{Hyperparameter Sweep:} Optionally, each round is run as a sweep over the discount factor $\gamma\in\{0.99, 0.995, 0.999\}$. Every policy in the sweep is evaluated after training, the $6$ policies with the highest evaluation return are retained, and only the replay buffers of the retained policies are kept for later use.

\noindent\textbf{Trajectory Extraction:} Every transition of every agent is stored in a replay buffer. To surface `interesting' subtrajectories, the rewards within a buffer are standardised into $z$-scores, and every step whose $z$-score exceeds $2.5$ is marked as the endpoint of a high-reward subtrajectory, which is extracted along with the steps of the episode that led to it.

\begin{figure}
    \centering
    \includegraphics[width=\linewidth]{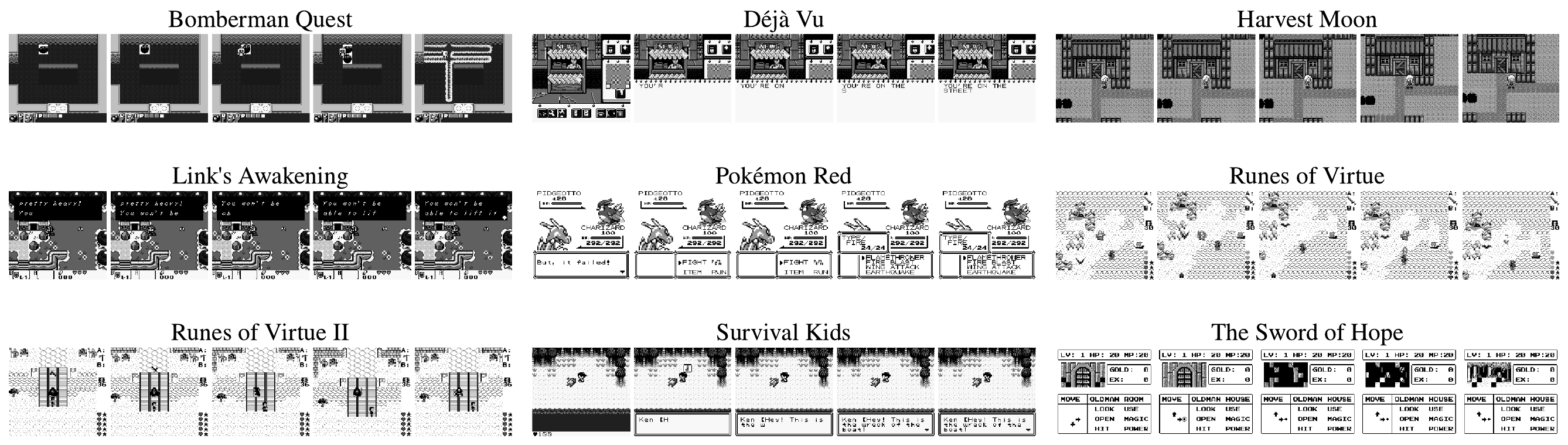}
    \caption{Random sample of high-reward trajectories from curiosity based exploration. Bomberman agents uncover bomb placement, Pokemon agents uncover specific attacks, Survival Kids agents reveal specific dialogue responses when attempting interaction, etc.}
    \label{fig:curexample}
\end{figure}

The result is a pipeline which filters out events that, in practice, often surfaces game-specific mechanics, dialogue and information rich moments of gameplay (Figure~\ref{fig:curexample}). 

\begin{tcolorbox}[breakable, title=Curiosity-Based Exploration: Fixed Hyperparameters]
\small
\begin{tabular}{@{}ll@{}}
\multicolumn{2}{@{}l}{\textit{Environment}} \\
Action space & Low-level button presses \\
Frame stack & 2 \\
Episode length & 30 steps \\
Extrinsic reward & None \\
\multicolumn{2}{@{}l}{\textit{PPO}} \\
Timesteps per agent & 100,000 \\
Parallel environments & 1 \\
Rollout length & 128 \\
Learning rate & $2.5\times10^{-4}$ (linearly annealed) \\
Discount $\gamma$ & 0.99 (unless swept) \\
GAE $\lambda$ & 0.95 \\
Minibatches & 4 \\
Update epochs & 4 \\
Clipping coefficient & 0.1 (policy and value) \\
Entropy coefficient & 0.01 \\
Value loss coefficient & 0.5 \\
Max gradient norm & 0.5 \\
Advantage normalisation & Yes \\
Seed & 1 \\
\multicolumn{2}{@{}l}{\textit{Observation Novelty}} \\
Embedder & Random patch projection \\
Patch size & $8\times8$ \\
Projection dimension per patch & 2 (720 per frame) \\
Similarity & Cosine \\
Duplicate tolerance & $10^{-3}$ \\
Maximum buffer size & 10,000 ($k$-means to 5,000) \\
\multicolumn{2}{@{}l}{\textit{Text Novelty}} \\
Strips per text region & 8 \\
Duplicate tolerance & $10^{-3}$ \\
\multicolumn{2}{@{}l}{\textit{Schedule}} \\
Buffer reset & Every episode \\
Sweep values of $\gamma$ & $\{0.99, 0.995, 0.999\}$ \\
Policies retained per sweep & 6 \\
\multicolumn{2}{@{}l}{\textit{Extraction}} \\
High-reward $z$-score threshold & 2.5 \\
\end{tabular}
\end{tcolorbox}

\subsubsection{Configuration for Insight Documents}
\label{app:curiosity_docs}
For insight document creation, exploration is run from every training state of every training game for $4$ rounds (an initial round followed by $3$ iterative rounds). Within every round, two agents are trained, with text weights $\alpha=0$ and $\alpha=0.75$, for a total of $8$ agents per training state. High-reward subtrajectories are then grouped by the similarity of their final frames, retaining only those with a $z$-score of at least $6.0$, as described in Section~\ref{app:synthetic}. We sweep $\gamma$ values of $0.95, 0.99, 0.995$ and 3 random seeds. 

\subsection{World Modelling}
\label{app:wm}

Our world modelling pipeline consists of three stages: (1) collecting experience in the training environments, (2) training an observation embedder that maps frames into a compact latent space, and (3) training a world model that predicts the latent of the next frame given the latents of recent frames and an action. The trained world model is then deployed within the executor. All stages use only data generated by the agent's own interaction with the training games, with no human annotation.

\subsubsection{Experience Collection}
For every training game and every training state declared for that game, we collect two replay buffers. The first is collected by a \textbf{random agent}, which selects button presses uniformly at random, and is intended to provide the observation embedder with broad coverage of the game's screens. The second is collected by the \textbf{curiosity agent} of Section~\ref{app:curiosity}, which surfaces states that random play is unlikely to reach. For world model training, we instantiate curiosity-based exploration in its simplest form: the text weight is set to $\alpha=0$, so that the agent is rewarded for visual novelty alone, and exploration is run for $2$ rounds (an initial round and a single iterative round, which is rewarded only for observations that the first round did not surface), with no hyperparameter sweep. Unlike insight document creation, no subtrajectories are extracted: every transition from both agents is stored and used for training. Both agents share the environment configuration of Section~\ref{app:curiosity}, and the settings specific to world model data collection are given in Figure~\ref{fig:wm_collection_hparams}.

\begin{figure}[h]
\centering
\small
\begin{tabular}{ll}
\toprule
Setting & Value \\
\midrule
\multicolumn{2}{l}{\textit{Random Agent}} \\
Timesteps per training state & 10,000 \\
\midrule
\multicolumn{2}{l}{\textit{Curiosity Agent}} \\
Text weight $\alpha$ & 0 (observation novelty only) \\
Rounds & 2 \\
Agents per round & 1 \\
Hyperparameter sweep & None \\
Timesteps per training state and round & 100,000 \\
Transitions kept & All \\
\midrule
\multicolumn{2}{l}{\textit{Other}} \\
Fixed hyperparameters & As in Section~\ref{app:curiosity} \\
\bottomrule
\end{tabular}
\caption{Configuration of experience collection for world model training.}
\label{fig:wm_collection_hparams}
\end{figure}

\subsubsection{Observation Embedder}
Predicting raw $144\times160$ frames directly is expensive and dominated by pixel-level detail. Instead, we first train an observation embedder: a patch-based convolutional autoencoder that encodes a frame into a compact latent, and decodes a latent back into a frame. Each frame is divided into non-overlapping $8\times8$ patches, giving $18\times20 = 360$ patches. Every patch is normalised (with non-learned batch normalisation) and independently encoded by a shared convolutional network into a $4$-dimensional vector, which is unit-normalised. The frame latent is the concatenation of all patch vectors, re-normalised to unit length, for a total dimension of $1440$. A shared decoder maps each patch vector back to an $8\times8$ patch, and the patches are reassembled into a frame. The architecture is given in Figure~\ref{fig:wm_embedder_hparams}.

The embedder is trained to reconstruct the normalised patches under a mean squared error loss, on the most recent frame of every observation stored in the replay buffers of a game (from both the random and curiosity agents). A random $10\%$ of the frames is held out for validation, and training stops once the validation loss fails to improve for $3$ consecutive epochs.

\begin{figure}[h]
\centering
\small
\begin{tabular}{ll}
\toprule
Component & Specification \\
\midrule
\multicolumn{2}{l}{\textit{Architecture}} \\
Patch size & $8\times8$ (360 patches per frame) \\
Patch normalisation & BatchNorm (non-affine) \\
Encoder & Conv $1\times1$ ($1\to2$), ReLU \\
 & Conv $2\times2$ ($2\to4$), ReLU \\
 & Conv $2\times2$ ($4\to4$), ReLU \\
 & Flatten, Linear ($144\to4$), Sigmoid, LayerNorm \\
Patch latent & 4 dimensions, L2-normalised \\
Frame latent & 1440 dimensions, L2-normalised \\
Decoder & Linear ($4\to144$) \\
 & ConvTranspose $2\times2$ ($4\to4$), ReLU \\
 & ConvTranspose $2\times2$ ($4\to2$), ReLU \\
 & ConvTranspose $1\times1$ ($2\to1$) \\
\midrule
\multicolumn{2}{l}{\textit{Training}} \\
Objective & MSE reconstruction of normalised patches \\
Optimiser & Adam \\
Learning rate & $10^{-3}$ \\
Weight decay & $10^{-5}$ \\
Batch size & 256 \\
Train / validation split & 90\% / 10\% \\
Early stopping patience & 3 epochs \\
Seed & 1 \\
\bottomrule
\end{tabular}
\caption{Architecture and training hyperparameters of the observation embedder.}
\label{fig:wm_embedder_hparams}
\end{figure}

\subsubsection{World Model}
The world model $\mathcal{W}$ operates entirely in the latent space of the frozen observation embedder. Given the latents of the $2$ most recent frames and a discrete action, it predicts the latent of the next frame. The stacked frame latents are projected by a linear layer, the action is mapped to a learned embedding, and the two are concatenated and passed through a small MLP whose output is unit-normalised. The world model is trained on every transition in the replay buffers, excluding the final step of each episode, to minimise the mean squared error between the predicted latent and the embedder's latent of the true next frame. As with the embedder, $10\%$ of transitions are held out for validation, and training stops once the validation loss fails to improve for $3$ consecutive epochs. The architecture and hyperparameters are given in Figure~\ref{fig:wm_hparams}.

\begin{figure}[h]
\centering
\small
\begin{tabular}{ll}
\toprule
Component & Specification \\
\midrule
\multicolumn{2}{l}{\textit{Architecture}} \\
Input & Latents of 2 stacked frames ($2\times1440$) and action index \\
Observation encoder & Linear ($2880\to512$) \\
Action encoder & Embedding ($|\mathcal{A}|\to512$) \\
Predictor & Linear ($1024\to512$), ReLU, LayerNorm \\
 & Linear ($512\to1440$) \\
Output & Next frame latent, L2-normalised \\
Observation embedder & Frozen (Figure~\ref{fig:wm_embedder_hparams}) \\
\midrule
\multicolumn{2}{l}{\textit{Training}} \\
Objective & MSE between predicted and true next frame latent \\
Optimiser & Adam \\
Learning rate & $10^{-3}$ \\
Weight decay & $10^{-5}$ \\
Batch size & 256 \\
Train / validation split & 90\% / 10\% \\
Early stopping patience & 3 epochs \\
Seed & 1 \\
\bottomrule
\end{tabular}
\caption{Architecture and training hyperparameters of the world model.}
\label{fig:wm_hparams}
\end{figure}

\subsubsection{World Model Executor}
The world model executor replaces the executor of Section~\ref{app:zeroshot_executor}, and is deployed under the same supervisor and step budget. At every step, the executor maintains a stack of the $2$ most recent frames (repeating the first frame at the start of an attempt), and embeds it with the observation embedder. For every action in the action space, the world model predicts the latent of the resulting frame, which is decoded and de-normalised into a predicted frame. The VLM is then shown the current frame followed by one predicted frame per action, and selects the predicted frame that most directly advances its task. The action associated with the selected frame is executed. The prompt is given in Prompt~\ref{prompt:world_model_executor_step}.

\subsection{Autonomous Skill Discovery}
\label{app:skill}

Our implementation of the Proposer-Agent-Evaluator framework consists of five stages: (1) proposing tasks from training states, (2) attempting each proposed task and judging success, (3) inferring guidance from successful attempts, (4) practising each task from perturbed starting states, and (5) cleaning the practice data and fine-tuning a VLA on it. Every stage uses Gemma-4-31B, and all attempts and practice episodes use the single action executor with visual history (Section~\ref{app:zeroshot_executor}).

\subsubsection{Task Proposal}
For every training state of a training game, the VLM is shown the initial frame and the full action space, and asked to propose an exhaustive list of tasks that can be attempted from that state. The prompt is given in Prompt~\ref{prompt:task_proposal}.

Table~\ref{tab:proposed_task_examples} shows a sample of the tasks proposed from one training state of each game.

\begin{table}[h]
\centering
\small
\caption{Example tasks proposed by Gemma-4-31B from a single training state of each training game (a subset of the full proposal list for that state).}
\label{tab:proposed_task_examples}
\begin{tabular}{p{0.2\linewidth}p{0.72\linewidth}}
\toprule
Game (State) & Proposed Tasks \\
\midrule
Pokémon Red (Viridian City) & move right and enter the building using the a button. \newline move down through the gap in the hedge to reach the lower area. \newline move down and interact with the signpost using the a button. \newline open the game menu using the start button. \\
\midrule
Legend of Zelda: Link's Awakening (Near Monster) & defeat the chepom enemy using the b button. \newline move link to the right to reach the staircase. \newline descend/ascend the staircase to transition to the next area. \newline move link to the left wall to check for hidden secrets or boundaries. \\
\midrule
Deja Vu 1 (In Cellar) & interact with the square button on the control panel. \newline interact with the diamond button on the control panel. \newline move toward the wooden door to attempt to leave the cellar. \newline investigate the area directly beneath the hanging light bulb. \\
\midrule
Sword of Hope 1 (Battle) & use the fight command to attack moth1. \newline select the magic menu to view available spells. \newline attempt to escape from the battle. \newline defeat both moth1 and moth2 to end the encounter. \\
\midrule
Bomberman Quest (Next to Guide) & move forward (up) to reach the entrance of the building. \newline enter the building by pressing an action button (a or b) once positioned at the doorway. \newline move to the left (left) to investigate the wooded area. \newline navigate to the rightmost edge of the visible path. \\
\midrule
Harvest Moon 1 (Chicken Coop) & move to and collect the egg lying on the floor to the right of the player. \newline interact with the chicken standing to the left of the player. \newline move to and examine the large barrel on the left side of the room. \newline navigate to the upper right area of the room to exit the structure. \\
\midrule
Runes of Virtue 1 (Near Telescope) & move the character to the left edge of the screen to check for a screen transition. \newline move the character to the right edge of the screen to check for a screen transition. \newline walk up to any nearby npcs or objects and press button a to interact. \newline move the character in a circle to test movement responsiveness. \\
\midrule
Survival Kids 1 (Water Menu Open) & select the ``drink'' option to consume water. \newline select the ``fill'' option to fill a container with water. \newline select the ``leave'' option to exit the interaction menu. \newline move the character toward the trees in the upper right area of the screen. \\
\bottomrule
\end{tabular}
\end{table}

\subsubsection{Task Attempts}
Each proposed task is attempted from its training state by the executor, with a budget of $50$ steps and the ability to declare the task complete. Success is judged by the VLM in two stages. First, the frames of the attempt are split into windows, each window is described without reference to the task, and the descriptions are consolidated into a single description of the full trajectory that refers to frame ranges. Then, given this description and the final $8$ frames, the VLM judges whether the task was completed at any point, and identifies the earliest frame by which it was surely complete. If an attempt fails, the VLM critiques the failed trajectory in the same windowed fashion and writes a hint for the next attempt. Each task is attempted up to $5$ times, stopping at the first success. The prompts are given in Prompts~\ref{prompt:attempt_judgement_window_description}, \ref{prompt:attempt_judgement_consolidation}, \ref{prompt:attempt_judgement_verdict}, \ref{prompt:attempt_critique_window_summary} and~\ref{prompt:attempt_critique_hint}.

Table~\ref{tab:skill_attempts} reports the number of proposed tasks and the number judged successful for each game. Only the tasks judged successful are carried forward to guidance inference and practice.

\begin{table}[h]
\centering
\small
\caption{Task proposal and attempt statistics for the initial step of autonomous skill discovery with Gemma-4-31B. States is the number of training states from which tasks were proposed. Successful is the number of proposed tasks judged successful within $5$ attempts.}
\label{tab:skill_attempts}
\begin{tabular}{lcccc}
\toprule
Game & States & Proposed Tasks & Successful Tasks & Success Rate (\%) \\
\midrule
Pokémon Red                     & 40  & 267 & 164 & 61.4 \\
Legend of Zelda: Link's Awakening & 44  & 297 & 147 & 49.5 \\
Deja Vu 1                       & 42  & 335 & 233 & 69.6 \\
Sword of Hope 1                 & 18  & 127 & 102 & 80.3 \\
Bomberman Quest                 & 35  & 222 & 78  & 35.1 \\
\midrule
Harvest Moon 1                  & 28  & 199 & 86  & 43.2 \\
Runes of Virtue 1               & 41  & 246 & 111 & 45.1 \\
Survival Kids 1                 & 28  & 196 & 111 & 56.6 \\
\bottomrule
\end{tabular}
\end{table}

\subsubsection{Guidance Inference}
For every successful attempt, the VLM infers step-by-step guidance on how to complete the task, along with a visual goal condition that confirms completion. As with judgement, the trajectory is processed in windows of at most $8$ frames, and the guidance from each window is consolidated. The guidance prompt requires every step to be tied to the visual cues that indicate when to perform it. The prompts are given in Prompts~\ref{prompt:guidance_inference_window} and~\ref{prompt:guidance_inference_consolidation}.

\subsubsection{Practice}
Each successful task is practised in $3$ independent episodes, capped at $4{,}000$ episodes per game. At the start of each episode, the environment is reset to the task's training state, and $5$ uniformly random button presses (seeded per episode) are applied to perturb the starting state. The executor then attempts the task with a budget of $50$ steps. The inferred guidance is provided to the executor through its hint channel, while the goal condition is provided only to the judge, which applies it as a strict criterion for success. If an episode fails, a hint is derived from the failed trajectory, the same perturbation is replayed, and the episode is retried once.

The guidance appears only in the executor's prompt during practice, and is removed when the practice data is converted into training data (see below). The resulting VLA must therefore learn to perform the task from the task description alone, without access to the guidance that produced its demonstrations.

\subsubsection{Data Cleaning and Dataset Construction}
Only successful practice episodes are used. Each executor decision (the prompt, the frame and the executor's reasoning and action) becomes a candidate training example, and decisions made more than $2$ steps after the judged success point of an episode are discarded. Two passes are then applied with the VLM:

\noindent\textbf{Decision Filtering:} Every decision is reviewed, along with the frame that resulted from the action, and rejected only if there is clear visual evidence that the executor's reasoning badly misdescribes the frame or that the action failed to advance the task. The prompt errs strongly towards acceptance.

\noindent\textbf{Paraphrasing:} For every task, the VLM generates $3$ paraphrases of the task string. The final paraphrase is reserved for validation, and the others are used to augment the training data.

The guidance and step-count blocks are stripped from every prompt. Episodes are split into training and validation sets per task, with $20\%$ of each task's episodes held out for validation. All decisions of an episode are placed on the same side of the split, and validation examples are phrased with the held-out paraphrase, so that neither the validation episodes nor their phrasing are seen during training. The prompts are given in Prompts~\ref{prompt:decision_filtering} and~\ref{prompt:paraphrasing}.

\subsubsection{VLA Training}
For each training game, we fine-tune a separate LoRA adapter~\citep{hu2021lora} on Gemma-4-31B with supervised fine-tuning on the cleaned dataset. The fine-tuned model is deployed as the executor under the supervisor of Section~\ref{app:zeroshot_supervisor}. Hyperparameters are given in Figure~\ref{fig:vla_hparams}.

\begin{figure}[h]
\centering
\small
\begin{tabular}{ll}
\toprule
Hyperparameter & Value \\
\midrule
Base model & Gemma-4-31B \\
Adapters & One per training game \\
Objective & Supervised fine-tuning \\
LoRA rank & 64 \\
LoRA $\alpha$ & 128 \\
Learning rate & $2\times10^{-4}$ \\
Weight decay & 0.01 \\
Batch size (per device) & 8 \\
Epochs & 2 \\
Action loss weight & 0.25 \\
Model selection & Best validation loss (evaluated every epoch) \\
Early stopping patience & 2 evaluations \\
\bottomrule
\end{tabular}
\caption{Hyperparameters for VLA fine-tuning.}
\label{fig:vla_hparams}
\end{figure}

\subsection{Insight documents}
\label{app:insight}

Insight documents are built from the high-reward subtrajectories surfaced by curiosity-based exploration, using the configuration described in Section~\ref{app:curiosity_docs}. The pipeline consists of four stages: (1) grouping subtrajectories by their final frames, (2) inferring the task performed in each group, (3) extracting insights from each (task, trajectory) pair, and (4) merging the extracted insights into a single document per game. All stages use Gemma-4-31B. At test time, the document is filtered and distilled for the task at hand before being provided to the supervisor.

\subsubsection{Trajectory Grouping}
The high-reward subtrajectories of a training state are first cleaned of cycles: whenever a frame recurs within a subtrajectory (outside of the final $5$ frames), the steps between its occurrences are removed. The subtrajectories are then clustered by the similarity of their final frames, under the assumption that subtrajectories ending in similar frames expose similar game mechanics. Clustering proceeds by recursive bisection: the set of subtrajectories is split in half, each half is clustered, and two clusters from different halves are merged if any sampled final frame of one is sufficiently similar to any sampled final frame of the other (by cosine similarity of their random patch projections). Only subtrajectories with a reward $z$-score of at least $6.0$ are retained.

\subsubsection{Task Inference}
\label{app:synthetic}
For every group, up to $3$ subtrajectories are sampled. For each, the VLM is shown the final $8$ frames and asked to infer the single, most multistep task that the player performed, along with the frames at which it began and was completed, or to respond that no clear task is present. Each inferred task is then checked for validity: tasks that are too generic to be completed exactly (e.g. `explore the area') are discarded, and valid tasks are rewritten as concise imperative instructions. The valid tasks inferred for a group are distilled into a single task string that captures their common core without over-generalising. Finally, groups that distil to the same task string are merged. Groups for which no valid task is inferred are discarded, providing the VLM with a path to filter out noisy trajectories. The prompts are given in Prompts~\ref{prompt:task_inference}, \ref{prompt:task_validation_and_rewriting} and~\ref{prompt:group_task_distillation}.

\subsubsection{Insight Extraction}
For every (task, trajectory) pair, the VLM is shown $8$ frames sampled from the trajectory along with the sequence of actions taken, and asked to extract only the key, non-obvious insights that would help a future player attempting a similar task. The prompt explicitly rejects insights that merely restate the controls, and allows the VLM to respond that nothing non-obvious can be learned, in which case the pair is discarded. Each pair is also assigned two reusable categories: a \textit{task category} describing the kind of task performed, and an \textit{image category} describing the kind of screen on which the trajectory begins. Each surviving pair becomes a small document, with one entry in its task section and (if the starting screen is distinctive) one entry in its image section. Every entry consists of a category, a description, examples, the extracted insights and a representative frame (the first frame of the trajectory). The prompt is given in Prompt~\ref{prompt:insight_extraction}.

\subsubsection{Document Merging}
The per-pair documents are merged into a single document per game through a tournament: in each round, documents are paired and each pair is merged, until a single document remains. A merge folds the entries of the second document into the first, one entry at a time and separately for each section. For every incoming entry, the VLM is shown the entry and all existing entries of that section (with their insights and representative frames), and decides whether the incoming entry teaches the same lesson as an existing one. The matching prompt defaults strongly to no match, since merging entries with genuinely different insights produces vague, unusable entries. If no match is found, the entry is appended unchanged. If a match is found, the two insight lists are combined by the VLM into a single list that merges duplicates and refinements while keeping every insight concrete, and the examples, frames and training states of the incoming entry are appended to the existing entry. Since entries are only ever appended or combined, the merge cannot drop an entry. The prompts are given in Prompts~\ref{prompt:document_merging_entry_matching} and~\ref{prompt:document_merging_insight_combination}.

\subsubsection{Document Statistics}
Table~\ref{tab:insight_docs} reports the size of the insight document at each stage of the pipeline. Inferred Tasks is the number of groups with a valid task after deduplication. Extracted Insights counts the insights drawn from the pairs for which extraction did not return NONE. The final document is reported by section: each entry of the image section carries the insights of the trajectory that created it, so the two sections hold largely overlapping insights organised under different categories.

\begin{table}[h]
\centering
\small
\caption{Insight document statistics for each training game. Pairs is the number of (task, trajectory) pairs from which at least one insight was extracted.}
\label{tab:insight_docs}
\resizebox{\linewidth}{!}{%
\begin{tabular}{lcccccccc}
\toprule
 & & & & \multicolumn{2}{c}{Task Section} & \multicolumn{2}{c}{Image Section} \\
\cmidrule(lr){5-6} \cmidrule(lr){7-8}
Game & Inferred Tasks & Pairs & Extracted Insights & Entries & Insights & Entries & Insights \\
\midrule
Pokémon Red                       & 445 & 288 & 565 & 66 & 247 & 72 & 249 \\
Legend of Zelda: Link's Awakening & 275 & 100 & 200 & 52 & 151 & 57 & 157 \\
Deja Vu 1                         & 193 & 114 & 241 & 53 & 180 & 55 & 169 \\
Sword of Hope 1                   & 120 & 95  & 212 & 43 & 149 & 44 & 153 \\
Bomberman Quest                   & 228 & 167 & 322 & 66 & 208 & 66 & 206 \\
\midrule
Harvest Moon 1                    & 192 & 103 & 180 & 36 & 115 & 39 & 114 \\
Runes of Virtue 1                 & 348 & 86  & 148 & 45 & 120 & 51 & 118 \\
Runes of Virtue 2                 & 487 & 187 & 332 & 76 & 230 & 83 & 222 \\
Survival Kids 1                   & 79  & 61  & 114 & 27 & 81  & 23 & 78  \\
\bottomrule
\end{tabular}}
\end{table}

We show example task entries from the final document of each game below. Insights are reproduced verbatim.

\begin{tcolorbox}[breakable, title=Example Insights: Pokémon Red]
\small
\textbf{Check a Pokémon's stats} -- Navigate the menu system to view the detailed base stats of a specific Pokémon in the party.
\begin{itemize}
\item To open the POKéMON (party) menu from the main menu, use the directional pad (UP, DOWN, or RIGHT) to highlight the ``POKéMON'' option and press A.
\item In the POKéMON menu, the visual cue is a list of Pokémon showing their current HP and level; use the directional pad (UP/DOWN) to position the cursor (a small black triangle/arrow to the left of the Pokémon's icon) on the specific Pokémon's name.
\end{itemize}
\textbf{Navigate battle and world menus} -- Transition from a battle screen back to world exploration and opening the main menu.
\begin{itemize}
\item The ``Run'' command can only be executed after the current turn's animations and text (e.g., ``Enemy PIDGEY used GUST!'') have been cleared by pressing A.
\item To reach the ``RUN'' option from the default ``FIGHT'' position in the 2x2 command grid, the player can press RIGHT then DOWN, or DOWN then RIGHT.
\end{itemize}
\end{tcolorbox}

\begin{tcolorbox}[breakable, title=Example Insights: Legend of Zelda: Link's Awakening]
\small
\textbf{Overworld navigation} -- Moving the player character to a specific geographic coordinate or landmark on the game map.
\begin{itemize}
\item The player cannot walk through the dense clusters of dark, rounded foliage (trees); these act as hard collisions that block movement.
\item To reach the upper right area from the starting position, the player must first move RIGHT to bypass the central tree line before pressing UP, otherwise, the player will be blocked by the foliage.
\end{itemize}
\textbf{Menu navigation} -- Navigating system or game-state menus to save progress and exit.
\begin{itemize}
\item The transition to the Game Over menu is triggered automatically when the player's health (hearts in the top right) reaches zero.
\item On the Game Over screen, press DOWN once to move the cursor from ``SAVE \& CONTINUE'' to ``SAVE \& QUIT''.
\end{itemize}
\end{tcolorbox}

\begin{tcolorbox}[breakable, title=Example Insights: Deja Vu 1]
\small
\textbf{Interact with vehicle interior} -- Navigating a point-and-click interface to search for items inside a car.
\begin{itemize}
\item To interact with the notebook/planner, the cursor must be moved into the right-hand panel area; using RIGHT multiple times shifts the focus from the scene to the notebook.
\item Pressing A while the cursor is highlighting a specific dashboard element (like the glove box) triggers the interaction/search action.
\end{itemize}
\textbf{Inspect object} -- Using an interface to examine a specific environmental detail to gain information.
\begin{itemize}
\item To inspect the card slot, the cursor must be positioned precisely over the small rectangular indentation to the right of the double doors before pressing the interaction button.
\end{itemize}
\end{tcolorbox}

\begin{tcolorbox}[breakable, title=Example Insights: Sword of Hope 1]
\small
\textbf{Cast spell in battle} -- Selecting a specific magic spell from a menu during an encounter to attack an enemy.
\begin{itemize}
\item When navigating the battle menu, the cursor is not always on the desired action by default; you must use the DOWN button to move the selection circle from `FIGHT' to `MAGIC'.
\item In the magic selection menu, the cursor does not start at the top; it starts on the second spell (FIREBAL2), so press UP to select the first spell (FIREBALL).
\end{itemize}
\textbf{Access the item use menu} -- Navigating from the main gameplay screen to the item selection menu to use an item.
\begin{itemize}
\item To reach the ``USE'' command from a message box, press START to return to the action menu, then press DOWN to enter the action grid.
\item Once in the action grid, the cursor must be moved to the right column to access ``USE''. Press RIGHT multiple times until the circle cursor highlights ``USE'' before pressing A.
\end{itemize}
\end{tcolorbox}

\begin{tcolorbox}[breakable, title=Example Insights: Bomberman Quest]
\small
\textbf{Push object} -- Moving a physical object in a specific direction by interacting with it.
\begin{itemize}
\item Pressing B while adjacent to a bomb drops/places the bomb rather than pushing it.
\item To push a bomb, the player must move directly into the bomb's space using a directional button (e.g., LEFT) without pressing the action button B first.
\end{itemize}
\textbf{Access game encyclopedia} -- Navigating from active gameplay through the system menu to a specific information database.
\begin{itemize}
\item To reach the Menu from a screen showing item details, press B to close the detail window first.
\item The monster encyclopedia is accessed via the Menu by selecting the icon featuring a monster's head (third icon from the left on the bottom row).
\end{itemize}
\end{tcolorbox}

\begin{tcolorbox}[breakable, title=Example Insights: Harvest Moon 1]
\small
\textbf{Purchase an item from a shop} -- Navigate a shopkeeper's menu to select and buy a specific animal or item.
\begin{itemize}
\item When using the shop menu, the cursor (indicated by a small pointer above the icon) wraps around; if the desired item is at the end of the list, continue pressing RIGHT to return to the start.
\item To confirm a selection from the icon list, press A; this triggers a confirmation dialogue (``Would you like...?'') where A must be pressed again on ``Yes'' to finalize the purchase.
\end{itemize}
\textbf{Break a rock to obtain an item} -- Use a tool to destroy a breakable rock object to reveal a hidden item.
\begin{itemize}
\item To break a rock, the player must be positioned directly adjacent to it and facing it; the action button (A) only triggers the tool when the player's sprite is aligned with the object's hit-box.
\item Repeatedly pressing the tool button (A) without movement is required to break through the rock's durability before the item is spawned.
\end{itemize}
\end{tcolorbox}

\begin{tcolorbox}[breakable, title=Example Insights: Runes of Virtue 1]
\small
\textbf{Defeat a coastal enemy} -- Navigate to and engage an enemy located on the shoreline.
\begin{itemize}
\item Pressing the A button while adjacent to an enemy triggers an attack animation; multiple presses of A are required to deplete the enemy's health and defeat it to remove it from the path.
\item Combat occurs in the overworld without transitioning to a separate battle screen.
\end{itemize}
\textbf{Combat enemy in enclosed area} -- Navigating a maze-like environment to position the player for an attack on an enemy.
\begin{itemize}
\item Brick walls act as hard barriers; you must navigate through the arched openings to enter or exit the enclosure.
\item Exiting the room via the bottom door (walking DOWN into the doorway) occurs immediately upon contact, which may lead to failure if the enemy is not yet defeated.
\end{itemize}
\end{tcolorbox}

\begin{tcolorbox}[breakable, title=Example Insights: Runes of Virtue 2]
\small
\textbf{Select inventory item} -- Navigating the inventory menu to highlight and select a specific item for use.
\begin{itemize}
\item To select an item in the inventory, use the RIGHT directional button to move the selection cursor; an item is successfully highlighted when a box appearing around the item icon (e.g., the ankh) and the ``QUANTITY'' value updates to reflect the number of that specific item held.
\item Pressing B while an item is highlighted closes the inventory menu and returns the player to the world map.
\end{itemize}
\textbf{Navigate to outdoor area} -- Moving the player character from an interior building to an exterior environment.
\begin{itemize}
\item To exit the building, the player must locate the specific arched doorway on the far left wall that connects to the paved outdoor path; other archways lead to different internal rooms.
\item Navigating through interior archways requires precise alignment with the center of the opening; if the character does not transition screens, reposition the character slightly and press the direction button again.
\end{itemize}
\end{tcolorbox}

\begin{tcolorbox}[breakable, title=Example Insights: Survival Kids 1]
\small
\textbf{Decline item interaction} -- Interact with an item on the ground and choose an option other than taking it.
\begin{itemize}
\item When the interaction menu appears (e.g., Take, Eat, Leave), use the UP and DOWN buttons on the directional pad to navigate the cursor.
\item To decline the item and close the interaction menu, highlight ``Leave'' and press START to confirm.
\end{itemize}
\textbf{Drop an item} -- Navigating the item menu to remove an object from the player's inventory and place it in the game world.
\begin{itemize}
\item To access the ``Drop'' command, the item must first be selected in the ``Items'' list; once selected, a secondary action menu appears at the bottom containing ``Unequip'' and ``Drop''.
\item The ``Merge'' menu and ``Items'' menu are distinct; if the player is in the ``Merge'' screen, they must exit or switch modes to reach the standard item management menu where ``Drop'' is available.
\end{itemize}
\end{tcolorbox}

\subsubsection{Test-Time Usage}
The insight document is used by the Subgoal Planning supervisor of Section~\ref{app:zeroshot_supervisor}, and does not alter the executor. Before planning, the supervisor selects and condenses the knowledge relevant to the task at hand in three steps:

\noindent\textbf{Relevance Judgement:} For every entry in both sections of the document, the VLM is shown the task, the entry's category, description and examples, the current screen and the entry's representative frame, and judges whether the entry fits the current situation. The entry's insights are deliberately withheld from this judgement, so that relevance is decided by the situation an entry describes rather than by its advice. These judgements are independent, and are made in parallel.

\noindent\textbf{Insight Filtering:} The insights of all relevant entries are pooled and numbered, and the VLM removes those that are clearly about something absent from, and unrelated to, the current task. The prompt errs strongly towards keeping insights, since the player will move between rooms and menus while completing the task.

\noindent\textbf{Insight Distillation:} The remaining insights, which were recorded by different trajectories and may repeat or contradict one another, are rewritten into a short, ordered list of concrete statements. Redundant insights are merged into their most specific form, generic advice is removed, and contradictions are stated rather than resolved.

The resulting list fills the knowledge slot of the planning, self-revision and error correction prompts (Section~\ref{app:zeroshot_supervisor}), which is otherwise filled with \texttt{(nothing recorded)}. It is computed once per episode and reused by every supervisor call. If no entry is judged relevant, the episode proceeds exactly as with the zeroshot supervisor. The prompts are given in Prompts~\ref{prompt:test_time_usage_relevance_judgement}, \ref{prompt:test_time_usage_insight_filtering} and~\ref{prompt:test_time_usage_insight_distillation}.

\section{GameBoyWorlds-Playthrough}
\label{app:playthrough}
ROM hacks are fan-made games created by modifying the read-only memory (ROM) image of an official Game Boy title. Using community-developed tools, developers rewrite the original game's code, maps, text, music, and graphics to create a new game that runs on an emulator or modified hardware. Because a ROM hack retains the original game's engine, controls, and screen format, it looks and plays like a Game Boy Pokémon game while introducing a new world, story, and gameplay mechanics. Both hacks in our testbed were developed primarily by a software developer known in the community as \emph{Koolboyman}. Pokémon Brown is built on Pokémon Red, while its sequel, Pokémon Prism, is built on Pokémon Crystal. The two games share a continuous story and overlapping geography.

\begin{figure}[htbp]
    \centering
    \includegraphics[width=\linewidth]{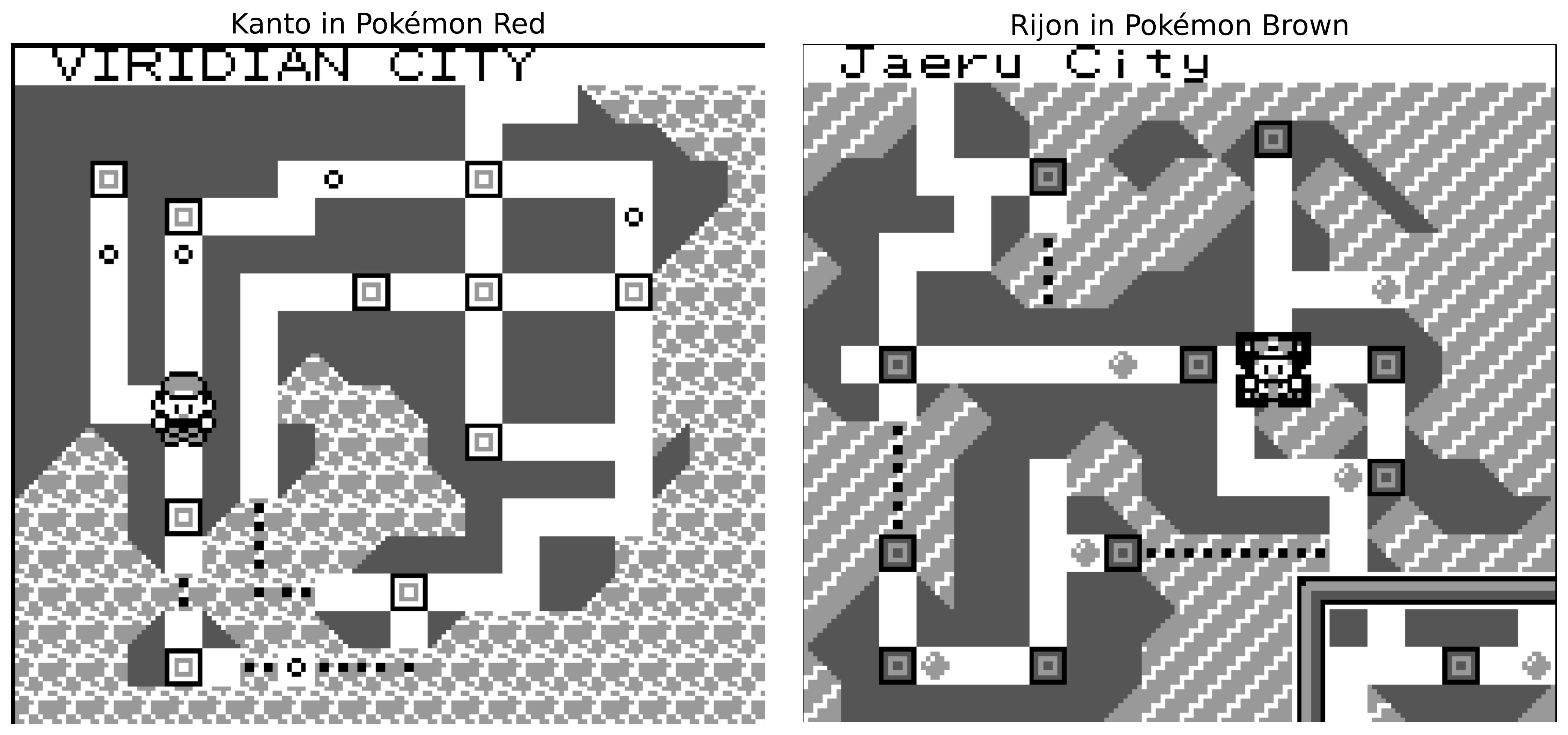}
    \caption{Comparison of the region maps in Pokémon Red and Pokémon Brown.}
    \label{fig:brown-vs-red-map}
\end{figure}

\noindent\textbf{Pokémon Brown}: Brown is a Generation~I hack of Pokémon Red set in \emph{Rijon}, a new industrial island region. The player
has just moved to \emph{Rijon} and, like in Red, receives a starter Pokémon and
a Pokédex from the local professor, then sets out with a rival to fill
the Pokédex and earn the region's eight gym badges. Along the way, Team
Rocket arrives from Kanto and takes over a city and its warehouse, and the
player must drive them out before going on to win the Rijon League. After
the League, a new leader of Team Rocket reveals a plan to derail the mine
cart of a boy, who becomes the protagonist of Prism, and strand him in a
distant region. Figure~\ref{fig:brown-vs-red-map} compares the
Rijon map with Red's Kanto map: the two share the same tile set and visual
style, but Rijon's towns, routes, and dungeons are entirely new. Brown
also differs from Red in its Pokémon and battle rules. Red contains only
the original 151 Pokémon, while Brown adds Pokémon from
Generations~II and III and a few fan-designed evolutions. Brown also adds
a new Wood type to Red's 15 types and moves some existing Pokémon into
it, which changes which attacks are strong or weak against them.

\begin{figure}[htbp]
    \centering
    \includegraphics[width=\linewidth]{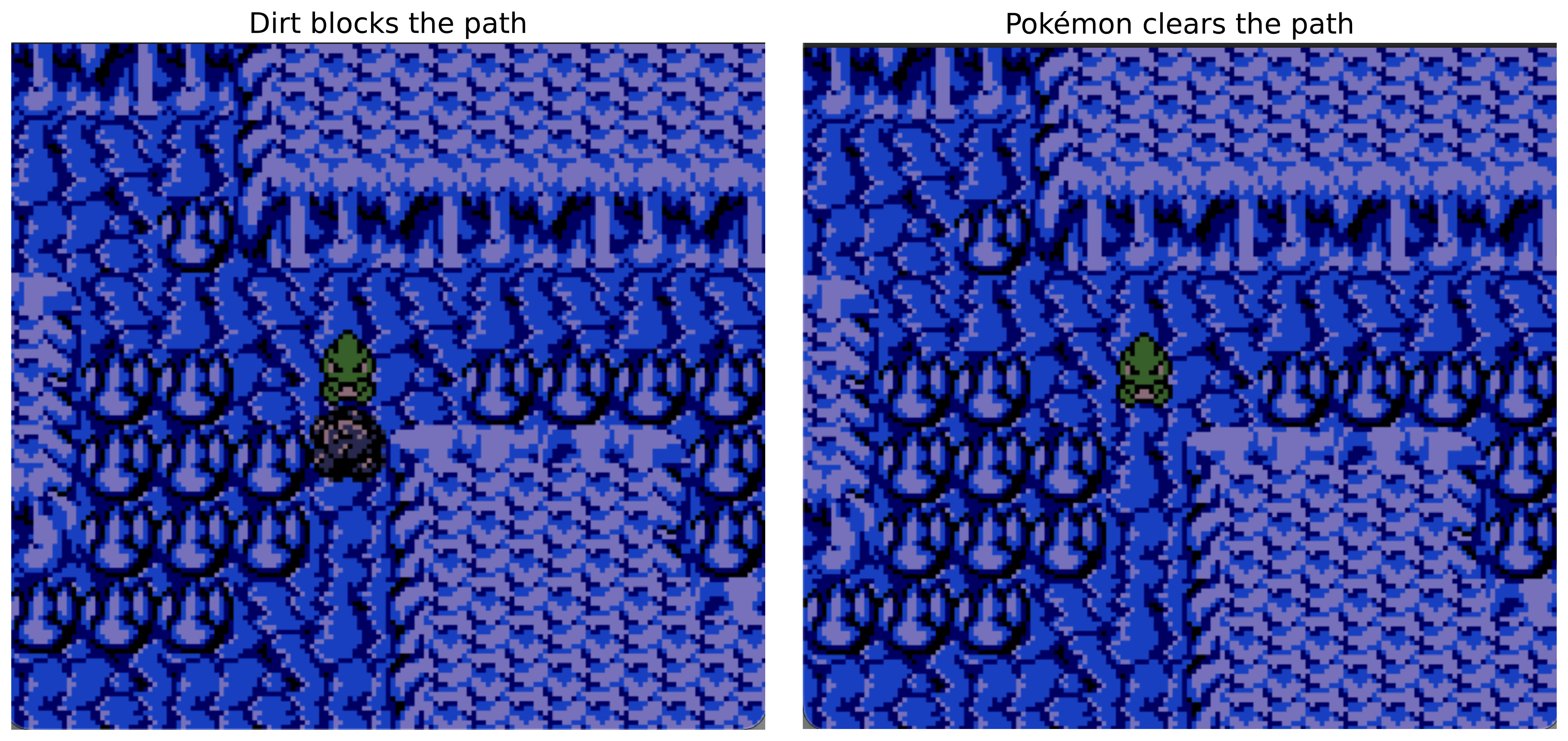}
    \caption{A player-controlled Pokémon clearing a blocked passage in Pokémon Prism by eating the dirt obstruction.}
    \label{fig:prism-pokemon-mode}
\end{figure}

\noindent\textbf{Pokémon Prism}: Prism is built on Pokémon Crystal (Generation~II) and is set mainly in \emph{Naljo}, another new region. The player is the child of Lance, the Dragon-type champion from Crystal, and ends up stranded in Naljo after the mine-cart accident set up at the end of
Brown. Naljo is caught up in a movement pushing for rapid, ``pure''
industrial progress, and a nationalist group called the Palette
Patrollers plans to awaken a set of ancient guardian Pokémon and corrupt
them for its own ends. The player travels across Naljo earning badges,
stops the Palette Patrollers, and wins the Naljo League. After that, the
game opens up to Rijon (from Brown) and parts of Johto and Kanto, where
the player continues collecting badges, 20 in total compared with
Crystal's 16, and uncovers who was behind the plot. Prism changes its base
game more than Brown does. It adds two new types, Gas and Sound, to
Crystal's 17. Its most distinctive new feature is shown in Figure~\ref{fig:prism-pokemon-mode}: in certain areas the player shrinks down and walks around the overworld as their own Pokémon sometimes in a side-scrolling view, to reach places and solve puzzles. In Crystal, as in the official Pokémon games generally, the player only ever walks around as the trainer, and Pokémon are used only in battle.

\subsection{Contamination Check}
\label{app:contamination}
We assess potential pretraining-data contamination by comparing model performance
on two official Pokémon titles, Pokémon Red and Pokémon Crystal, with performance on Pokémon Brown and Pokémon Prism. The official titles have been extensively documented in strategy guides, wikis, and online discussions for several decades. In contrast, the ROM hacks have a substantially smaller documentation footprint. The four games nevertheless share similar interfaces, terminology, progression structures, and factual question types, allowing documentation density to vary while largely preserving the form of the task.

\noindent\textbf{Knowledge-check construction:} We constructed 100 short-answer questions for each game, producing 400 questions in total. The questions target concrete facts required for game progression, including event triggers, non-player characters, key items, field moves, badges, and route connectivity. Questions about optional content, wild-Pokémon encounter tables, and subjective strategies were excluded. Reference answers were intentionally concise, with a median length of three tokens and a mean length of 4.62 tokens.

The Pokémon Brown questions were written from a first-hand playthrough and
associated walkthrough notes. The questions for Pokémon Red, Pokémon Crystal, and Pokémon Prism were derived from walkthrough material and manually checked against the corresponding source before inclusion. Table~\ref{tab:knowledge-examples} shows three representative examples from each game.

\begin{table*}[t]
\centering
\small
\caption{Representative knowledge-check questions. Each question requests a
single verifiable fact concerning the corresponding game.}
\label{tab:knowledge-examples}
\begin{tabular}{p{0.12\textwidth} p{0.60\textwidth} p{0.20\textwidth}}
\hline
Game & Question & Reference answer \\
\hline
Pokémon Red
& Which Badge does Brock award after his defeat?
& Boulder Badge \\

& Which HM does the player receive from the captain of the S.S.\ Anne?
& HM01, Cut \\

& Which item allows the player to identify the ghosts in Pokémon Tower?
& Silph Scope \\
\hline
Pokémon Crystal
& Which Badge does Falkner award after his defeat?
& Zephyr Badge \\

& Which item is used to move the Sudowoodo blocking the path?
& SquirtBottle \\

& Which field move allows the player to cross bodies of water?
& Surf \\
\hline
Pokémon Brown
& Which named tunnel must the player traverse from eastern Mt.\ Boulder?
& Rijon Tunnel \\

& What happens when the player first enters Gravel Town?
& Professor Tim meets the player and takes them to his laboratory \\

& In which region does Pokémon Brown primarily take place?
& Rijon \\
\hline
Pokémon Prism
& What must the player collect to light the fireplace near the beginning of
the game?
& Firewood \\

& Which Pokémon does the player obtain in the Acqua Mines?
& Larvitar \\

& Which professor assists the player during the early portion of the game?
& Professor Ilk \\
\hline
\end{tabular}
\end{table*}

\paragraph{Evaluation protocol:}
We evaluated GPT-5.6-sol, Claude Opus 5, Gemini 3.5 Flash-Lite, and Gemma:4-31B. Each question was submitted as an independent request containing only the game name and the question. No previous questions, previous responses, or reference answers were included in the model context. Retrieval and Web search were disabled where supported by the inference interface. For the components in which tools could not be disabled structurally, the execution metadata confirmed zero tool calls for every response.

Aside from automatic metrics, we manually classified each answer as \emph{Right}, \emph{Partially
Right}, or \emph{Wrong} according to its semantic agreement with the reference answer. Wording, capitalization, formatting, and answer order were ignored. The primary measure is strict human accuracy, under which only answers labelled \emph{Right} receive credit. The 32 partially correct responses across all models constitute only 2.0\% of the 1,600 evaluated responses and do not affect the principal conclusion.

\noindent\textbf{Human-evaluated results:} Table~\ref{tab:contamination-per-game} reports strict human accuracy separately for each game. All models perform substantially better on the official titles than on the ROM hacks. GPT-5.6-sol answers all 200 official-game questions correctly but only 58 of the 200 ROM-hack questions. Claude Opus 5 and Gemini 3.5 Flash-Lite achieve at least 92\% accuracy on each official title, while
their ROM-hack accuracy ranges from 8\% to 16\%. Gemma:4-31B exhibits lower
official-game accuracy, but its performance also falls to 8\%--16\% on the ROM hacks.

\begin{table}[t]
\centering
\small
\caption{Strict human-annotated accuracy by game. Each cell contains 100
questions.}
\label{tab:contamination-per-game}
\begin{tabular}{lrrrr}
\hline
Model & Red & Crystal & Brown & Prism \\
\hline
GPT-5.6-sol             & 100\% & 100\% & 29\% & 29\% \\
Claude Opus 5           &  97\% &  93\% &  8\% & 16\% \\
Gemini 3.5 Flash-Lite   &  92\% &  94\% & 13\% & 15\% \\
Gemma 4 31B             &  60\% &  61\% &  8\% & 16\% \\
\hline
\end{tabular}
\end{table}

Pooling the two official titles and the two ROM hacks produces accuracy
differences ranging from 48.5 to 83.0 percentage points
(Table~\ref{tab:contamination-pooled}). A two-sided Fisher exact test applied
to the binary \emph{Right}/\emph{not-Right} labels rejects independence between
game category and answer correctness for every model
(\(p < 10^{-24}\)). The result is therefore not attributable to a small number
of isolated questions or to a single evaluated model.

\begin{table}[t]
\centering
\small
\caption{Strict human accuracy on the pooled official and ROM-hack subsets.
Each subset contains 200 questions per model.}
\label{tab:contamination-pooled}
\begin{tabular}{lrrrr}
\hline
Model & Official & ROM hacks & Gap & Fisher \(p\) \\
\hline
GPT-5.6-sol
& 100.0\% & 29.0\% & 71.0 pp & \(6.0 \times 10^{-61}\) \\

Claude Opus 5
& 95.0\% & 12.0\% & 83.0 pp & \(7.5 \times 10^{-72}\) \\

Gemini 3.5 Flash-Lite
& 93.0\% & 14.0\% & 79.0 pp & \(7.5 \times 10^{-64}\) \\

Gemma 4 31B
& 60.5\% & 12.0\% & 48.5 pp & \(6.5 \times 10^{-25}\) \\
\hline
\end{tabular}
\end{table}

The difference also changes the apparent model hierarchy. Accuracy across
models spans 39.5 percentage points on the official titles, from 60.5\% to
100.0\%, but only 17.0 percentage points on the ROM hacks, from 12.0\% to
29.0\%. In particular, Claude Opus 5, Gemini 3.5 Flash-Lite, and Gemma 4 31B
are separated by only two percentage points on the pooled ROM-hack subset.
Thus, much of the separation observed on the official games does not persist
when the models are evaluated on less extensively documented games with a
closely matched task structure.

For completeness, Table~\ref{tab:automatic-evaluation} reports the automatic
evaluation measurements computed over all 400 questions for each model.

\begin{table}[t]
\centering
\small
\caption{Automatic evaluation results over all 400 questions. The LLM-judge
columns report classification-based accuracy, whereas BLEU-2 and BERTScore-F1
are mean similarity scores.}
\label{tab:automatic-evaluation}
\begin{tabular}{lrrrr}
\hline
Model & LLM judge & Judge partial & BLEU-2 & BERTScore-F1 \\
\hline
GPT-5.6-sol
& 53.2\% & 57.5\% & 0.291 & 0.557 \\

Claude Opus 5
& 49.2\% & 53.0\% & 0.157 & 0.330 \\

Gemini 3.5 Flash-Lite
& 37.0\% & 42.9\% & 0.173 & 0.455 \\

Gemma 4 31B
& 25.5\% & 30.0\% & 0.140 & 0.440 \\
\hline
\end{tabular}
\end{table}  

\begin{table}[t]
\centering
\small
\caption{Automatic evaluation results by game and model. LLM-judge accuracy
counts only answers classified as \emph{Right}. Judge partial credit assigns
1 point to \emph{Right}, 0.5 points to \emph{Partially Right}, and 0 points to
\emph{Wrong}. BLEU-2 and BERTScore-F1 are mean similarity scores over the 100
questions for each game.}
\label{tab:automatic-metrics-by-game}
\begin{tabular}{llrrrr}
\hline
Model & Game & LLM judge & Judge partial & BLEU-2 & BERTScore-F1 \\
\hline
GPT-5.6-sol
& Pokémon Red     & 91.0\% & 94.0\% & 0.458 & 0.699 \\
& Pokémon Crystal & 90.0\% & 94.5\% & 0.401 & 0.732 \\
& Pokémon Brown   & 13.0\% & 19.0\% & 0.164 & 0.409 \\
& Pokémon Prism   & 19.0\% & 22.5\% & 0.140 & 0.387 \\
\hline
Claude Opus 5
& Pokémon Red     & 91.0\% & 93.5\% & 0.244 & 0.421 \\
& Pokémon Crystal & 90.0\% & 91.5\% & 0.268 & 0.460 \\
& Pokémon Brown   &  5.0\% & 10.5\% & 0.049 & 0.208 \\
& Pokémon Prism   & 11.0\% & 16.5\% & 0.069 & 0.231 \\
\hline
Gemini 3.5 Flash-Lite
& Pokémon Red     & 67.0\% & 75.0\% & 0.312 & 0.591 \\
& Pokémon Crystal & 72.0\% & 80.0\% & 0.272 & 0.601 \\
& Pokémon Brown   &  4.0\% &  7.0\% & 0.050 & 0.322 \\
& Pokémon Prism   &  5.0\% &  9.5\% & 0.056 & 0.307 \\
\hline
Gemma:4-31B
& Pokémon Red     & 47.0\% & 51.5\% & 0.212 & 0.526 \\
& Pokémon Crystal & 45.0\% & 51.5\% & 0.191 & 0.542 \\
& Pokémon Brown   &  4.0\% &  8.5\% & 0.070 & 0.353 \\
& Pokémon Prism   &  6.0\% &  8.5\% & 0.088 & 0.340 \\
\hline
\end{tabular}
\end{table}

\paragraph{Conclusion.}
The consistent and statistically significant performance reduction on Pokémon Brown and Pokémon Prism indicates that these ROM-hack benchmarks are substantially less contaminated than the official-game benchmarks. Because the four games use closely related terminology, mechanics, and question formats, the 48.5--83.0 percentage-point reduction is most consistent with a large difference in prior corpus exposure rather than a corresponding change in the basic form of the task. These results support using the ROM-hack subset as a more conservative measure of knowledge that cannot be explained primarily by recall of extensively documented official-game material.

\subsection{Agent Design}
\label{app:playthroughagent}
The agent is built in two layers over the game. A \textbf{supervisor} takes one short-horizon task and carries it out through the action primitives below, in a single episode. A \textbf{strategist} sits above it and owns everything that outlives one task: what the agent is trying to achieve, what it has learned about the game, and where in the world it is. The strategist chooses each task, hands it down, and folds the result back into its own state.

Neither layer reads the emulator's memory, and neither is given any knowledge of the game beyond its name. Everything the agent knows about Rijon or Naljo it has seen on the screen. The strategist's state is therefore not configuration but the product of play: a \textbf{goal tree}, a \textbf{knowledge tree}, a \textbf{location store} and a \textbf{notepad}, over a \textbf{tile database} built by the perception layer. These four artifacts are what a playthrough actually accumulates, and we describe each in turn. Every prompt named below is given in Appendix~\ref{app:prompts_strategist} and~\ref{app:prompts_executors}.

\paragraph{Perception.} The screen is treated as a grid of $16\times16$ pixel cells with the player fixed at the origin, $x$ increasing to the right and $y$ upward. Each cell is split into four $8\times8$ tiles, and each visually distinct tile is classified exactly once: a VLM is shown the full screen with the tile outlined, the tile on its own, and the cell containing it, and assigns one of fifteen categories (ground, ledge, obstacle, entrance, water, tall grass, item, NPC, sign, staircase, and so on) together with a short description naming the whole object the tile belongs to, such as ``door of the Poke Mart'' rather than ``top-left corner of a door''. Because the classification is keyed to the tile's appearance, it is reused everywhere that tile recurs, and the tile database grows into a game-specific vocabulary over the playthrough.

Two views are derived from a classified screen. The first is a verbalisation, listing the cell coordinates of every object, entrance and character while suppressing plain ground and obstacles; this is the \texttt{[SCREEN\_TILES]} block that every layer of the agent is shown alongside the frame. The second is a walkability map, collapsing the categories into walkable, ledge and blocked cells, which turns movement into a search problem rather than a perception problem.

\paragraph{Action primitives.} The supervisor acts through seven primitives, each of which assumes a particular game state, refuses to run outside it, and returns a structured outcome code rather than free text. This split matters because the modes of a Pokémon game demand different control: walking is a geometry problem, battles and menus are cursor problems, and dialogue is neither.

\begin{center}
\small
\begin{tabular}{@{}lll@{}}
\toprule
\textbf{Primitive} & \textbf{Assumed state} & \textbf{Mechanism} \\
\midrule
MoveTo & free roam & VLM names the target cell; search plans the route \\
MoveOff & free roam & VLM picks a screen-edge cell; search plans the route \\
Interact & free roam & VLM locates the target among adjacent cells \\
Battle & in battle & VLM chooses a battle-menu option each turn \\
Menu & in menu & VLM chooses one cursor press each turn \\
Dialogue & in dialogue & scripted advance until the dialogue ends \\
Free & any & VLM chooses one raw button press each turn \\
\bottomrule
\end{tabular}
\end{center}

Movement is the clearest case. The VLM is asked only to name the cell holding the target, or to declare the description ambiguous and list the candidates; the route to that cell is then found by breadth-first search over the walkability map, with ledges traversable in one direction only, and if the target itself cannot be stood on the agent walks to an adjacent cell and turns to face it. The model is thus never asked to count button presses, which it is poor at, only to say what it can see, which it is better at.

\paragraph{Episodes.} Given a task, the supervisor writes a plan of subgoals expressed as states of the game to reach, then works through them: it chooses one primitive per step, runs it, and judges the result from the screens before and after, any text read from the screen, and the primitive's outcome code. Failures are retried, an unexpected battle or menu is cleared and the subgoal resumed, and a subgoal that will not yield triggers a replan or an admission that the task cannot be done from here.

The episode ends as a \textbf{digest}: the task, the guidance it came with, the outcome, the final plan, an ordered account of what happened, and every piece of text the player read along the way. This digest is the \emph{only} record of the episode that reaches the strategist. It sees no frames from within the episode and nothing of the primitives' internals, so every update to the artifacts below is made from this text and from the screen at the episode's end.

\paragraph{Goal tree.} The goal tree holds what the agent is trying to do, as a tree whose root is clearing the game. It is seeded with the structure the series itself makes explicit and nothing more: obtain a first Pokémon, obtain each of the eight badges, become the Champion, with the badge goals split into reaching the gym's city and defeating its leader. This skeleton mirrors the progress markers of Appendix~\ref{app:markers} and is the only game-structural prior the agent receives. It contains no information about \emph{where} any gym is, who leads it, or what the game requires in between, all of which must be discovered.

Everything below the skeleton is grown by the agent. The \textbf{frontier} is the deepest leftmost goal still open, found by descending from the root through open goals, and is what the agent is working on at any moment. When the frontier moves to a goal that has not been planned, or when a goal has sat on the frontier for several episodes without being achieved, the agent breaks it down in two steps. It first writes what it would need to know to plan the goal, as questions put to the knowledge tree and as destinations put to the location store; crucially, these questions are answered without any of the surrounding context, so the prompt requires each to be self-contained, naming places and people rather than saying ``here'' or ``the next gym''. It then writes the subgoals themselves, given the recalled facts, the routes that came back, and the places on its map that may lead somewhere unvisited. Each subgoal carries a description, optional details, and a justification of why the parent goal needs it. A goal the agent judges to be directly achievable is marked as such and left as a leaf, so the tree deepens only where planning actually requires it.

After every episode the tree is reconciled against the digest. Goals are closed from the frontier upward for as long as the evidence shows them achieved, which is how finishing one subgoal can complete its parent and move the frontier on. A frontier goal that has not been achieved is separately reconsidered: it may be kept, rewritten if it was aimed at the wrong thing, or abandoned outright. Abandonment is restricted to goals the agent created, so it can discard its own bad plans but cannot drop the milestones it was seeded with.

\paragraph{Knowledge tree.} The knowledge tree is the agent's long-term memory of the game, organised by topic rather than by time. Each entry is a title and a description; the tree is seeded with seven headings, covering the active story, background lore, NPCs, locations, items, Pokémon, and a last heading for the agent's own control of the game, in which it records which kinds of task and guidance it has found workable. Everything beneath these headings is written by the agent.

Facts enter from two places: extraction from an episode digest, and distillation from the notepad when a goal closes. Neither path writes to the tree directly. A new fact is instead routed down from the root one level at a time, the agent choosing at each step which entry it belongs under, or declaring that none fits and it needs a new entry at that level. At a leaf, it decides whether the entry already says this, whether the entry is about the same thing but is incomplete or wrong and should be rewritten, or whether the fact concerns something else and deserves an entry of its own. Creating an entry may also introduce a heading above it, so the taxonomy deepens as material accumulates instead of being reorganised on a schedule.

Reading is the same descent made branching: given a query, the agent picks the entries at each level most likely to hold something relevant, up to three, and prunes the rest, returning the leaves it arrives at. Because entries are retrieved far from the episode that produced them, both writing prompts insist that a fact be intelligible on its own: it must name the people, places and things it concerns rather than referring to them as ``he'', ``there'' or ``it''.

\paragraph{Location store.} The location store is the agent's map. Places are of three kinds: \emph{major} locations, the towns and routes the game's own map names; \emph{bridge} locations, whose purpose is to connect major ones, such as gatehouses, caves and tunnels; and \emph{internal} locations, the rooms inside a major location, each recorded with the location it sits in and each floor of a building counted separately. Connections are held as a graph over major and bridge locations, and a second graph within each major location for the rooms inside it. Every connection records how the player made the crossing and whether the same way leads back. Each place also accumulates a visual signature from the tile database, tying the map back to what the screen looked like there.

The map is built by watching for transitions rather than by being told about them. The Game Boy blanks the screen when the player changes area, so the agent scans an episode's frames for runs of black, and for each one compares the frames before and after. It first asks whether a move between places actually occurred, since battles, menus and cutscenes blank the screen too and must be rejected. If it did, it names the place arrived at, its kind, the location it belongs to if it is a room, how the player got there, and whether the route is reversible, and records the connection. If the screens do not show enough to name the place, the agent marks its position as unknown, which is what prompts it to locate itself from a fresh screen at the start of the next episode.

The map is then read at planning time in two ways. A destination named in a plan, whether a specific place or a kind of place such as somewhere to heal, is matched against the known places and turned into a route by shortest-path search, reported as the sequence of crossings to make. Separately, the agent identifies its own frontier: places with no known way out of them, or with at most one known neighbour, each offered with a route from where the player stands. This is the mechanism by which the agent is meant to notice that part of the world remains unopened, and it is what the divergence in Section~\ref{sec:playthrough} implicates. In Pokémon Red the agent has enough prior knowledge to name its destination outright; in Pokémon Brown, where it does not, reaching the neighbouring town requires the map to tell it that the starting town has an unexplored exit at all.

Two limits follow from building the map this way. It contains only crossings the player has actually made, so a place the agent has merely been told about cannot be routed to until it has been reached once. And it is purely topological: it records that one place connects to another and how, never where either lies in the world.

\paragraph{Notepad.} The notepad is a free-text scratchpad for the goal currently being pursued, holding what has been tried, what worked, what failed, and what to try next. After each episode the agent either appends to it or rewrites it in full, the latter when parts of it have gone stale. It is read back into the dispatch and planning prompts, and is the one place where the agent can keep a thought that is not yet a fact and not yet a goal.

Unlike the other artifacts the notepad is deliberately temporary. When the goals it was written for are closed, the agent is shown the notepad one last time, asked which of its contents will still matter long after those goals, and those facts are routed into the knowledge tree; the notepad is then cleared for the next goal. It is thus the passage between the working detail of an episode and the agent's permanent memory of the game.

\paragraph{The loop.} An episode proceeds as follows. The agent perceives the current screen and, if it does not know where it is, locates itself. It plans the frontier goal into subgoals if that goal is new or has stalled. It then writes one task and its guidance from the frontier goal, the current place, the notepad and the screen, and hands them to the supervisor. When the episode returns, the agent scans its frames for transitions and updates the map, re-perceives the screen, checks the goal tree against the digest, summarises the episode for its log, extracts facts into the knowledge tree, and updates the notepad, distilling and clearing it if a goal has closed. The next episode begins from the artifacts as they now stand.

\section{Complete Set of Prompts}
\label{app:prompts}

\newtcolorbox[auto counter]{promptbox}[2][]{breakable, title={Prompt~\thetcbcounter: #2}, #1}

This section lists every prompt used by the agents and pipelines in this appendix. Text in square brackets (e.g. \texttt{[TASK]}) denotes a slot that is filled in at runtime.

\subsection{Zeroshot Agent}

\begin{promptbox}[label=prompt:subgoal_planning]{Subgoal Planning}
\ttfamily
You are planning how a player should complete a task in a game of [GAME].

Task: "[TASK]"

The image is the screen the player is looking at right now.

Here is what has been learned from past playthroughs of this game that may be relevant:\\
(nothing recorded)

Break the task into an ordered sequence of steps, separated by the token [STEP].

Each step will be given to a player who CANNOT see the other steps and does not know how many remain. They see only the current screen and the step you wrote. So each step must stand entirely on its own.

Requirements for every step:

- Describe the step by what is VISIBLE on screen: objects, icons, cursors, doors, characters, menu entries, text. Refer to things the player can point at.\\
- Do NOT name buttons or directions. Write "move the cursor to the coat" rather than "press RIGHT twice to reach the coat", and "select the hand tool" rather than "press A". Which button achieves it is the player's problem, and the button that worked in a past playthrough may be wrong from this screen.\\
- Every step MUST carry a termination condition that is visually checkable -- a state of the screen the player can look at and confirm. Write it as "... until <what the screen shows>". If you cannot name a visible condition that ends the step, the step is too vague: merge it into a neighbour or rewrite it.\\
- One step should be one coherent sub-goal, not a single input and not the whole task.\\
- Prefer few steps. Three or four good steps beat ten brittle ones.

Do not include a step for something the screen shows is already done.

Respond in exactly this format:\\
Plan: <step one, ending in a visible condition> [STEP] <step two, ending in a visible condition> [STEP] <...>\\
{[STOP]}
\end{promptbox}

\begin{promptbox}[label=prompt:subgoal_judgement_window_summary]{Subgoal Judgement: Window Summary}
\ttfamily
You are examining a segment of a player's attempt at one step of a plan in a game of [GAME].

The step they were asked to complete: "[TASK]"

Actions taken in this segment (steps [START\_IDX]-[END\_IDX] of [TOTAL] total):\\
{[ACTION\_SEQUENCE]}

The images show frames [START\_IDX]-[END\_IDX] of the attempt, from left to right.

Describe what visibly changed on screen across this segment, and whether anything in it shows the step's termination condition being met. Report what you can see, not what you assume the player intended.

Respond in exactly this format:\\
Segment summary: <one or two sentences describing what visibly happened>\\
{[STOP]}
\end{promptbox}

\begin{promptbox}[label=prompt:subgoal_judgement_verdict]{Subgoal Judgement: Verdict}
\ttfamily
You are deciding whether a player completed one step of a plan in a game of [GAME].

The step they were asked to complete: "[TASK]"

Summaries of each segment of their attempt:\\
{[SEGMENT\_SUMMARIES]}

The image is the screen as it stands NOW, at the end of the attempt. It is your primary evidence: the step is complete if and only if this screen shows its termination condition met.

The player stopped because: [STOP\_REASON]. Note that a player who declared itself finished may be wrong -- judge the screen, not the claim.

Answering "yes" when the step is not done sends the plan onward from a state it does not expect, and everything after it is built on a false premise. Answering "no" when it is done wastes the step budget repeating work. Judge honestly in both directions.

Respond in exactly this format:\\
Reasoning: <one or two sentences, referring to what is visible in the final screen>\\
Complete: <yes or no>\\
{[STOP]}
\end{promptbox}

\begin{promptbox}[label=prompt:regression_check]{Regression Check}
\ttfamily
You are checking whether a player of [GAME] has undone progress they had already made.

Earlier in this task they completed this step:\\
"[PREVIOUS\_STEP]"

[EARLIER\_STEPS\_BLOCK]You are given two images. The FIRST is the screen at the moment that step was judged complete. The SECOND is the screen now, after a later step was attempted and failed.

What happened in between:\\
{[SEGMENT\_SUMMARIES]}

Compare the two screens. Has the state that made the earlier step complete been lost? Examples of losing it: a tool that was selected is no longer selected, a menu that was open has closed, a door that was opened is shut again, an item that was held has been put back, the player has left the room they had reached.

Judge only what the two images show. Do not guess from the actions described -- if the second screen still shows the earlier step's result, it was not undone, however erratic the play looks. If the images are too similar to tell, say no.

Respond in exactly this format:\\
Reasoning: <one or two sentences comparing the two screens>\\
Undone: <yes or no>\\
What was lost: <if yes, name the specific thing that is no longer true; otherwise write NONE>\\
{[STOP]}
\end{promptbox}

\begin{promptbox}[label=prompt:self_revision]{Self-Revision}
\ttfamily
You are reviewing whether a plan for a task in [GAME] is still worth following.

The overall task: "[OVERALL\_TASK]"

The plan, with progress marked:\\
{[PLAN\_BLOCK]}

The current step has just failed. What happened:\\
{[FAILURE\_HISTORY]}

Why it is judged incomplete: [JUDGEMENT]\\
{[REGRESSION\_LINE]}\\
What past playthroughs of this game recorded:\\
(nothing recorded)

The image is the screen the player is looking at right now.

A plan can fail for two very different reasons, and you are deciding which:

**The plan is sound, the player is fumbling it.** The steps describe the right route; the player misread the screen, pressed the wrong control, or acted on the wrong object. A better hint fixes this. Answer **no**.

**The plan is wrong.** The screens show something the plan did not anticipate: the route it assumes does not exist, an object it names is not there, a step depends on a state that cannot be reached from here, the game works differently from what the plan assumed, or the player is somewhere the plan has no path from. No hint fixes this, because the player is being asked to do the wrong thing. Answer **yes**.

Be strict. Repeated failure alone is not evidence of a bad plan -- a fumbled step fails repeatedly too. You need something visible on the screens that the plan is incompatible with. If you cannot name that thing, answer no.

If you answer yes, write a replacement for the current step and everything after it. Steps already marked DONE are finished and must not be re-planned; start from where the player is now. Keep the original plan's rules: describe what is VISIBLE, never name buttons or directions, and end every step with a visually checkable condition ("... until <what the screen shows>").

Respond in exactly this format:\\
Reasoning: <what on the screens does or does not contradict the plan>\\
Flawed: <yes or no>\\
Plan: <the replacement steps separated by [STEP], or NONE if not flawed>\\
{[STOP]}
\end{promptbox}

\begin{promptbox}[label=prompt:error_correction]{Error Correction}
\ttfamily
You are advising a player of [GAME] who has just failed to complete one step of a plan and is about to try again.

The step: "[TASK]"

Summaries of what they just did:\\
{[SEGMENT\_SUMMARIES]}

Why it is judged incomplete: [JUDGEMENT]

[REGRESSION\_BLOCK]What past playthroughs of this game recorded that may bear on this:\\
(nothing recorded)

[PRIOR\_HINT\_BLOCK]The image is the screen the player is looking at RIGHT NOW. They are NOT starting over -- the game is exactly as this screen shows, including any progress or damage from the failed attempt.

Work in two parts.

**First, diagnose.** Say what is actually going wrong, using the reasons the player gave for each button beside what the frames show happened. Name the mechanism, not the symptom: not "they failed to select the tool" but why the presses that should have selected it did not.

**Then instruct.** Unlike the plan, which describes goals without mentioning controls, your hint names the actual controls: UP, DOWN, LEFT, RIGHT, A, B, START. Say **what each button does towards this goal** -- which one moves the cursor, which one confirms, which one backs out of the menu they are stuck in. Use the recorded knowledge above wherever it names a control or what it does; that is what it is for.

**Do not give a count or a sequence.** Not "press DOWN four times, then A". The player acts one button at a time and looks at the screen again after each one, so a recipe written from this screen is wrong by its second step, and a player following it stops watching the screen. Give them the function of each control and the visible condition that tells them to stop: "DOWN moves the selection down the list -- keep going until KEY1 is the circled entry, then A confirms it."

Requirements:

- The instruction must start from THIS screen. If the failed attempt left the player somewhere unexpected, say which button gets them out of it first.\\
- Correct a false belief explicitly before instructing: "you are two tiles left of the icon, not on it -- RIGHT moves the cursor towards it, and A selects once it is highlighted" beats restating the goal.\\
- Do not repeat an instruction the summaries show already failed. If pressing A did nothing three times, do not say press A; say which button does the thing they were trying to do.\\
- Tie every button to an effect the player can see. A button named without saying what it changes on screen is no more useful than the plan step was.

Respond in exactly this format:\\
Diagnosis: <one or two sentences naming what is actually going wrong>\\
Hint: <which buttons do what towards this goal, and the visible condition to stop at; two sentences at most, no counts>\\
{[STOP]}
\end{promptbox}

\begin{promptbox}[label=prompt:executor_step]{Executor Step}
\ttfamily
Task: [TASK][HINT\_BLOCK]

You are playing a GameBoy game. The current screen is shown in the image.

[ERROR\_BLOCK]Available environment actions:\\
{[ACTION\_LIST]}

[CONTEXT\_SECTION]Reason about the best next action, then respond in exactly this format:\\
Reasoning: <your reasoning>\\
Action: <one environment action>\\
{[STOP]}
\end{promptbox}

\begin{promptbox}[label=prompt:executor_step_hint_block]{Executor Step: Hint Block}
\ttfamily
[STEP\_INFO] You have already taken [N] actions so far in this attempt. Note: this may not be the first step of the overall task and one action does not correspond to one step in the hint plan -- earlier actions may already have been taken before this attempt began, so reason from what you currently see on screen rather than assuming a fresh start. [STEP\_INFO\_END]\\
{[HINT\_START]}\\
Hint: [HINT]\\
Note: This hint block is a secret. You must use it to guide your decision making, but in the reasoning you say, you should pretend as if you actually just know the content of the hint. Do not refer to it explicitly. So if the hint gives you a direction, instead of saying 'the hint says go here', your reasoning should just say 'next I must go here'. [HINT\_END]
\end{promptbox}

\begin{promptbox}[label=prompt:visual_history_change_description]{Visual History: Change Description}
\ttfamily
You are watching someone play the GameBoy game [GAME].

Image 1 is the screen BEFORE they pressed [ACTION]. Image 2 is the screen AFTER.

In one short sentence, say what changed between the two screens as a result of that action. If nothing changed, say exactly: nothing changed.\\
{[STOP]}
\end{promptbox}

\begin{promptbox}[label=prompt:visual_history_context_section]{Visual History: Context Section}
\ttfamily
What your recent actions did (oldest first):\\
\hspace*{1em}[ACTION] -> [CHANGE DESCRIPTION]\\
\hspace*{1em}...

If you have been trying to execute the same action repeatedly (specifically A or B) and especially if you get the [no change] message on your recent actions, consider that you may be stuck in a loop, and should try something else. Look at the screen deeply and use the visual cues to guide your decision making. If you are trying to interact with something, you likely have the incorrect orientation and need to slightly adjust your positioning
\end{promptbox}

\begin{promptbox}[label=prompt:completion_check]{Completion Check}
\ttfamily
Task: [TASK][HINT\_BLOCK]

You are judging whether a task being played on a GameBoy has been FULLY completed.

Image 1 is the screen BEFORE the most recent action. Image 2 is the screen AFTER it.

Most recent action: [LAST\_ACTION]\\
The reasoning given for that action was:\\
{[LAST\_REASONING]}

[HISTORY\_BLOCK]Decide whether the task as stated is now COMPLETELY accomplished -- not partially, not nearly, not "the next step is obvious". If any part of the task remains to be done, the answer is no. If you cannot tell from what is visible, the answer is no.

Respond in exactly this format, with the verdict FIRST:\\
Complete: <yes or no>\\
Reasoning: <why, referring to what is visible in image 2>\\
{[STOP]}
\end{promptbox}

\begin{promptbox}[label=prompt:executor_step_scored_variant]{Executor Step: Scored Variant}
\ttfamily
Score each available action on how useful it would be RIGHT NOW for making progress toward the task (1=useless, 5=very useful), and justify each score in one sentence.

Respond with one line per action in exactly this format:\\
<action>: <one-sentence reason for the score>: <score>\\
...\\
{[STOP]}
\end{promptbox}

\begin{promptbox}[label=prompt:executor_step_sequence_variant]{Executor Step: Sequence Variant}
\ttfamily
Plan a short sequence of actions (1-5) to make progress on the task. Respond in exactly this format:\\
Reasoning: <your reasoning>\\
Action: <ACTION1, ACTION2, ...>\\
{[STOP]}
\end{promptbox}

\begin{promptbox}[label=prompt:executor_step_action_history_variant]{Executor Step: Action History Variant}
\ttfamily
Recent actions (oldest first):\\
\hspace*{1em}[ACTION]\\
\hspace*{1em}[ACTION] [no change]\\
\hspace*{1em}...
\end{promptbox}

\subsection{World Modelling}

\begin{promptbox}[label=prompt:world_model_executor_step]{World Model Executor Step}
\ttfamily
Task: [TASK][HINT\_BLOCK]

You are playing a GameBoy game.

[ERROR\_BLOCK]Image 1 is the CURRENT screen.

The remaining images are predictions from a learned world model of what the screen would look like after each available action. They are reconstructions, so they are blurry and imperfect -- judge them on the overall change they show, not on fine detail.

Predicted outcomes:\\
\hspace*{1em}Image 2: after taking [ACTION 1]\\
\hspace*{1em}Image 3: after taking [ACTION 2]\\
\hspace*{1em}...

Pick the single predicted screen that most directly advances the task. Respond in exactly this format:\\
Reasoning: <your reasoning>\\
Choice: <the image number you pick>\\
{[STOP]}
\end{promptbox}

\subsection{Autonomous Skill Discovery}

\begin{promptbox}[label=prompt:task_proposal]{Task Proposal}
\ttfamily
You are observing the initial frame of a game of [GAME].

You are an expert game analyst. Your job is to look carefully at this frame and reason about:\\
- What is visible in the immediate surroundings (objects, NPCs, structures, interactables)\\
- What areas or items are accessible from this position\\
- What mechanisms or interactions are available given the action space: [ACTION\_SPACE]

Propose an exhaustive list of distinct tasks that a player could meaningfully attempt to achieve starting from this exact state. Focus on tasks that are:\\
- Grounded in what is actually visible or reachable from this state\\
- Specific and concrete (not vague like "explore the area")\\
- Achievable as a single coherent goal\\
- Varied in scope (include both short and longer-horizon tasks)

Respond in exactly this format:\\
Reasoning: <brief analysis of what is visible and what interactions are possible>\\
Tasks:\\
- <task 1>\\
- <task 2>\\
- <task 3>\\
...\\
{[STOP]}
\end{promptbox}

\begin{promptbox}[label=prompt:attempt_judgement_window_description]{Attempt Judgement: Window Description}
\ttfamily
You are watching frames [START\_IDX]-[END\_IDX] of [TOTAL] total frames from a game of [GAME].

Describe what the player does and what changes visually in this segment. Focus on actions taken and their outcomes. Do not assume any particular goal.

Respond in exactly this format:\\
Description: <concise description of the player's actions and visual changes in this segment>\\
{[STOP]}
\end{promptbox}

\begin{promptbox}[label=prompt:attempt_judgement_consolidation]{Attempt Judgement: Consolidation}
\ttfamily
You are consolidating segment descriptions from a game of [GAME] into a single complete trajectory description.

Segment descriptions (in chronological order), each labelled with the frame range it covers:\\
{[SEGMENT\_DESCRIPTIONS]}

Produce a single coherent description of the full trajectory from start to finish. Explicitly reference the frame ranges (e.g. "frames 1-10", "frames 11-20") as you describe what happens, so the reader can tell which part of the trajectory each event belongs to. Keep these frame-range references in the same form they appear in the segment labels above.

Respond in exactly this format:\\
Description: <complete description of the full trajectory, with frame ranges referenced inline>\\
{[STOP]}
\end{promptbox}

\begin{promptbox}[label=prompt:attempt_judgement_verdict]{Attempt Judgement: Verdict}
\ttfamily
Task: "[TASK]"

A player attempted to complete this task. Here is a description of what happened across the FULL trajectory:\\
"[DESCRIPTION]"

The images show only the FINAL frames of the trajectory. Task completion may have occurred earlier and may not be visible in these images.

Did the player successfully complete the task at any point during the trajectory? Use the description as your primary evidence -- if it mentions something that closely matches task completion, count it as success even if it is not visible in the final frames shown.\\
{[GOAL\_CONDITION\_NOTE]}\\
The description references frame ranges (e.g. "frames 11-20"). Using these, identify the safe success point: the single frame number by which the task has SURELY been achieved. Pick the earliest frame you are confident the task is already complete. If the task was never completed, or you cannot tell from the description, respond with N/A.

Respond in exactly this format:\\
Reasoning: <your reasoning, referencing the description and any visual evidence>\\
Success: <yes or no>\\
Safe success point: <frame number, or N/A if never completed or unknown>\\
{[STOP]}
\end{promptbox}

\begin{promptbox}[label=prompt:attempt_critique_window_summary]{Attempt Critique: Window Summary}
\ttfamily
You are analysing a segment of a failed attempt to complete a task in a game of [GAME].

Task: "[TASK]"

Actions taken in this segment (steps [START\_IDX]-[END\_IDX] of [TOTAL] total):\\
{[ACTION\_SEQUENCE]}

The images show frames [START\_IDX]-[END\_IDX] of the trajectory, from left to right.

Describe what happened in this segment: what the player did, what went wrong (if anything), and any observations relevant to why the task was not completed.

Respond in exactly this format:\\
Segment summary: <one or two sentences describing what happened in this segment>\\
{[STOP]}
\end{promptbox}

\begin{promptbox}[label=prompt:attempt_critique_hint]{Attempt Critique: Hint}
\ttfamily
You are analysing a failed attempt to complete a task in a game of [GAME].

Task: "[TASK]"

Below are summaries of each segment of the failed trajectory:\\
{[SEGMENT\_SUMMARIES]}

[PRIOR\_HINT\_BLOCK]Based on the full trajectory above, provide a concise hint for how to better approach the task on the next attempt.

Respond in exactly this format:\\
Critique: <what went wrong overall>\\
Hint: <one or two sentence hint for a better approach>\\
{[STOP]}
\end{promptbox}

\begin{promptbox}[label=prompt:guidance_inference_window]{Guidance Inference: Window}
\ttfamily
You are an expert player of [GAME]. You are given frames [START\_IDX]-[END\_IDX] (out of [TOTAL] total frames) from a game trajectory, along with the task being accomplished:

Task: "[TASK]"

Describe what the player does in this section of the trajectory. Base your guidance strictly on what you can observe -- do not invent details not visible.

Guidelines:\\
- Write each step as a short, clear imperative instruction (e.g. "Press A to speak to the NPC", "Walk left towards the door").\\
- Order steps chronologically within this section.\\
- Be specific about directions, button presses and targets when clearly visible.\\
- If the exact button labels are intuitive, describe the action instead.\\
- You must make sure that for each step, you explicitly refer to the visual cues in the frame that indicate WHEN to do a particular step, and when to move on to the next. For example, instead of saying "walk upwards", say "when the character is past the tree line, walk upwards until the door is visible at the top of the frame." Be very specific about the visual cues that indicate what to do and when.

Respond in exactly this format:\\
Summary: <one sentence describing what happens in frames [START\_IDX]-[END\_IDX]>\\
Goal condition: <a visual description of the final frame in this section>\\
Steps:\\
- <step 1>\\
- <step 2>\\
...\\
{[STOP]}
\end{promptbox}

\begin{promptbox}[label=prompt:guidance_inference_consolidation]{Guidance Inference: Consolidation}
\ttfamily
You are consolidating partial guidance from multiple sections of a trajectory in a game of [GAME].

Task: "[TASK]"

Here are the descriptions of each section, in chronological order:\\
{[SECTION\_DESCRIPTIONS]}

Combine these into a single, complete, coherent set of step-by-step guidance that a new player could follow from start to finish to accomplish the task.

Respond in exactly this format:\\
Summary: <one sentence describing the overall approach to complete the task>\\
Goal condition: <a visual description of the state that confirms task completion>\\
Steps:\\
- <step 1>\\
- <step 2>\\
...\\
{[STOP]}
\end{promptbox}

\begin{promptbox}[label=prompt:decision_filtering]{Decision Filtering}
\ttfamily
You are reviewing a single decision an agent made while playing [GAME].

The agent was working toward this task:\\
"[TASK]"

It was shown the game frame(s) provided as image(s) (every image EXCEPT the last) and given this prompt context:\\
{[AGENT\_PROMPT]}

The agent's reasoning and action (its response) was:\\
{[AGENT\_RESPONSE]}

The FINAL image is the game frame AFTER the agent's action was executed. Use the change from the agent's frame(s) to this resulting frame as your main evidence of whether the action helped.

Your job is to catch only CRITICAL errors. Consider:\\
- Is the agent's description of the frame clearly wrong given the visual evidence?\\
- Does the action's visible effect (the before -> after change) clearly fail to advance the task, or move away from it?

Err strongly on the side of ACCEPT. Only REJECT if there is clear visual evidence that the response badly misdescribes the frame, or that the resulting transition shows the action was counterproductive or unlikely to advance the task. If you are unsure, ACCEPT.

Respond in exactly this format:\\
Reason: <one short sentence justifying your decision>\\
Decision: <ACCEPT or REJECT>\\
{[STOP]}
\end{promptbox}

\begin{promptbox}[label=prompt:paraphrasing]{Paraphrasing}
\ttfamily
You are given a canonical task string:\\
"[CORE\_TASK]"

Generate at least [N] diverse, valid paraphrases of this task. Vary the wording, phrasing style, and structure but preserve the exact core meaning and level of specificity. Use imperative tone throughout.

Respond in exactly this format (one paraphrase per line):\\
- <paraphrase 1>\\
- <paraphrase 2>\\
...\\
{[STOP]}
\end{promptbox}

\subsection{Insight documents}

\begin{promptbox}[label=prompt:task_inference]{Task Inference}
\ttfamily
You are analysing multiple screenshots in sequence from a game of [GAME].

Over the course of some of these frames, a single primary task may have been performed by the player, with the task being completed either at the very end or in some frame close to the end.\\
Describe, with a single phrase, the action or task the player performed over the course these frames? Do not use conjunctions like "and" or "while" in your description. If there are multiple distinct tasks that seem to be happening, try to describe the whole subtrajectory wholistically and omit the less important subtasks. If there is no clear task, say "NO TASK".\\
Be specific but concise, each task should be a single, specific and meaningful action and not trivial. Describe only what is clearly supported by the evidence above. Make the task description as unambiguous as possible -- include enough distinguishing detail that it cannot be confused with other similar tasks that could occur in the same game.

Always try to pick the longest horizon, most multistep version of the task that is present in the trajectory. If there is no clear task, respond with "NO TASK"\\
Otherwise, respond in exactly this format:\\
Visual Description: <a description of the individual frames and changes that occur from leftmost frame to rightmost frame>\\
Reasoning: <one single, short sentence describing your thinking. Reference visual evidence of the key frames and overall actions that led you to infer this task.>\\
Task: <one or two sentence description of what the player did or is doing>\\
Start: <integer index of starting frame> this is to indicate when the player seems to be acting with the intent to perform the task, not simply the step right before the task is executed. This may be several frames before the task is completed, and will depend on the specific task and context.\\
End: <integer index of frame where task is performed or executed or completed or first detected>\\
{[STOP]}
\end{promptbox}

\begin{promptbox}[label=prompt:task_validation_and_rewriting]{Task Validation and Rewriting}
\ttfamily
You are given a description of what a player did over the course of some game frames:\\
"[CANDIDATE\_TASK]"

First judge whether this is a well-formed task. A VALID task is SPECIFIC and CONCRETE -- a clearly-defined action with an unambiguous completion state that could be checked from the screen (e.g. "select the diamond from the inventory", "open the cellar door", "exit the taxi"). Judge it INVALID if it is too generic or vague to complete exactly -- e.g. "navigate the game environment", "interact with an object", "explore the area", "manage inventory", "pick up an object" -- cases where many different behaviours would all satisfy it.

If VALID, rewrite it as a concise imperative instruction:\\
- Use second-person imperative tone (no subject).\\
- Keep it short (under 10 words if possible).\\
- Do not add any detail that was not in the original description.

Respond in exactly this format:\\
Verdict: <VALID or INVALID>\\
Task: <imperative task string if VALID, or NONE if INVALID>\\
{[STOP]}
\end{promptbox}

\begin{promptbox}[label=prompt:group_task_distillation]{Group Task Distillation}
\ttfamily
You are given several candidate descriptions of a task performed in a game of [GAME], all inferred from similar game states:

[CANDIDATE\_LIST]

Generate a single, unifying task string that captures the core commonality between all of these candidates, while omitting any extraneous detail or noise. Use imperative tone.

The distilled task must stay SPECIFIC, CONCRETE and UNAMBIGUOUS -- a clearly-defined action with a checkable completion state (e.g. "select the diamond from the inventory", "open the cellar door"). Do not over-generalise into a vague or generic instruction (e.g. "interact with an object", "navigate the environment", "manage inventory") that many different behaviours would satisfy. Keep the most specific meaning shared by the candidates.

Respond in exactly this format:\\
Reasoning: <one single, short sentence describing your thinking. Reference the commonalities between the candidates that led you to infer this distilled task.>\\
Task: <single distilled imperative task string>\\
{[STOP]}
\end{promptbox}

\begin{promptbox}[label=prompt:insight_extraction]{Insight Extraction}
\ttfamily
You are analysing a gameplay trajectory from [GAME] to build a knowledge base for future players.

Task attempted: "[TASK]"\\
Actions taken, in order: [ACTIONS]

The images are frames sampled from the run, in chronological order.

Extract only the KEY, NON-OBVIOUS insights -- the things that would genuinely help someone attempting a SIMILAR task in this game in future. Good insights are things like:\\
- A game mechanic that is not visible from a single screen (what triggers an interaction, what a state change means).\\
- Exactly when to press which button, and the visual cue that tells you it is time.\\
- A positioning or orientation requirement that is easy to get wrong.\\
- A failure mode and how to avoid it.

Do NOT state the obvious. "Press the direction you want to walk", "press START to open the menu" ,"the player must reach the goal", "press A to interact or speak to an NPC or pick up an object, Press B to go close a dialogue or menu" with no further condition -- these are worthless. If nothing non-obvious can be learned from this trajectory, reply with exactly NONE on the Insights line.

Every insight must be concrete and actionable: name the actual button, the actual visual cue, the actual object, the actual condition.

You must also categorise this trajectory twice:\\
1. A TASK category -- the general kind of task this is, so that similar tasks group under it. Short and reusable (e.g. "open a locked container", "talk to an NPC to receive an item"), not specific to this one instance.\\
2. An IMAGE category -- the kind of screen the run STARTS on, based on the first frame. Short and reusable (e.g. "top-down interior room", "dialogue box", "inventory menu"). If the starting screen is not distinctive enough to be worth naming, reply NONE for it.

Respond in exactly this format:\\
Task category: <short reusable name>\\
Task description: <one line: what this category of task is, in general>\\
Task example: <one line: what was attempted here and what actually happened>\\
Image category: <short reusable name, or NONE>\\
Image description: <one line: how to recognise this kind of screen from what is visible, or NONE>\\
Image example: <one line: what is on this particular screen, or NONE>\\
Insights:\\
- <non-obvious, concrete, actionable insight>\\
- <another>\\
{[STOP]}
\end{promptbox}

\begin{promptbox}[label=prompt:document_merging_entry_matching]{Document Merging: Entry Matching}
\ttfamily
You are consolidating a knowledge base for the game [GAME] and must decide whether a new entry teaches something already in it.

Here is the NEW entry, with its insights:\\
{[CANDIDATE]}

Here are the EXISTING entries, numbered, with their insights:\\
{[EXISTING]}

The images are: first the NEW entry's representative frame, then the representative frame of each existing entry in the order listed above.

Decide primarily on the INSIGHTS. Two entries match when their insights say the same thing, or say things so close that one consolidated list would be strictly better than two separate ones -- the same mechanic, the same cue, the same button at the same moment, the same failure mode. The category, description, examples and frames are supporting evidence for that judgement, not the judgement itself: two entries can carry different category names and still match if the insights teach the same lesson, and two entries can share a category name and NOT match if their insights are about different mechanics.

Default to NONE. Answer with a number only when the insights are highly similar -- essentially the same lesson about the same [KIND], not merely related or adjacent. If you are weighing whether two insight lists are close enough, they are not: say NONE.

Merging entries whose insights are genuinely different produces a mushy, useless entry, and a wrong match is far more damaging than a missed one.

Respond in exactly this format:\\
Reasoning: <one or two sentences, comparing the insights and referring to the frames>\\
Match: <the number of the matching entry, or NONE>\\
{[STOP]}
\end{promptbox}

\begin{promptbox}[label=prompt:document_merging_insight_combination]{Document Merging: Insight Combination}
\ttfamily
You are merging what has been learned about the same [KIND] in the game [GAME], from two separate observations.

The entry is: [CATEGORY] -- [DESCRIPTION]

Insight list A:\\
{[INSIGHTS\_A]}

Insight list B:\\
{[INSIGHTS\_B]}

Combine these into a SINGLE consolidated list of insights. Decide for yourself which points are the same point stated twice (merge them), which refine or qualify each other (state the refined version), and which are independent (keep both).

CRITICAL -- never let a combined insight become generic, vague, or non-actionable. Every insight must stay concrete and specific enough to act on: name the actual button, the actual visual cue, the actual object, the actual condition. If two insights are specific in DIFFERENT ways, keep both as separate points rather than generalising them into one weaker statement. Do not soften a precise claim into a vague one. Preferring two sharp insights over one blurred insight is always correct.

Never drop a point just because the list is getting long. A longer list of sharp insights is better than a short list of blunt ones.

Respond in exactly this format:\\
Insights:\\
- <insight>\\
- <insight>\\
{[STOP]}
\end{promptbox}

\begin{promptbox}[label=prompt:test_time_usage_relevance_judgement]{Test-Time Usage: Relevance Judgement}
\ttfamily
You are deciding whether a piece of recorded knowledge about [GAME] is relevant to the situation a player is in right now.

The player's current task is: "[TASK]"

Here is the recorded entry:\\
{[ENTRY]}

The images are: first the CURRENT screen the player is looking at, then the representative frame recorded with this entry.

Could this entry's knowledge be relevant to the player's current task on this current screen? Answer yes only if the entry genuinely fits the situation -- the same or a very similar [KIND]. The frames are your primary evidence: compare what is actually visible in them.

Answering yes to something that does not fit produces a misleading hint, which is worse than no hint at all. Answering no to something that does fit wastes knowledge that was already paid for. Judge honestly in both directions.

Respond in exactly this format:\\
Reasoning: <one or two sentences, referring to the frames>\\
Relevant: <yes or no>\\
{[STOP]}
\end{promptbox}

\begin{promptbox}[label=prompt:test_time_usage_insight_filtering]{Test-Time Usage: Insight Filtering}
\ttfamily
You are pruning recorded knowledge about [GAME] down to what could matter for one task.

The player's task is: "[TASK]"

The image is the screen they are looking at right now.

Here is everything recorded from past playthroughs that was retrieved for this situation:\\
{[CANDIDATES]}

Some of it will be about things that have nothing to do with this task or this place -- advice about talking to a character when nobody is here, about a menu that this task never opens, about a room the player is not in and will not enter. That is what you are removing.

Keep an item if it could plausibly matter at ANY point while doing this task, not only on the screen as it looks this instant. The player will move, open menus and change rooms while working, and knowledge about where they are heading is exactly what is worth keeping. Something you drop is gone for the whole task.

So: drop only what is clearly about something absent and unrelated. **If you are unsure, KEEP it.** Removing one useful item costs more than leaving three useless ones.

Respond in exactly this format:\\
Keep: <comma-separated numbers, or ALL>\\
{[STOP]}
\end{promptbox}

\begin{promptbox}[label=prompt:test_time_usage_insight_distillation]{Test-Time Usage: Insight Distillation}
\ttfamily
You are consolidating what is known about [GAME] into a briefing for one task.

The player's task is: "[TASK]"

The image is the screen they are looking at right now.

Here is the knowledge kept from past playthroughs. It was recorded piecemeal, by different runs, so it repeats itself, contradicts itself in places, and states the same thing at several levels of detail:\\
{[CANDIDATES]}

Rewrite it as a short, ordered list of concrete statements.

- **Aggregate.** Where several items describe one thing, merge them into a single statement that carries every specific detail any of them had. Prefer the most specific version: if one says "an icon in the toolbar" and another says "the third icon from the left", the merged statement says the third icon from the left.\\
- **Cut redundancy.** Two items that say the same thing become one. An item that is a vaguer restatement of another is dropped entirely.\\
- **Be concrete.** Name the object, the place on the screen, the observable result. Drop anything that survives only as generic advice -- "be careful", "explore thoroughly", "pay attention to the surroundings" -- that is not knowledge, it is filler.\\
- **Stay faithful.** Every statement must be supported by the items above. Do not add knowledge, do not resolve a contradiction by inventing a third version, and do not promote a guess into a fact. If two items genuinely disagree, say so in one statement and keep both readings.\\
- Do not narrow a statement to only what is on this screen. The player will move and change rooms while doing this task, and knowledge about where they are going still belongs here.

Respond in exactly this format, one statement per line:\\
Insights:\\
- <statement>\\
- <statement>\\
{[STOP]}
\end{promptbox}

\subsection{Playthrough Agent: Strategist}
\label{app:prompts_strategist}

The strategist's prompts are grouped by the artifact they read from or write to, following
Appendix~\ref{app:playthroughagent}.

\subsubsection{Episode Level}

\begin{promptbox}[label=prompt:playthrough_dispatch]{Strategist: Dispatch}
\ttfamily
You are playing [GAME] and are the strategist: you decide what is worth doing next, and hand one short task at a time to a player who does the actual playing.

{[RECENT\_BLOCK]}\\
The goal you are working towards, from the broadest goal down to it:\\
{[GOAL]}

The player is at: [CURRENT]

Your notepad:\\
{[NOTEPAD]}

The image is the current game screen. Here is what has been recognised on it:\\
{[SCREEN\_TILES]}

Write one task for the player that moves them towards the goal. The player is competent but has no memory of anything before this moment, gets roughly [MAX\_STEPS] actions, and can only walk, talk to people, use menus and fight battles. So the task must be something reachable from this screen in that budget: "walk into the Pokemon Center and heal your team", not "beat the next gym".

Then write the guidance: everything the player needs and could not work out from the screen. Where the thing they want is, what has already been tried and failed, what to avoid. Do not repeat the task in it.

Respond in exactly this format:\\
Reasoning: <where the player is and what the next concrete step is>\\
Task: <one sentence, concrete, achievable from this screen>\\
Guidance: <one to four sentences for the player, or None>\\
{[STOP]}
\end{promptbox}

\begin{promptbox}[label=prompt:playthrough_review]{Strategist: Episode Review}
\ttfamily
You are playing [GAME] and are the strategist. One episode has just finished.

Here is the player's report of the episode:\\
{[DIGEST]}

Write one line summarising the episode for your own log.

Respond in exactly this format:\\
Reasoning: <what actually happened and what it means>\\
Summary: <one sentence>\\
{[STOP]}
\end{promptbox}

\subsubsection{Goal Tree}

\begin{promptbox}[label=prompt:playthrough_subgoal_query]{Goal Tree: Subgoal Query}
\ttfamily
You are playing [GAME] and are the strategist: you break long-term goals down into smaller subgoals that a player can work through one at a time.

The goal you are planning for, from the broadest goal down to it:\\
{[GOAL]}\\
{[EXTRAS]}\\
Your notepad:\\
{[NOTEPAD]}

Before planning, you can look things up in your database of what you know about the game. Write the questions whose answers would help you decide how this goal should be broken down.

Each question is looked up on its own, in a database that does not know where you are, what you are doing or what has just happened. So every question must make sense to someone who knows the game but knows nothing about your situation: name the places, people, items and Pokemon it is about, and never write "here", "this town", "the next gym" or "right now". Keep each question specific and about one thing.\\
{[ROUTE\_OPTION]}\\
If the plan is already clear, or looking things up would not help, write None instead.

Respond in exactly this format:\\
Reasoning: <what you would need to know to plan this goal>\\
Query: <one self-contained question>\\
Query: <one self-contained question>[ROUTE\_FORMAT]\\
([LINE\_KINDS], or instead the single line `Queries: None`)\\
{[STOP]}
\end{promptbox}

\begin{promptbox}[label=prompt:playthrough_subgoal_create]{Goal Tree: Subgoal Creation}
\ttfamily
You are playing [GAME] and are the strategist: you break long-term goals down into smaller subgoals that a player can work through one at a time.

The goal you are planning for, from the broadest goal down to it:\\
{[GOAL]}\\
{[EXTRAS]}\\
Your notepad:\\
{[NOTEPAD]}

What you looked up in your database of what you know about the game:\\
{[KNOWLEDGE]}

The player is at: [CURRENT]

Routes you looked up on your map:\\
{[ROUTES]}

Places on your map that may lead somewhere you have not been yet:\\
{[EXPLORE]}

Routes are only true from where the player is now, and the player will move while working through the subgoals. So describe each subgoal by where to get to and what to do there, not by copying the steps of a route.

Decide whether this goal needs to be broken down. If it is one clear objective the player can go straight for, write None. Otherwise, list the subgoals that together achieve it, in the order they should be done. Each subgoal is something to achieve in the game, not a button to press.

For each subgoal, give a short description, any details that would help achieve it (or None), and why it is needed for the goal. Separate the three with `|`, and do not use `|` anywhere else. No two subgoals may have the same description, and none may repeat the goal itself or one of its existing subgoals.

Respond in exactly this format:\\
Reasoning: <what the goal needs, and whether it must be broken down>\\
Subgoal: <description> | <details, or None> | <why it is needed>\\
Subgoal: <description> | <details, or None> | <why it is needed>\\
(one `Subgoal:` line per subgoal, or instead the single line `Subgoals: None`)\\
{[STOP]}
\end{promptbox}

\begin{promptbox}[label=prompt:playthrough_goal_check]{Goal Tree: Completion Check}
\ttfamily
You are playing [GAME] and are the strategist. One episode has just finished.

The goal to check, from the broadest goal down to it:\\
{[GOAL]}

Your notepad:\\
{[NOTEPAD]}

Here is the player's report of the episode:\\
{[DIGEST]}

Has this goal been achieved, in this episode or before it? Answer Yes only if the report or your notepad clearly shows it has happened.

Respond in exactly this format:\\
Reasoning: <the evidence for and against>\\
Achieved: <Yes or No>\\
{[STOP]}
\end{promptbox}

\begin{promptbox}[label=prompt:playthrough_goal_decision]{Goal Tree: Goal Decision}
\ttfamily
You are playing [GAME] and are the strategist. One episode has just finished, and this goal has not been achieved yet:\\
{[GOAL]}

Your notepad:\\
{[NOTEPAD]}

Here is the player's report of the episode:\\
{[DIGEST]}

Decide what to do with the goal:\\
CONTINUE: the goal is still right and is on its way; it just needs longer. Nothing changes.\\
UPDATE: the goal is roughly right but aimed at the wrong thing or described wrongly. Rewrite it.[ABANDON\_OPTION]

Respond in exactly this format:\\
Reasoning: <how the goal is going, and what should happen to it>\\
Choice: <one of [CHOICES]>\\
New Description: <the rewritten goal if UPDATE, otherwise None>\\
New Details: <the rewritten details if UPDATE, or None>\\
Reason: <why you chose to update or abandon it, or None if CONTINUE>\\
{[STOP]}
\end{promptbox}

\subsubsection{Knowledge Tree}

\begin{promptbox}[label=prompt:playthrough_knowledge_extract]{Knowledge Tree: Fact Extraction}
\ttfamily
You are playing [GAME] and are the strategist. One episode has just finished, and you keep a database of what you know about the game.

Here is the player's report of the episode:\\
{[DIGEST]}

The first image is the screen when the episode started, the second is the screen when it ended.

List the facts from this episode that are worth remembering for the rest of the game: what people said, where places are and how they connect, what items and Pokemon were found or used, what worked, what failed and why, and what the story now asks of the player. Leave out the play-by-play of what the player pressed, and anything that will not matter later.

Each fact is stored on its own, away from this report, so it must make sense without it: name the people, places and things it is about rather than saying "he", "there" or "it".

Write at most [MAX\_FACTS] facts, one per line. If nothing is worth remembering, write None.

Respond in exactly this format:\\
Reasoning: <what this episode taught you>\\
Facts:\\
- <fact>\\
- <fact>\\
{[STOP]}
\end{promptbox}

\begin{promptbox}[label=prompt:playthrough_knowledge_navigate]{Knowledge Tree: Navigation}
\ttfamily
You are playing [GAME] and keep a database of what you know, organised as a tree of topics. Here is some knowledge you wish to remember:\\
{[KNOWLEDGE]}

You may already have this in your database. First, navigate to the place where it is best stored.

You are at: [PATH]\\
{[DESCRIPTION]}

The entries under it:\\
{[CHILDREN]}

Choose the entry this knowledge belongs in, or NEW if none of them fit and it needs a new entry here.

Respond in exactly this format:\\
Reasoning: <which entry fits and why>\\
Choice: <the number of an entry, or NEW>\\
{[STOP]}
\end{promptbox}

\begin{promptbox}[label=prompt:playthrough_knowledge_leaf]{Knowledge Tree: Leaf Decision}
\ttfamily
You are playing [GAME] and keep a database of what you know, organised as a tree of topics. Here is some knowledge you wish to remember:\\
{[KNOWLEDGE]}

The closest entry in your database is:\\
{[PATH]}\\
{[DESCRIPTION]}

Decide what to do with it:\\
KNOWN: the entry already says this. Nothing changes.\\
UPDATE: the entry is about the same thing but is missing this knowledge or gets it wrong. Rewrite its description so it includes the new knowledge and keeps everything still true.\\
NEW: the knowledge is about something else. It gets its own entry next to this one.

Respond in exactly this format:\\
Reasoning: <how the knowledge compares to the entry>\\
Choice: <KNOWN, UPDATE or NEW>\\
New Description: <the full rewritten description if UPDATE, otherwise None>\\
{[STOP]}
\end{promptbox}

\begin{promptbox}[label=prompt:playthrough_knowledge_create]{Knowledge Tree: Entry Creation}
\ttfamily
You are playing [GAME] and keep a database of what you know, organised as a tree of topics. Here is some knowledge you wish to remember:\\
{[KNOWLEDGE]}

It will be stored under: [PATH]\\
{[DESCRIPTION]}

The entries already there:\\
{[SIBLINGS]}

Write the new entry. If it belongs directly under [LAST], write None for both parent lines. If it is the first of a group of related entries that deserves its own heading, give that heading as the parent and the entry will go under it.

No title may repeat one of the entries already there.

Respond in exactly this format:\\
Reasoning: <what the knowledge is and where it fits>\\
Parent Title: <a short heading, or None>\\
Parent Description: <one sentence on what the heading covers, or None>\\
Entry Title: <a short title>\\
Entry Description: <the knowledge itself, distilled to what is worth remembering>\\
{[STOP]}
\end{promptbox}

\begin{promptbox}[label=prompt:playthrough_knowledge_recall]{Knowledge Tree: Recall}
\ttfamily
You are playing [GAME] and keep a database of what you know, organised as a tree of topics. You want to recall what you know about:\\
{[QUERY]}

You are at: [PATH]\\
{[DESCRIPTION]}

The entries under it:\\
{[CHILDREN]}

Choose up to [MAX\_BRANCHES] entries that are most likely to hold something relevant, or NOT FOUND if none of them could.

Respond in exactly this format:\\
Reasoning: <which entries could be relevant and why>\\
Choice: <the numbers of at most [MAX\_BRANCHES] relevant entries, separated by commas, or NOT FOUND>\\
{[STOP]}
\end{promptbox}

\subsubsection{Location Store}

\begin{promptbox}[label=prompt:playthrough_locate]{Location Store: Locate}
\ttfamily
You are playing [GAME] and are the strategist. You keep a map of the places the player has been, and you do not currently know where the player is.

The image is the current game screen. Here is what has been recognised on it:\\
{[SCREEN\_TILES]}

Places come in three kinds:\\
Major: a town, city or route, the places the game's own town map names.\\
Bridge: a place whose purpose is to connect major locations, such as a gatehouse between a route and a town, a cave or a tunnel.\\
Internal: a room inside a major location, such as a Pokemon Center, a shop, a gym or a house. Each floor of a building is its own room. Give the major location it is inside as its parent.

The places you already know:\\
{[KNOWN]}

If the place is one of these, use its name and kind exactly as written. Otherwise give it the name the game uses for it, or a short descriptive name if the game does not show one. If the screen does not show a place at all (a battle, a menu, dialogue covering the view, a black screen), write Unknown.

Respond in exactly this format:\\
Reasoning: <what the screen shows and which place it is>\\
Kind: <Major, Bridge, Internal or Unknown>\\
Name: <the place's name, or None if Unknown>\\
Parent: <for Internal, the major location it is inside; otherwise None>\\
{[STOP]}
\end{promptbox}

\begin{promptbox}[label=prompt:playthrough_transition_check]{Location Store: Transition Check}
\ttfamily
You are playing [GAME] and are the strategist. While the player was playing, the screen went black for a moment. The first [N\_BEFORE] images are the screen before it went black, oldest first, and the last [N\_AFTER] images are the screen after, oldest first.

Decide whether the player moved to a different place: through a door into or out of a building, up or down stairs, into or out of a cave or gatehouse, or to another town or route.

Only a move to a different place counts. These do not: a battle starting or ending, a menu, the Pokemon or item screens opening or closing, dialogue, a screen flash, or a cutscene that ends back where it started. If the screen after the black shows the same place as before it, answer No.

Respond in exactly this format:\\
Reasoning: <what the screens before and after show>\\
Transition: <Yes or No>\\
{[STOP]}
\end{promptbox}

\begin{promptbox}[label=prompt:playthrough_transition_name]{Location Store: Transition Naming}
\ttfamily
You are playing [GAME] and are the strategist. You keep a map of the places the player has been. The player has just moved from one place to another, and the screen went black in between. The first [N\_BEFORE] images are the screen before, oldest first, and the last [N\_AFTER] images are the screen after, oldest first.

The player was in: [PREVIOUS]\\
Places already known to connect to it: [CONNECTED]

Places come in three kinds:\\
Major: a town, city or route, the places the game's own town map names.\\
Bridge: a place whose purpose is to connect major locations, such as a gatehouse between a route and a town, a cave or a tunnel.\\
Internal: a room inside a major location, such as a Pokemon Center, a shop, a gym or a house. Each floor of a building is its own room. Give the major location it is inside as its parent.

The places you already know:\\
{[KNOWN]}

If the place is one of these, use its name and kind exactly as written. Otherwise give it the name the game uses for it, or a short descriptive name if the game does not show one. If the screen after does not show enough to tell where the player is, write Unknown.

Then say how the player got from the previous place to this one, and whether the same way leads back.

Respond in exactly this format:\\
Reasoning: <what the screens show and which place the player is in now>\\
Kind: <Major, Bridge, Internal or Unknown>\\
Name: <the place's name, or None if Unknown>\\
Parent: <for Internal, the major location it is inside; otherwise None>\\
How: <how the player got here from the previous place, e.g. "through the door at the top of the building", or None>\\
Two-way: <Yes if the same way leads back, No if it does not or you cannot tell>\\
{[STOP]}
\end{promptbox}

\begin{promptbox}[label=prompt:playthrough_navigate_select]{Location Store: Route Selection}
\ttfamily
You are playing [GAME] and keep a map of the places you have been. You want to go to:\\
{[REQUEST]}

You are at: [CURRENT]

The places on your map, nearest first:\\
{[PLACES]}

Choose every place that fits where you want to go. The nearest one you know a route to will be used. If none of them fit, write None.

Respond in exactly this format:\\
Reasoning: <which places fit and why>\\
Choice: <the numbers of the places that fit, separated by commas, or None>\\
{[STOP]}
\end{promptbox}

\subsubsection{Notepad}

\begin{promptbox}[label=prompt:playthrough_notepad_update]{Notepad: Update}
\ttfamily
You are playing [GAME] and are the strategist. You keep a notepad of working thoughts about the goal you are pursuing: what has been tried, what worked, what failed, and what to try next. One episode has just finished.

The goal you are working towards, from the broadest goal down to it:\\
{[GOAL]}

The player is at: [CURRENT]

Your notepad:\\
{[NOTEPAD]}

Here is the player's report of the episode:\\
{[DIGEST]}

Your one-line summary of the episode: [SUMMARY]

Update the notepad with what this episode taught you about reaching the goal. Choose one:\\
APPEND: add new notes to the end of the notepad, keeping everything already there.\\
REWRITE: replace the whole notepad, when parts of it are now wrong, out of date or repeated. Keep everything that is still true and useful.

Keep the notes short and useful for the episodes to come. Do not narrate the episode step by step.

Respond in exactly this format:\\
Reasoning: <what this episode changes about your notes>\\
Action: <APPEND or REWRITE>\\
Notes:\\
<the notes to append, or the whole new notepad>\\
{[STOP]}
\end{promptbox}

\begin{promptbox}[label=prompt:playthrough_notepad_extract]{Notepad: Fact Extraction}
\ttfamily
You are playing [GAME] and are the strategist. You have just finished with these goals:\\
{[CLOSED]}

This is your notepad of working thoughts from pursuing them. It is about to be cleared:\\
{[NOTEPAD]}

List the facts in the notepad that are worth remembering for the rest of the game, long after these goals: where places are and how they connect, what people said, what items and Pokemon were found or used, what worked, what failed and why. Leave out what only mattered for finishing these goals.

Each fact is stored on its own, away from this notepad, so it must make sense without it: name the people, places and things it is about rather than saying "he", "there" or "it".

Write at most [MAX\_FACTS] facts, one per line. If nothing is worth remembering, write None.

Respond in exactly this format:\\
Reasoning: <what in the notepad will still matter later>\\
Facts:\\
- <fact>\\
- <fact>\\
{[STOP]}
\end{promptbox}

\subsubsection{Context Blocks}

\begin{promptbox}[label=prompt:playthrough_strategist_blocks]{Strategist: Context Blocks}
\ttfamily
The blocks substituted into the slots of the prompts above.

\textbf{Recent episodes.} Substituted for \texttt{[RECENT\_BLOCK]}, holding the task, status and summary of each of the last few episodes.

The last few episodes, oldest first:\\
{[RECENT]}

\textbf{Retry.} Appended to any prompt above whose reply could not be parsed, together with the reply itself and the specific problem with it.

Your previous reply was:\\
{[PREVIOUS]}

That reply could not be used: [PROBLEM] Respond again in the required format.

\textbf{Existing subgoals.} Substituted for \texttt{[EXTRAS]} when the goal being planned already has subgoals that are all finished with.

It already had these subgoals, which are all finished with:\\
{[CHILDREN]}

\textbf{Stalled goal.} Also substituted for \texttt{[EXTRAS]}, when the goal has been on the frontier for several episodes without being achieved.

You have been working towards this goal for [EPISODES] episodes and have not achieved it yet. Your notepad may say what went wrong.

\textbf{Route lookups.} Substituted for \texttt{[ROUTE\_OPTION]} and \texttt{[ROUTE\_FORMAT]} in the subgoal query prompt, and offered only when the player's current place is known.

You can also look up routes on your map of the places you have been. The player is at: [CURRENT]. For each place you want to know how to reach, write a `Route:` line saying where you want to go: a particular place you know by name, or a kind of place, such as "somewhere to heal your Pokemon". Unlike the questions, routes are always worked out from where the player is now.\\
Route: <where you want to go from here>

\textbf{Abandon option.} Substituted for \texttt{[ABANDON\_OPTION]} in the goal decision prompt, and offered only for goals below the seeded milestones.

ABANDON: the goal cannot be achieved, or is no longer worth pursuing. Drop it.

\end{promptbox}

\subsection{Playthrough Agent: Supervisor}

\begin{promptbox}[label=prompt:playthrough_primitives]{Supervisor: Action Primitives}
\ttfamily
Substituted for \texttt{[PRIMITIVES]} in the planning and decision prompts below.

1. MoveTo(description): walk to an object or entity that is visible on the screen, e.g. MoveTo(wooden signpost at (2, -4)).\\
2. MoveOff(direction description): walk off the edge of the screen in a direction, e.g. MoveOff(up, through the gap in the ledge) or MoveOff(left).\\
3. Interact(target): interact with the object or entity directly in front of the player, e.g. to talk to someone or read a sign. Name the target, e.g. Interact(wooden signpost).\\
4. Battle(optional instructions): fight the battle that is on screen, optionally with instructions, e.g. Battle() or Battle(use Thundershock and do not switch Pokemon).\\
5. Menu(specific instruction): use the menu system to do something specific, e.g. Menu(use a Potion on Pikachu). Only works in the main menu opened with Start, or in a menu that is already open.\\
6. Dialogue(): press through the dialogue that is on screen until it ends.\\
7. Free(instruction; stop when: condition): press single buttons (a, b, start, arrows) towards something specific, and say what to stop at. The condition must be something visible on screen, e.g. Free(walk up and down inside the patch of tall grass; stop when: a wild Pokemon appears). Use this when none of the others fit.
\end{promptbox}

\begin{promptbox}[label=prompt:playthrough_plan]{Supervisor: Subgoal Planning}
\ttfamily
You are playing [GAME] and are planning how to complete a task.

Your task: "[TASK]"\\
{[GUIDANCE\_BLOCK][LOG\_BLOCK]}\\
The image is the current game screen. Here is what has been recognised on it:\\
{[SCREEN\_TILES]}

These are the only things the player can be told to do:\\
{[PRIMITIVES]}

Break the task into subgoals, in order. Write each subgoal as the state of the game to reach, in plain words, not as one of the calls above: "the player is standing in front of the nurse", never "Interact(nurse)". Each subgoal should be a small step that one of the above can plausibly achieve from the screen it starts on, and should be worth checking off on its own. Write as few as the task needs. Start from where the player is now.

Respond in exactly this format:\\
Reasoning: <what the screen shows and how you intend to get the task done>\\
Subgoals:\\
- <first subgoal>\\
- <second subgoal>\\
...\\
{[STOP]}
\end{promptbox}

\begin{promptbox}[label=prompt:playthrough_decision]{Supervisor: Action Decision}
\ttfamily
You are playing [GAME] and deciding what the player should do next.

Your task: "[TASK]"\\
{[GUIDANCE\_BLOCK]}

Your plan:\\
{[PLAN]}

You are working on this subgoal: "[SUBGOAL]"\\
{[LOG\_BLOCK][ATTEMPTS\_BLOCK]}\\
The image is the current game screen. Here is what has been recognised on it:\\
{[SCREEN\_TILES]}

You can make the player do exactly one of the following:\\
{[PRIMITIVES]}

Pick the single action that makes the most progress on the subgoal from this screen.

Respond in exactly this format:\\
Reasoning: <what you see, and why this action is the best next step>\\
Action: <exactly one action from the list, written as a call, e.g. MoveTo(door of the Poke Mart)>\\
{[STOP]}
\end{promptbox}

\begin{promptbox}[label=prompt:playthrough_judge]{Supervisor: Subgoal Judgement}
\ttfamily
You are playing [GAME] and are checking what the player just did.

Your task: "[TASK]"\\
{[GUIDANCE\_BLOCK]}\\
You are working on this subgoal: "[SUBGOAL]"

You asked the player to: [REQUEST]

The first image is the screen before, and the second image is the screen after. The player is always drawn in the centre of the screen, so when the player walks, the rest of the screen shifts the other way.

Before, this had been recognised on the screen, with the player at (0, 0):\\
{[SCREEN\_TILES]}\\
{[AFTER\_BLOCK][TEXT\_BLOCK]}\\
The game reported: [EXECUTOR\_REPORT]\\
That report can be wrong about where the player ended up, so judge from the images.

Say whether the subgoal is now finished, and write a short note for the log of this playthrough. The note is the only record that survives, so say what actually happened and anything learned about the game or the map.

Respond in exactly this format:\\
Reasoning: <compare the two screens and the text>\\
Subgoal complete: <yes or no>\\
Summary: <one or two sentences for the log>\\
{[STOP]}
\end{promptbox}

\begin{promptbox}[label=prompt:playthrough_reconsider]{Supervisor: Reconsider}
\ttfamily
You are playing [GAME] and are stuck.

Your task: "[TASK]"\\
{[GUIDANCE\_BLOCK]}

Your plan:\\
{[PLAN]}

You have been trying this subgoal and it is not working: "[SUBGOAL]"\\
{[LOG\_BLOCK]}\\
What you tried for this subgoal:\\
{[ATTEMPTS]}

The image is the current game screen. Here is what has been recognised on it:\\
{[SCREEN\_TILES]}

Decide whether to carry on with a new plan or to give up. Give up only if the task cannot be done from here at all, for example because it needs something the player does not have and cannot get.

If you carry on, write a fresh plan for what is left, starting from the current screen. It may drop the subgoal that failed, or reach it another way.

Respond in exactly this format:\\
Reasoning: <why it is not working, and what to do about it>\\
Decision: <replan or quit>\\
Reason: <if quitting, why the task cannot be done; otherwise None>\\
Subgoals:\\
- <first subgoal of the new plan, if replanning>\\
- <second subgoal>\\
...\\
{[STOP]}
\end{promptbox}

\begin{promptbox}[label=prompt:playthrough_task_check]{Supervisor: Task Check}
\ttfamily
You are playing [GAME] and have worked through your plan.

Your task: "[TASK]"\\
{[GUIDANCE\_BLOCK][LOG\_BLOCK]}\\
The image is the current game screen. Here is what has been recognised on it:\\
{[SCREEN\_TILES]}

Decide whether the task is now complete.

Respond in exactly this format:\\
Reasoning: <what was achieved, and whether it adds up to the task>\\
Task complete: <yes or no>\\
Summary: <one or two sentences describing how the playthrough went>\\
{[STOP]}
\end{promptbox}

\begin{promptbox}[label=prompt:playthrough_clarify]{Supervisor: Target Clarification}
\ttfamily
You are playing [GAME] and deciding what the player should do next.

Your task: "[TASK]"\\
{[GUIDANCE\_BLOCK]}\\
The image is the current game screen. Here is what has been recognised on it:\\
{[SCREEN\_TILES]}

You asked the player to move to "[TARGET]", but that description matches more than one thing on the screen:\\
{[CANDIDATES]}

Rewrite the description so that it matches exactly one thing on the screen: the one you meant. Include its coordinates and whatever sets it apart from the others.

Respond in exactly this format:\\
Reasoning: <which of the candidates you meant, and why>\\
Target: <the new description>\\
{[STOP]}
\end{promptbox}

\begin{promptbox}[label=prompt:playthrough_menu_objective]{Supervisor: Menu Objective}
\ttfamily
You are playing [GAME] and are about to use the game's menus.

Your task: "[TASK]"\\
{[GUIDANCE\_BLOCK]}\\
The image is the current game screen. Here is what has been recognised on it:\\
{[SCREEN\_TILES]}

The player is currently [AGENT\_STATE].

You decided to use the menus for this: "[INSTRUCTION]"

Write one clear, specific objective for whoever operates the menus. They see only the screen and your objective, not the task, so name exactly what should end up selected, used, or read.

Then say whether this needs the main menu, the one opened with the Start button from the overworld, which holds the Pokedex, your Pokemon, the bag and the trainer card. Answer no if it is about a menu that is already open on the screen, such as a PC, a shop, a yes/no choice or a battle menu.

Respond in exactly this format:\\
Reasoning: <what the task needs from the menus>\\
Objective: <one sentence, specific>\\
Start menu: <yes or no>\\
{[STOP]}
\end{promptbox}

\begin{promptbox}[label=prompt:playthrough_battle_instructions]{Supervisor: Battle Instructions}
\ttfamily
You are playing [GAME] and a battle has just started.

Your task: "[TASK]"\\
{[GUIDANCE\_BLOCK]}\\
The image is the current battle screen.

Say how this battle should be fought, for whoever gives the orders in it. Keep it short: which Pokemon to use, which kind of move to favour, whether to try to run, or whether to just win normally. Answer None if there is nothing special to say and the battle should simply be won.

Respond in exactly this format:\\
Reasoning: <what the battle screen shows and what it means for the task>\\
Instructions: <one short sentence, or None>\\
{[STOP]}
\end{promptbox}

\begin{promptbox}[label=prompt:playthrough_supervisor_blocks]{Supervisor: Context Blocks}
\ttfamily
\textbf{Guidance block.} Included only when the strategist supplied guidance with the task:

Guidance for this task:\\
{[GUIDANCE]}

\textbf{Log block.} The narrative accumulated so far in this episode, one line per attempt:

What has happened so far, in order:\\
{[LOG]}

\textbf{Attempts block.} Shown on a retry, so the model can see what it has already tried:

What you already tried for this subgoal, and what came of it:\\
{[ATTEMPTS]}

\textbf{After block.} Included in the judgement prompt when the player ended in free roam:

After it, this is recognised on the screen, with the player at (0, 0):\\
{[AFTER\_TILES]}

\textbf{Text block.} Included when text was read during the action:

Text that appeared on screen during it:\\
{[TEXT]}
\end{promptbox}

\subsection{Playthrough Agent: Executors and Perception}
\label{app:prompts_executors}

These prompts are issued by the action primitives of Appendix~\ref{app:playthroughagent} and by
the tile recogniser that supplies \texttt{[SCREEN\_TILES]} to every layer above them.

\begin{promptbox}[label=prompt:playthrough_move_target]{MoveTo: Target Location}
\ttfamily
You are helping a player move to something on the screen in the Game Boy game Pokemon Red.

The player wants to move to: "[TARGET]"

The image is the current game screen. [GRID\_DESCRIPTION] Here is what has been recognised on the screen:\\
{[SCREEN\_TILES]}

Find the exact cell of the target.\\
- If exactly one thing on the screen matches the target, give its coordinates. If it covers several cells, give the one closest to the player.\\
- If more than one thing on the screen could be the target, answer Ambiguous and list every candidate.\\
- If nothing on the screen matches the target, answer Not Found.

Respond in exactly this format:\\
Reasoning: <what matches the target and where>\\
Location: <(x, y), Ambiguous, or Not Found>\\
Candidates: <if Ambiguous, each candidate with its coordinates and what sets it apart from the others; otherwise None>\\
{[STOP]}
\end{promptbox}

\begin{promptbox}[label=prompt:playthrough_move_edge]{MoveOff: Edge Selection}
\ttfamily
You are helping a player walk off the edge of the screen in the Game Boy game Pokemon Red.

The player wants to: "[TARGET]"

The image is the current game screen. [GRID\_DESCRIPTION] Here is what has been recognised on the screen:\\
{[SCREEN\_TILES]}

These cells on each edge of the screen can be walked on:\\
{[EDGE\_CELLS]}

Choose the edge cell the player should walk to so that they can then keep walking off the screen as described. Pick a cell from the list above on the edge in the direction the player wants to go. If no listed cell fits the description, answer Not Found.

Respond in exactly this format:\\
Reasoning: <which edge and which cell, and why>\\
Direction: <up, down, left or right>\\
Location: <(x, y) or Not Found>\\
{[STOP]}
\end{promptbox}

\begin{promptbox}[label=prompt:playthrough_interaction_cell]{Interact: Target Cell}
\ttfamily
You are playing Pokemon. The image is the current screen, divided into a grid of 16x16 cells. The player is at (0, 0), x increases to the right and y increases upward.

The target to interact with is: [TARGET]\\
{[NEIGHBOURHOOD\_BLOCK]}\\
Which of the eight cells around the player holds the target? The player can only interact with something directly above, below, to the left or to the right, never diagonally, so prefer one of those if the target covers several cells. Answer None if the target is not in any of the eight cells.

Respond in exactly this format:\\
Reasoning: <where the target is on the screen>\\
Cell: <(x, y) of one of the eight cells around the player, or None>\\
{[STOP]}
\end{promptbox}

\begin{promptbox}[label=prompt:playthrough_battle_step]{Battle: Turn Decision}
\ttfamily
You are playing Pokemon Red and are in a battle.

{[INSTRUCTIONS]}\\
The image is the current battle screen.\\
{[HISTORY]}\\
Choose what to do next:\\
- fight: attack, and say which of the four moves to use (1 is the top one)\\
- progress: press through battle text, or continue when there is no choice to make\\
- run: try to escape. This never works against another trainer\\
- bag: open the bag to use an item. This hands over to the menu system, so only choose it if an item is needed\\
- pokemon: open your team to switch. This hands over to the menu system, so only choose it if switching is needed

Respond in exactly this format:\\
Reasoning: <what the screen shows and what to do about it>\\
Action: <one of fight, progress, run, bag, pokemon>\\
Move: <1, 2, 3 or 4 if the action is fight, otherwise None>\\
{[STOP]}
\end{promptbox}

\begin{promptbox}[label=prompt:playthrough_menu_open]{Menu: Main Menu Entry}
\ttfamily
You are playing Pokemon Red and are about to open the main menu with the Start button.

Objective: "[TASK]"

The image is the current game screen.

Which entry of the main menu should be opened first?\\
- pokedex: the Pokedex\\
- pokemon: your Pokemon team, for checking, reordering, healing or using their moves outside battle\\
- bag: your items\\
- trainer: your trainer card

Respond in exactly this format:\\
Reasoning: <which entry the objective needs and why>\\
Entry: <one of pokedex, pokemon, bag, trainer>\\
{[STOP]}
\end{promptbox}

\begin{promptbox}[label=prompt:playthrough_menu_step]{Menu: Cursor Press}
\ttfamily
You are playing Pokemon Red and are using the game's menus.

Objective: "[TASK]"

The image is the current screen.\\
{[HISTORY]}\\
First decide whether the objective has already been achieved on this screen. If it has not, choose the single button press that best moves towards it.

The presses you can make:\\
- up, down, left, right: move the cursor\\
- confirm: press A to choose what the cursor is on\\
- back: press B to go back, or to close the current menu

Respond in exactly this format:\\
Reasoning: <what is on the screen, and what should be pressed next>\\
Complete: <yes or no>\\
Press: <one of up, down, left, right, confirm, back>\\
{[STOP]}
\end{promptbox}

\begin{promptbox}[label=prompt:playthrough_free_step]{Free: Button Press}
\ttfamily
You are playing Pokemon Red, pressing one button at a time.

What you have been asked to do: "[TASK]"

Stop when: [TERMINATION]

The image is the current screen.\\
{[HISTORY]}\\
First decide, from this screen, whether the condition to stop at has been met. If it has not, choose the single button press that best moves towards it.

The buttons are: a, b, start, up, down, left, right.

Respond in exactly this format:\\
Reasoning: <what is on the screen, and whether the stopping condition is met>\\
Complete: <yes or no>\\
Press: <one of a, b, start, up, down, left, right>\\
{[STOP]}
\end{promptbox}

\begin{promptbox}[label=prompt:playthrough_tile_classify]{Perception: Tile Classification}
\ttfamily
You are looking at a screen from the Game Boy game Pokemon Red. The screen is divided into a grid of 8x8 pixel tiles.

You are given three images:\\
1. The full game screen, with one tile outlined by a red box.\\
2. The contents of that tile on its own.\\
3. The 16x16 pixel grid cell that contains the tile. The tile is the [QUARTER] quarter of this cell.

Classify what is in the outlined tile into exactly one of these categories:\\
- Ground: plain navigable ground the player can walk on, including ground with a repeating texture\\
- Ground Decoration: walkable ground with small decorations on it, such as flowers or little tufts that look like tiny Digletts\\
- Ledge: a ledge the player can jump down from one side\\
- Obstacle: a boundary or obstacle that blocks movement and cannot be interacted with (trees, rocks, walls, fences, buildings)\\
- Entrance: a door or entrance the player can walk into to enter or leave a building, cave or another area, including doorways, cave mouths, gates between routes, and the exit mats on the floor at the edge of a room\\
- Water\\
- Tall Grass: tall grass where wild Pokemon can appear\\
- Interactable Object: an object the player can interact with (machines, computers, bookshelves, etc.)\\
- Item: an item lying on the ground that the player can pick up. An item on the ground always looks like a Poke Ball\\
- NPC: a non-player character\\
- Sign: a sign the player can read\\
- Staircase: stairs, a ladder, or a hole in the floor that takes the player up or down to another floor or level\\
- Elevator: an elevator, or the panel or floor of an elevator, that moves the player between floors\\
- Special Ground Tile: a tile the player can walk on that may move them around, such as arrow or spinner tiles that push the player, ice floors the player slides across, or teleportation pads\\
- Unknown: none of the above, or you cannot tell\\
{[PLAYER\_NOTE]}\\
If the category is NPC, also give a short name or description of the NPC and where it is on the screen.\\
If the category is Sign, also give where it is on the screen.

Respond in exactly this format:\\
Reasoning: <brief description of what is in the tile and its surroundings>\\
Category: <one category from the list>\\
Description: <a few words naming the whole object or thing this tile is part of, specific enough to tell it apart from similar things. Name the whole object, not which part of it this tile shows: write "wooden signpost", never "top-left corner of a signpost". Other examples: "Poke Ball on the floor", "door of the Poke Mart", "Pokemon Center nurse">\\
Name: <NPC name or description, or None>\\
Location: <where the NPC or sign is on the screen, or None>\\
{[STOP]}
\end{promptbox}

\begin{promptbox}[label=prompt:playthrough_executor_blocks]{Executors and Perception: Context Blocks}
\ttfamily
The blocks substituted into the slots of the prompts above.

\textbf{Grid description.} Substituted for \texttt{[GRID\_DESCRIPTION]} in both movement prompts.

The screen is a grid of cells. The player is at (0, 0). x increases to the right, from -4 at the left edge to 5 at the right edge. y increases upward, from 4 at the top edge to -4 at the bottom edge.

\textbf{Neighbourhood.} Substituted for \texttt{[NEIGHBOURHOOD\_BLOCK]} in the interaction prompt, holding the recognised contents of the eight cells around the player.

This is what has already been recognised in the eight cells around the player. Use it together with the image:\\
{[NEIGHBOURHOOD]}

\textbf{Player note.} Substituted for \texttt{[PLAYER\_NOTE]} when the tile being classified lies next to the player, whose sprite is drawn larger than its own cell.

Note: the player character stands in the grid cell [DIRECTION] this one. The player sprite is drawn slightly larger than its own cell, so part of it (for example the top of the player's cap) may spill into the [SIDE] of this cell. Ignore any part of the player sprite: it is not an NPC or an object, and you should classify only what is underneath or around it.

\end{promptbox}

\section{Hardware}
Experiments were run on multiple clusters with varying GPU configurations: 

\begin{enumerate}
\item Most open-source LLM experiments were run on nodes with 8xA40 NVIDIA GPUs. Some results were collected on nodes with 4xH200 NVIDIA GPUs. 
\item All API model results used the OpenRouter API and did not require GPU access. 
\item Curiosity-based exploration, observation embedder training and world model training were primarily conducted on nodes with 1xV100 NVIDIA GPUs. 
\item Autonomous Skill Discovery was run on a node with 8xA100 NVIDIA GPUs
\end{enumerate}

\section{Use of LLMs}
We used LLMs for the following:
\begin{itemize}
    \item Help with code implementation (Claude Code). No design decisions were made unilaterally by the LLM. 
    \item Appendix writing: Claude Code was used to verbalize the experimental details and hyperparameters during the writing of certain Appendix sections. All text was checked by the authors to confirm veracity.
    \item Error and spell checking. 
\end{itemize}

We did not use LLMs for any of the following:
\begin{itemize}
    \item Ideation/project design
    \item Interpretation of results
\end{itemize}

\end{document}